\documentclass[]{seed/bytedance_seed}
\usepackage[T1]{fontenc}

\usepackage[toc,page,header]{appendix}

\usepackage{amsmath,amsfonts,bm}

\def\eqref#1{equation~\ref{#1}}

\def\1{\bm{1}}

\DeclareMathAlphabet{\mathsfit}{\encodingdefault}{\sfdefault}{m}{sl}
\SetMathAlphabet{\mathsfit}{bold}{\encodingdefault}{\sfdefault}{bx}{n}

\usepackage{hyperref}
\usepackage{xurl}
\usepackage{graphicx}
\usepackage{float}
\usepackage{makecell}

\usepackage{minitoc}
\usepackage{amsfonts}
\usepackage{amsmath}
\usepackage{booktabs}
\usepackage{algorithm}
\usepackage{algpseudocode}
\usepackage{cleveref}
\usepackage{pgfplots}
\pgfplotsset{compat=1.18}
\usepackage{xspace}
\usepackage{amsmath,amssymb}
\usepackage{hyperref}
\usepackage{url}
\usepackage{cleveref}
\usepackage{pgfplots}
\pgfplotsset{compat=1.18}
\usepackage{graphicx}
\usepackage{array}
\usepackage{longtable}
\usepackage{multirow}

\DeclareRobustCommand{\modify}[1]{#1}
\usepackage[most]{tcolorbox}

\newtcblisting{promptbox}{
  listing only,
  breakable,                      
  colback=gray!5,
  colframe=gray!40,
  boxrule=0.4pt,
  left=2mm, right=2mm, top=1.5mm, bottom=1.5mm,
  listing options={
    basicstyle=\ttfamily\scriptsize,
    breaklines=true,              
    breakatwhitespace=true,
    columns=fullflexible,
    keepspaces=true,
  },
}

\title{SAE Interprets MoE: Sparse AutoEncoders Reveal\\That MoE Experts Exhibit Human-Interpretable \\Fine-Grained Specialization}

\author[1,2]{Anyuan Zhuo}
\author[1]{Shaohua Zhang}
\author[1,\dagger]{Yuan Lin}
\author[2,\dagger]{Pinyan Lu}

\affiliation[1]{ByteDance Seed}
\affiliation[2]{Shanghai University of Finance and Economics}

\contribution[\dagger]{Corresponding author}

\abstract{

Mixture-of-Experts (MoE) models route each token to a few of many experts, yet what an
individual expert is responsible for remains unsettled. \modify{Prior analyses count how often
each expert is selected on domain-labeled corpora. However, such labels impose a coarse taxonomy
in advance, and selection frequency ignores how much an expert actually contributes. We instead
introduce a general framework for analyzing specialization in feature space by training a sparse
autoencoder (SAE) on the output of any MoE layer. Because this output is a router-weighted sum
of expert outputs and the SAE encoder is affine, each latent's pre-activation decomposes into
exact per-expert contributions. Aggregating these contributions over a corpus gives each expert
a dominant set of latents that it primarily drives. Applying this framework to layers 4, 30 and
46 of Qwen3-30B-A3B, we find that expert contributions are far more concentrated than gating
weights, and most latents are dominated by only one or two experts. Across experts, dominant
sets are largely disjoint. Auto-interpreting these sets reveals a hierarchical division of
labor. At a coarse level, experts cluster into domains of very unequal sizes, led by code,
mathematics and text formatting. Within each domain, individual experts specialize in distinct
subfields, such as Java versus front-end web code, or geometry versus arithmetic and
polynomials. Finally, substantial feature overlap occurs almost exclusively between neighboring
subfields of the same domain.}
Our project page is at \url{https://bytedance-seed.github.io/SAE-interprets-MoE/}, where we provide open-source code and links to the SAE models.
}

\correspondence{\email{{linyuan.0@bytedance.com}}, \email{{lu.pinyan@mail.shufe.edu.cn}}}

\begin{document}
\maketitle

\begin{figure}[htb]
    \centering
    \includegraphics[width=\textwidth]{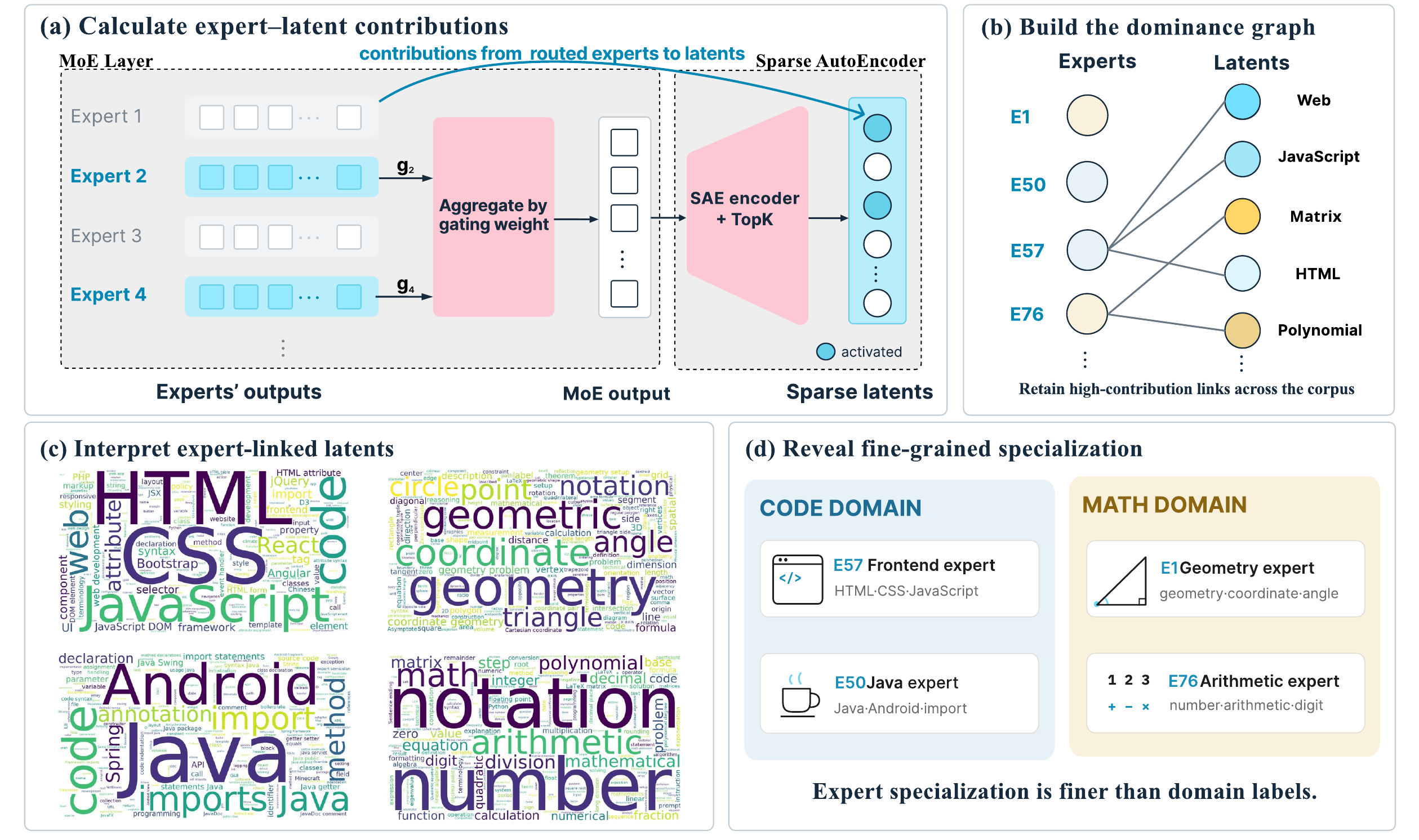}
    \caption{An MoE layer outputs a router-weighted sum of its
    active experts (left), and since the SAE encoder is affine, each latent's activation
    decomposes exactly over these experts (middle). Thresholding the contribution fractions
    averaged over a corpus gives the dominance graph (right), and auto-interpreting the latents an expert
    dominates describes its role (bottom), such as a Java expert (E50) and a front-end web
    expert (E57) at layer 30 of Qwen3-30B-A3B.}
    \label{fig:teaser}
\end{figure}

\section{Introduction}
\label{sec:intro}

Mixture-of-Experts (MoE) architectures underlie much of the recent progress in large language
models. \modify{By routing each token to a small subset of experts, an MoE model increases
parameter capacity while keeping per-token inference cost close to a much smaller dense model.
Although this design relies on the router, its learned routing mechanism has received little
scrutiny. How a trained model assigns tokens to experts, and what each expert is ultimately
responsible for, remains poorly understood.}

The existing evidence is mixed. \modify{Some studies report domain-specialized experts that fire
preferentially on specific domains~\citep{liu2024deepseek,muennighoff2025olmoe}. Others conclude
that routing does not follow human-defined domains and instead reflects surface properties such
as token identity~\citep{jiang2024mixtral,xue2024openmoe}. Yet both sides rely on the same
method, counting how often each expert is selected on domain-labeled corpora such as GitHub for
code or arXiv for academic papers.} This approach has four limitations: (a) \modify{The taxonomy
is imposed by the analyst rather than recovered from the model, testing agreement with a preset
partition rather than uncovering what the model learned.} (b) The labels available at corpus
scale are coarse, typically a handful of categories such as code and math. This is far coarser
than the division of labor one would expect among the tens to hundreds of experts in a single
layer. (c) The label attaches to a corpus while the routing decision being measured is made per
token. For example, a math problem statement may contain a great deal of story narrative
framing, and counting every token of it toward ``math'' blurs precisely the distinctions the
analysis is meant to detect. (d) \modify{Selection counts ignore router weights, even though a
selected expert with a small weight contributes little to the layer output.}

\modify{We introduce a general framework for analyzing expert specialization in the feature
space of any MoE layer.} We train a
Sparse AutoEncoder (SAE;~\citealp{huben2024sparse}) on the output of an MoE layer. SAE latents
approximate the features the model itself uses, and are discovered from its activations rather
than stipulated in advance. Moreover, they also admit a natural decomposition into the
contributions of individual experts. The output of an MoE layer is a router-weighted sum of the
outputs of the active experts; because the SAE encoder is affine in that sum, each latent's
pre-activation decomposes exactly over the experts that produced it. \modify{For every (expert,
latent, token) triple, this gives the expert's share of the latent's activation. Aggregating
these shares over a corpus yields an expert-level contribution profile (\Cref{fig:teaser}).}

\modify{As a general analysis framework, our method improves upon selection-count approaches
along two axes. In the unit of analysis, we replace analyst-defined domains with unsupervised
SAE latents. These latents require no preset taxonomy and capture much finer distinctions than
corpus-level labels.} In
resolution, we replace corpus-level selection frequency with a per-token, per-expert
contribution score, which both avoids attributing every token of a document to a single domain
and measures functional contribution rather than mere co-occurrence. \modify{Moreover, the
framework requires only expert outputs and a layer-output SAE, making it applicable to any MoE
model without predefined labels.}

\modify{To demonstrate this framework, we apply it to layers 4, 30, and 46 of
Qwen3-30B-A3B~\citep{yang2025qwen3}, covering shallow, mid-late, and deep positions in the
network.} Our first observation is that contribution is highly concentrated. Most latents receive the bulk of their activation from a small number of experts. Building on this, we say that an expert \emph{dominates} a latent when it is a principal contributor to that latent's activation, and we collect, for each expert, the set of latents it dominates. Interpreting the latents in an expert's dominant set therefore describes what that expert does in the model's own vocabulary rather than in categories chosen in advance. We obtain such interpretations by auto-interpreting each latent from activating contexts and logit effects (\Cref{app:autointerp}).

\modify{Dominant sets are largely disjoint across experts. About half of the latents in an expert's
dominant set are unique to that expert, and most expert pairs share almost no latents. Interpreting these sets at layer 30 reveals a hierarchical
division of labor. First, grouping the interpreted experts recovers coarse domains of strongly
unequal sizes, with large allocations to computing, mathematics and text formatting. Second,
each coarse domain is divided among experts that specialize in finer subfields, such as Java
versus front-end web code. Third, these subfields are not a hard partition. Meaningful overlap
between dominant sets concentrates within coarse domains and links experts on neighboring
subfields. Together, these results show that MoE experts organize into coarse domains internally
divided into largely distinct, locally overlapping subfields.}

\section{Related work}
\label{sec: related}

\paragraph{Expert specialization in MoE models.}
Prior analyses report either domain-specialized
experts~\citep{liu2024deepseek,muennighoff2025olmoe} or routing driven by surface properties
such as token identity~\citep{jiang2024mixtral,xue2024openmoe}, and both rest on selection
counts over domain-labeled corpora, whose limitations we discuss in \Cref{sec:intro}.
\modify{\citet{herbst2026expertstrikesbackinterpreting} move beyond coarse domains. Motivated
by $k$-sparse probing showing that expert neurons are less polysemantic than dense-network
neurons, they automatically label hundreds of whole experts as fine-grained task specialists
rather than domain or token-level processors. Their specialization score uses clusters of the
model's own output embeddings, keeping the categories model-derived. However, because the
cluster count is chosen from a small hand-set grid, the score only selects the best-fitting
preset resolution rather than uncovering the model's own granularity. By contrast, our SAE
latents are learned unsupervised, allowing each expert's dominant set and its natural
granularity to emerge directly without matching against a preset partition.}

\paragraph{SAEs, interpretability and tracing.}
Our unit of analysis comes from dictionary learning. \modify{Because a model represents more
features than it has dimensions, it stores them in
superposition~\citep{elhage2022toymodelssuperposition}, making individual neurons polysemantic.
An SAE addresses this by learning an overcomplete sparse dictionary of much more monosemantic
directions~\citep{huben2024sparse}, an approach since scaled to frontier models and millions of
latents~\citep{templeton2024scaling,gao2025scaling}.} We adopt the top-$k$ variant of
\citet{gao2025scaling} in \Cref{subsec:sae} and read latents by auto-interpretation from
activating contexts and logit effects (\Cref{app:autointerp}),
following~\citet{paulo2024automatically}. \modify{SAE latents are also used as nodes in sparse
feature circuits~\citep{marks2025sparse} and attribution graphs~\citep{ameisen2025circuit} to
trace computation across layers and components. By contrast, we decompose each latent's
activation within a single MoE layer into per-expert contributions.} The decomposition of the
pre-activation is also exact rather than approximate, since \Cref{eq:decomposition} follows
algebraically from the linearity of the layer output in the expert outputs.

\section{Method}
\label{sec: method}

\subsection{The MoE block linearly combines expert outputs into the layer output}
\label{subsec:moe-block}

Consider an MoE LLM with hidden dimension $d$, where one MoE block contains $N$ experts.
Given a token sequence $x = (x_1, \dots, x_T)$, let $x_{1:t}=(x_1, \dots, x_t)$ be its prefix and
$h(t) \in \mathbb{R}^d$ be the corresponding input to the MoE block.
The router scores this input, takes the top-$k$ experts and renormalizes their weights.
\modify{It assigns each expert $j$ a non-negative weight $g_j(t) \in [0,1)$, where only $k$
weights are positive and $\sum_{j=1}^{N} g_j(t) = 1$. We call an expert with $g_j(t) > 0$ an
activated expert. Let $\mathcal{A}(t)$ be the set of activated experts and
$u_j(t) \in \mathbb{R}^d$ be the output of expert $j$. The layer output is}
\begin{equation}
  \label{eq:layer-output}
  y(t) = \sum\nolimits_{j \in \mathcal{A}(t)} g_j(t)\, u_j(t) \in \mathbb{R}^d.
\end{equation}

\subsection{Encoding the MoE output into a latent space with an SAE}
\label{subsec:sae}

We train an SAE whose objective is to reconstruct $y$, with latent dimension
$m \gg d$. Its encoding and decoding are
\begin{align}
  f(t) &= \operatorname{TopK-ReLU}_\kappa\!\bigl(W_{\mathrm{enc}}\, y(t) + b_{\mathrm{enc}}\bigr),
  \label{eq:encode} \\
  \hat{y}(t) &= W_{\mathrm{dec}}\, f(t) + b_{\mathrm{dec}} \approx y(t),
  \label{eq:decode}
\end{align}
where $W_{\mathrm{enc}} \in \mathbb{R}^{m \times d}$,
$W_{\mathrm{dec}} \in \mathbb{R}^{d \times m}$, and the columns of $W_{\mathrm{dec}}$ have unit
norm. The $i$-th component $f_i(t)$ of $f(t)$ is the activation magnitude of latent
$i$ on this prefix, and at most $\kappa$ components are non-zero. We denote the activated
set by $\mathcal{F}(t) = \{\, i \mid f_i(t) > 0 \,\}$.

\subsection{The contribution of an expert to a latent}
\label{subsec:contribution}

We observe that the activated part of a latent is an affine transformation of $y$. Let
$W_{\mathrm{enc},(i)} \in \mathbb{R}^d$ denote the $i$-th row of the encoder weight and
$b_{\mathrm{enc},(i)}$ the $i$-th entry of the encoder bias. Then, for an activated latent $i$,
\begin{equation}
  \label{eq:affine}
  f_i(t) = W_{\mathrm{enc},(i)} \cdot y(t) + b_{\mathrm{enc},(i)}.
\end{equation}
Substituting \Cref{eq:layer-output} into \Cref{eq:affine} gives
\begin{equation}
  \label{eq:decomposition}
  f_i(t) = \sum\nolimits_{j \in \mathcal{A}(t)} g_j(t)\, W_{\mathrm{enc},(i)} \cdot u_j(t)
             + b_{\mathrm{enc},(i)}.
\end{equation}
\Cref{eq:decomposition} decomposes the activation of a latent exactly into the encoder
bias plus the sum of the contributions of several experts. \modify{For an activated expert
$j \in \mathcal{A}(t)$ and an activated latent $i \in \mathcal{F}(t)$, we define the positive
contribution of $j$ to $i$ as}
\begin{equation}
  \label{eq:contribution}
  c(j,i,x_{1:t}) = \operatorname{ReLU}\!\bigl(g_j(t)\, W_{\mathrm{enc},(i)} \cdot u_j(t)\bigr).
\end{equation}
A negative contribution strong enough to suppress a latent leaves it inactive
and is never observed, so any statistics over negative contributions would be
systematically biased. We therefore keep only the positive part in
\Cref{eq:contribution} and use it exclusively in all subsequent analyses.

\subsection{Computing the contribution fraction}
\label{subsec:fraction}

On a single prefix, the relative magnitude of an expert's contribution to a latent
reflects whether this expert is important for that latent. Normalizing $c$ over the set
of activated experts, we obtain
\begin{equation}
  \label{eq:token-fraction}
  C(j,i,x_{1:t}) =
    \frac{c(j,i,x_{1:t})}{\sum_{j' \in \mathcal{A}(t)} c(j',i,x_{1:t})}
\end{equation}
for $j \in \mathcal{A}(t)$ and $i \in \mathcal{F}(t)$, and $C(j,i,x_{1:t}) = 0$ otherwise.
We are interested in which latents an expert is generally better at contributing to, and
therefore take the expectation of $C$ over the entire dataset $X$, obtaining
\begin{equation}
  \label{eq:expected-fraction}
  \mathcal{C}(j,i) =
  \mathop{\mathbb{E}}\nolimits_{(x,t) \,\sim\, \mathcal{U}\{(x,t)\,\mid\, x \in X,\; 1 \le t \le \lvert x \rvert\}}
  \left[C(j,i,x_{1:t})\right].
\end{equation}
Finally, in order to assess whether an expert is generally important for a latent, we
normalize $\mathcal{C}$ over the set of experts,
\begin{equation}
  \label{eq:contribution-fraction}
  s(j,i) = \frac{\mathcal{C}(j,i)}{\sum_{j'=1}^{N} \mathcal{C}(j',i)}.
\end{equation}
We call \Cref{eq:contribution-fraction} the contribution fraction of $j$ to $i$.

\subsection{Dominance graph}
\label{subsec:dominance}

Given a threshold $\tau$, $s(j,i) \ge \tau$ indicates that $j$ generally makes a substantial
contribution to $i$; in this case we say that $j$ dominates $i$. We define the
dominant set of an expert $j$ to be the set consisting of all latents dominated by $j$,
denoted $\mathcal{S}(j) = \{\, i \mid \text{$i$ is dominated by $j$} \,\}$.

We construct a bipartite graph $G(U,V,E)$, whose left vertex set $U$ is the expert set
and whose right vertex set $V$ is the latent set of the SAE. Its edge set is
$E = \{\, (j,i) \mid j \in U,\, i \in \mathcal{S}(j) \,\}$. We call this bipartite graph
the dominance graph.

\section{Qwen3-30B-A3B has specialized experts}
\label{sec:qwen3specialized}

We apply \Cref{sec: method} to layers 4, 30, and 46 of Qwen3-30B-A3B, which we select to
represent the model's shallow, mid-late, and deep-layer behaviors. The model has 48 layers in
total, where each MoE layer of this model contains $N=128$ experts, with $k=8$ experts
activated per token and no shared experts. We train one individual SAE with latent size of
$2^{20}$ for each of the three layers. Statistics are computed on a validation set of 1.5B
tokens that shares the same distribution as the SAE training set and has no data overlap with
it. Every latent of each SAE activates at least once on this set, so $s(j,i)$ is defined for
all latents. \modify{We first verify that our contribution score captures genuine feature
preferences beyond router gating weights (\Cref{subsec:gating-vs-contribution}). We then
analyze the dominance graph from two complementary perspectives to test for specialization,
examining feature degrees (\Cref{sec:dominance_graph meta}) and pairwise overlap across experts
(\Cref{subsec:dominance-overlap}).}

\subsection{\modify{Validating the contribution score against router gating}}
\label{subsec:gating-vs-contribution}

The contribution defined in \Cref{eq:contribution} carries $g_j(t)$ as a factor, and the gating
weights are themselves unevenly distributed. The router takes the top-$k$ experts and
renormalizes their weights, so higher-ranked experts naturally receive larger $g_j(t)$.
\modify{Even if experts had no preference over latents, uneven gating weights alone would make
their contributions unequal. Before adopting this statistic, we therefore test whether
contribution is concentrated beyond what gating weights alone can explain.}

\modify{We compare the two quantities after aggregating each by rank.} For each prefix
$x_{1:t}$ we sort the gating weights of the 8 activated experts in descending order, take the
value in position $r$ ($r = 0, \dots, 7$), and average it over the entire validation set; we
denote the result by $\bar{g}_r$. Analogously, for each (prefix, activated latent) pair
$(x_{1:t}, i)$ with $i \in \mathcal{F}(t)$, we sort the 8 values $C(j,i,x_{1:t})$ in descending
order, take the value in position $r$, and average it over all such pairs; we denote the result
by $\bar{C}_r$. \modify{Position $r$ can correspond to different experts on different tokens,
because we aggregate by rank rather than by expert identity. Because both the router weights
and the normalized contributions in \Cref{eq:token-fraction} sum to $1$, we have
$\sum_{r=0}^{k-1} \bar{g}_r = \sum_{r=0}^{k-1} \bar{C}_r = 1$, making the two rank profiles
directly comparable.}

\Cref{fig:rank-profile} reports the results for the three layers, and in all three $\bar{C}_r$
is markedly steeper than $\bar{g}_r$. \modify{At layer 30, for example, the top three gating
weights decay gently ($0.231$, $0.169$ and $0.137$). By contrast, the top three contribution
shares are $0.512$, $0.185$ and $0.115$, so the top-ranked expert alone accounts for over half
of the total contribution. Beyond rank $r = 1$, the contribution curve stays below the gating
curve, dropping to $\bar{C}_7 = 0.005$ versus $\bar{g}_7 = 0.074$ at $r = 7$.} Layers 4 and 46
exhibit the same shape.

The ReLU in \Cref{eq:contribution} could also steepen $\bar{C}_r$, \modify{since setting
truncated experts to $c = 0$ concentrates the normalized sum of $1$ on the remaining experts.}
The truncation is mild, however. Of the 8 activated experts of a (prefix, activated latent)
pair, on average only $2.09$, $1.29$ and $1.64$ are truncated at layers 4, 30 and 46, so most
activated experts keep a positive contribution and the truncation is not the main source of the
steepness.

The concentration of the contribution is therefore not an artifact of the top-$k$ routing
structure. \modify{While gating weights decay gently, expert contributions concentrate sharply
on one or two experts, confirming that $s(j,i)$ provides a meaningful signal for constructing
the dominance graph.}

\newcommand{\rankprofile}[4]{%
  \begin{tikzpicture}
    \begin{axis}[
      width = 0.33\textwidth, height = 0.25\textwidth,
      title = {Layer #1},
      xlabel = {rank $r$},
      xmin = -0.3, xmax = 7.3,
      ymin = 0,    ymax = 0.68,
      xtick = {0,1,2,3,4,5,6,7},
      grid = major, grid style = {gray!20},
      tick align = outside,
      legend to name = rankprofilelegend,
      legend columns = -1,
      legend cell align = left,
      legend style = {
        draw = none,
        /tikz/every even column/.append style = {column sep = 2em},
      },
      #4
    ]
      \addplot[thick, mark = *, color = black!65] coordinates {#2};
      \addlegendentry{gating $\bar g_r$}
      \addplot[thick, dashed, mark = square*, color = red!70!black] coordinates {#3};
      \addlegendentry{contribution $\bar{C}_r$}
    \end{axis}
  \end{tikzpicture}%
}

\begin{figure}[t]
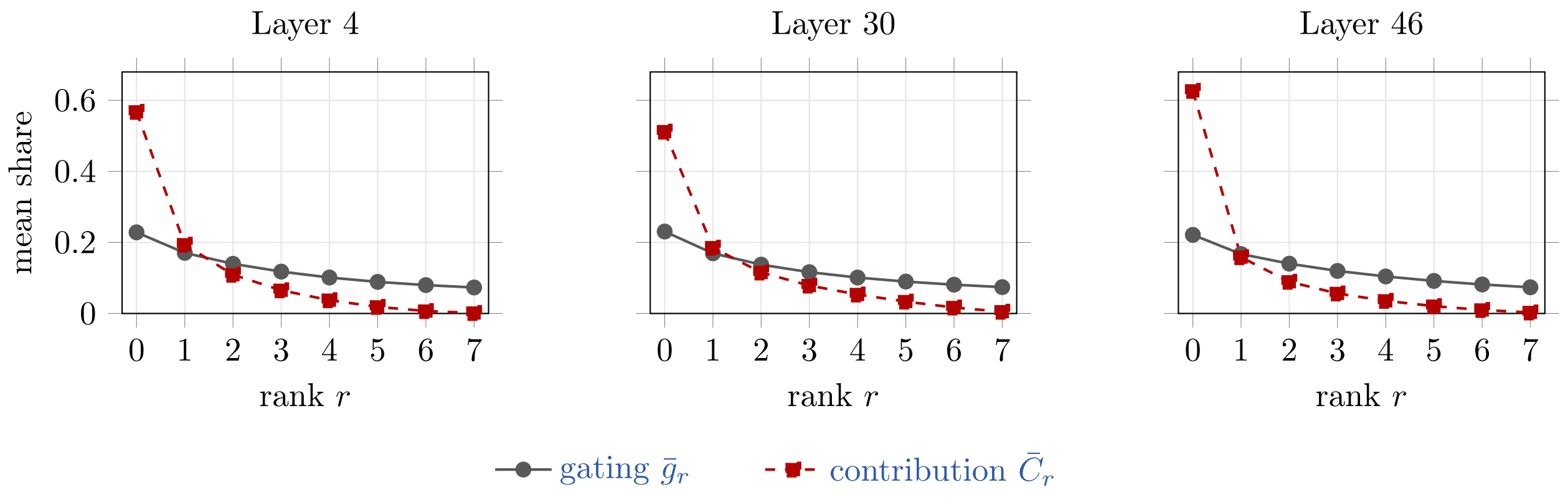

  \centering
  \rankprofile{4}{
    (0,0.228385) (1,0.170646) (2,0.139857) (3,0.117821)
    (4,0.101261) (5,0.088947) (6,0.079887) (7,0.073193)
  }{
    (0,0.567694) (1,0.193636) (2,0.109129) (3,0.065290)
    (4,0.037442) (5,0.018584) (6,0.006854) (7,0.001371)
  }{ylabel = {mean share}}\hfill
  \rankprofile{30}{
    (0,0.230709) (1,0.169310) (2,0.137491) (3,0.116342)
    (4,0.101124) (5,0.089777) (6,0.081058) (7,0.074186)
  }{
    (0,0.512017) (1,0.185317) (2,0.115389) (3,0.078591)
    (4,0.053420) (5,0.033443) (6,0.016858) (7,0.004964)
  }{yticklabels = {}}\hfill
  \rankprofile{46}{
    (0,0.221303) (1,0.167379) (2,0.140202) (3,0.119784)
    (4,0.104272) (5,0.091635) (6,0.081666) (7,0.073756)
  }{
    (0,0.626773) (1,0.158527) (2,0.089251) (3,0.056370)
    (4,0.035613) (5,0.020709) (6,0.009856) (7,0.002901)
  }{yticklabels = {}}

  \vspace{0.6em}
  \ref{rankprofilelegend}

  \caption{Rank profiles of the gating weight and of the contribution. $\bar{C}_r$ is markedly steeper than $\bar{g}_r$. }
  \label{fig:rank-profile}
\end{figure}

\subsection{\modify{From the feature side, most latents have one or two dominators}}
\label{sec:dominance_graph meta}

\modify{We build the dominance graph of \Cref{subsec:dominance} with $\tau = 0.2$ and first
examine it from the perspective of individual features. Note that $\tau = 0.2$ sits far above
the equal-share baseline $1/N \approx 0.008$, so every edge reflects a substantial
contribution. If experts specialize, most latents should have at least one dominator rather
than receiving small, even contributions from many experts. At the same time, each latent
should be dominated by only a few experts, well below the $k = 8$ experts selected per token.}

\begin{figure}[t]
  \centering
  \begin{tikzpicture}
    \begin{axis}[
      width = 0.47\textwidth, height = 0.26\textwidth,
      title = {(a) Latent-side degree},
      ybar = 0pt,
      bar width = 4.2pt,
      enlarge x limits = 0.16,
      xlabel = {number of dominators},
      ylabel = {share of latents (\%)},
      xtick = {0,1,2,3,4},
      ymin = 0, ymax = 72,
      grid = major, grid style = {gray!20},
      tick align = outside,
      legend pos = north east,
      legend cell align = left,
      legend style = {draw = none, fill = none, font = \scriptsize},
    ]
      \addplot[fill = blue!25,  draw = blue!60!black] coordinates {
        (0,7.03) (1,57.37) (2,32.51) (3,3.08) (4,0.009)};
      \addplot[fill = red!25,   draw = red!70!black]  coordinates {
        (0,13.61) (1,62.10) (2,23.04) (3,1.25) (4,0.002)};
      \addplot[fill = black!20, draw = black!70]      coordinates {
        (0,3.69) (1,56.65) (2,36.02) (3,3.63) (4,0.014)};
      \legend{Layer 4, Layer 30, Layer 46}
    \end{axis}
  \end{tikzpicture}\hfill
  \begin{tikzpicture}
    \begin{axis}[
      width = 0.47\textwidth, height = 0.26\textwidth,
      title = {(b) Expert-side degree},
      xlabel = {expert rank (sorted by degree)},
      ylabel = {dominated latents},
      xmin = -2, xmax = 129,
      xtick = {0,32,64,96,127},
      ymode = log,
      ymin = 60, ymax = 120000,
      grid = major, grid style = {gray!20},
      tick align = outside,
      legend pos = north west,
      legend cell align = left,
      legend style = {draw = none, fill = none, font = \scriptsize},
    ]
      \addplot[thick, mark = none, color = blue!60!black]
        table[x expr = \coordindex, y index = 0, row sep = \\, header = false] {
        362\\ 457\\ 547\\ 593\\ 712\\ 800\\ 872\\ 1155\\
        1217\\ 1243\\ 1451\\ 1679\\ 1702\\ 1939\\ 2016\\ 2496\\
        2618\\ 2660\\ 2709\\ 2807\\ 2967\\ 3087\\ 3112\\ 3299\\
        3334\\ 3684\\ 3702\\ 3759\\ 3778\\ 3818\\ 4006\\ 4009\\
        4203\\ 4208\\ 4217\\ 4228\\ 4262\\ 4290\\ 4510\\ 4664\\
        4680\\ 4814\\ 4939\\ 4973\\ 4984\\ 5026\\ 5182\\ 5363\\
        5492\\ 5584\\ 5618\\ 5744\\ 5940\\ 6017\\ 6100\\ 6331\\
        6965\\ 7050\\ 7246\\ 7296\\ 7522\\ 7597\\ 7664\\ 8000\\
        8040\\ 8085\\ 8467\\ 8487\\ 8505\\ 8520\\ 8553\\ 8557\\
        8576\\ 8683\\ 8724\\ 9067\\ 9207\\ 10315\\ 10367\\ 10743\\
        10996\\ 11000\\ 11057\\ 11593\\ 12122\\ 12140\\ 12379\\ 13992\\
        14014\\ 14225\\ 14274\\ 14517\\ 14762\\ 14944\\ 14945\\ 15268\\
        15532\\ 15722\\ 16108\\ 16403\\ 16759\\ 17342\\ 17500\\ 17572\\
        17656\\ 17899\\ 18583\\ 18798\\ 20774\\ 20805\\ 21381\\ 21771\\
        22122\\ 23313\\ 23570\\ 23679\\ 24443\\ 24589\\ 27819\\ 28527\\
        29517\\ 30029\\ 31032\\ 34288\\ 35686\\ 38410\\ 41150\\ 43412\\
      };
      \addlegendentry{Layer 4}

      \addplot[thick, mark = none, color = red!70!black, dashed]
        table[x expr = \coordindex, y index = 0, row sep = \\, header = false] {
        224\\ 1084\\ 1298\\ 1358\\ 1430\\ 1708\\ 2544\\ 2863\\
        3536\\ 3911\\ 3957\\ 4038\\ 4072\\ 4294\\ 4317\\ 4343\\
        4368\\ 4368\\ 4374\\ 4386\\ 4433\\ 4443\\ 4698\\ 4769\\
        4771\\ 4943\\ 5026\\ 5083\\ 5085\\ 5090\\ 5189\\ 5226\\
        5268\\ 5288\\ 5295\\ 5415\\ 5486\\ 5700\\ 5706\\ 5906\\
        5923\\ 6012\\ 6020\\ 6178\\ 6377\\ 6471\\ 6479\\ 6549\\
        6565\\ 6671\\ 6681\\ 6695\\ 6854\\ 6881\\ 6920\\ 6930\\
        6952\\ 6971\\ 7061\\ 7116\\ 7280\\ 7324\\ 7327\\ 7337\\
        7453\\ 7805\\ 7820\\ 8113\\ 8175\\ 8324\\ 8398\\ 8719\\
        8891\\ 9093\\ 9136\\ 9194\\ 9297\\ 9401\\ 9467\\ 9516\\
        9586\\ 9758\\ 9820\\ 9896\\ 9964\\ 10124\\ 10265\\ 10308\\
        10363\\ 10387\\ 10545\\ 10545\\ 10645\\ 10918\\ 11344\\ 11487\\
        11971\\ 12041\\ 12048\\ 12226\\ 12247\\ 12358\\ 12390\\ 13093\\
        13569\\ 13662\\ 13702\\ 13808\\ 13930\\ 14367\\ 14484\\ 14868\\
        15143\\ 15345\\ 16011\\ 16222\\ 16276\\ 16432\\ 16808\\ 18934\\
        19254\\ 20851\\ 21550\\ 23374\\ 24739\\ 27800\\ 27962\\ 34605\\
      };
      \addlegendentry{Layer 30}

      \addplot[thick, mark = none, color = black!70, dotted]
        table[x expr = \coordindex, y index = 0, row sep = \\, header = false] {
        83\\ 149\\ 206\\ 224\\ 260\\ 307\\ 331\\ 400\\
        517\\ 543\\ 611\\ 737\\ 768\\ 819\\ 1218\\ 1377\\
        1456\\ 1682\\ 1818\\ 2308\\ 2905\\ 3048\\ 3170\\ 3417\\
        3501\\ 3874\\ 3913\\ 3984\\ 4156\\ 4272\\ 4626\\ 4692\\
        4783\\ 5004\\ 5015\\ 5131\\ 5167\\ 5232\\ 5234\\ 5297\\
        5304\\ 5361\\ 5509\\ 5666\\ 5758\\ 5901\\ 6081\\ 6338\\
        6699\\ 6719\\ 6778\\ 6921\\ 6952\\ 7035\\ 7140\\ 7150\\
        7286\\ 7334\\ 7342\\ 7355\\ 7470\\ 7576\\ 7641\\ 7784\\
        7885\\ 8360\\ 8370\\ 8614\\ 8750\\ 8763\\ 8773\\ 8805\\
        9227\\ 9287\\ 9613\\ 9858\\ 10003\\ 10083\\ 10254\\ 10552\\
        10800\\ 10867\\ 10870\\ 11012\\ 11723\\ 12001\\ 12206\\ 12254\\
        12512\\ 12999\\ 13214\\ 13397\\ 13460\\ 13532\\ 13650\\ 13714\\
        13777\\ 13969\\ 13970\\ 14609\\ 15045\\ 15600\\ 15804\\ 16478\\
        18494\\ 18667\\ 18905\\ 19281\\ 19391\\ 19417\\ 20118\\ 22481\\
        22598\\ 22913\\ 23031\\ 23290\\ 23388\\ 23507\\ 28913\\ 29229\\
        30084\\ 32437\\ 32812\\ 35136\\ 42547\\ 61340\\ 65372\\ 82867\\
      };
      \addlegendentry{Layer 46}
    \end{axis}
  \end{tikzpicture}

  \caption{\modify{Degree distributions of the dominance graph ($\tau = 0.2$). (a) Share of
  the $2^{20}$ latents dominated by $0, \dots, 4$ experts (degree $4$ is at most $0.014\%$).
  (b) Number of latents each expert dominates, with the $128$ experts sorted in ascending order
  of degree.}}
  \label{fig:degree}
\end{figure}
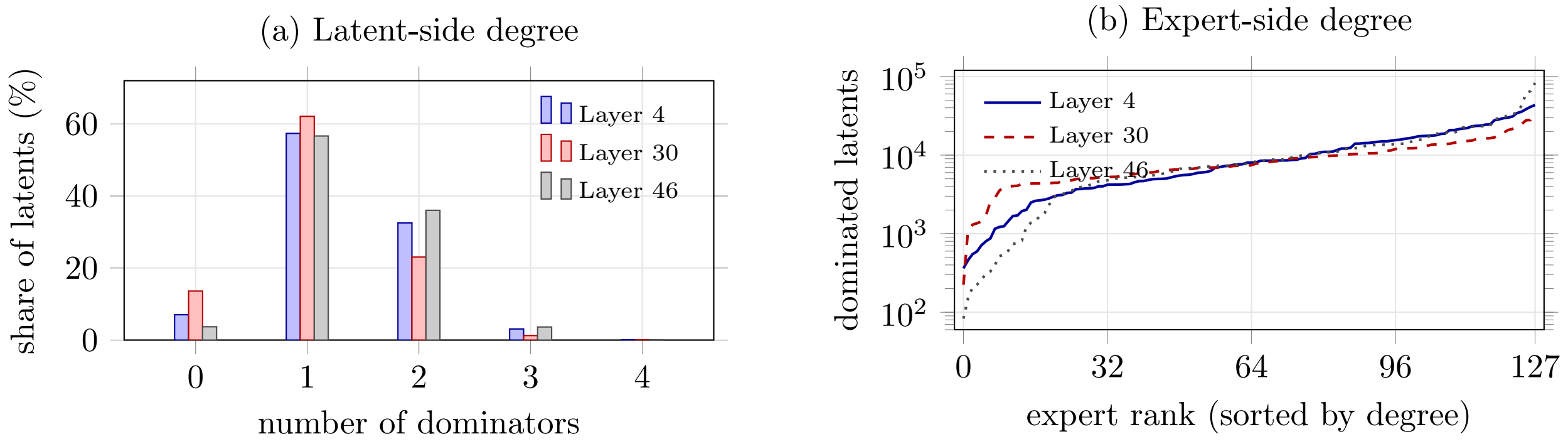

\modify{\Cref{fig:degree}(a) confirms both predictions.} In
all three layers over half of the $2^{20}$ latents have exactly one dominator, namely $57.4\%$
at layer 4, $62.1\%$ at layer 30 and $56.7\%$ at layer 46. \modify{Latents with one or two
dominators together account for $89.9\%$, $85.1\%$ and $92.7\%$ of all latents at layers 4, 30
and 46. At most $3.7\%$ of latents have three or more dominators, and none reaches the
theoretical maximum of five ($1/\tau$). Only $7.0\%$, $13.6\%$ and $3.7\%$ of latents have no
dominator because no single expert reaches $\tau = 0.2$. Most features are therefore driven by
one or two dedicated experts rather than shared across all routed experts.}

\subsection{\modify{From the expert side, dominant sets are nearly disjoint}}
\label{subsec:dominance-overlap}

\modify{We next examine the dominance graph from the expert perspective. Effective division of
labor requires two conditions on the expert sets $\mathcal{S}(j)$. First, dominance should
spread across all experts rather than collapsing onto a few hubs. \Cref{fig:degree}(b) shows
that all $128$ experts hold nonempty dominant sets in every layer (with minimum sizes of $362$,
$224$ and $83$ at three layers), and the ten largest sets hold at most $30.1\%$ of all
edges. Second, even when all experts participate, they could still duplicate one another on the
same features. Genuine specialization requires that different experts hold nearly disjoint
dominant sets. We test this second condition by computing the pairwise Jaccard similarity}
\(
  J(j, j') = \frac{\lvert \mathcal{S}(j) \cap \mathcal{S}(j') \rvert}
                  {\lvert \mathcal{S}(j) \cup \mathcal{S}(j') \rvert},
\)
\modify{over all $\binom{128}{2} = 8{,}128$ unordered expert pairs in each layer.}

\begin{figure}[t]
  \centering
  \includegraphics[width = 0.96\textwidth]{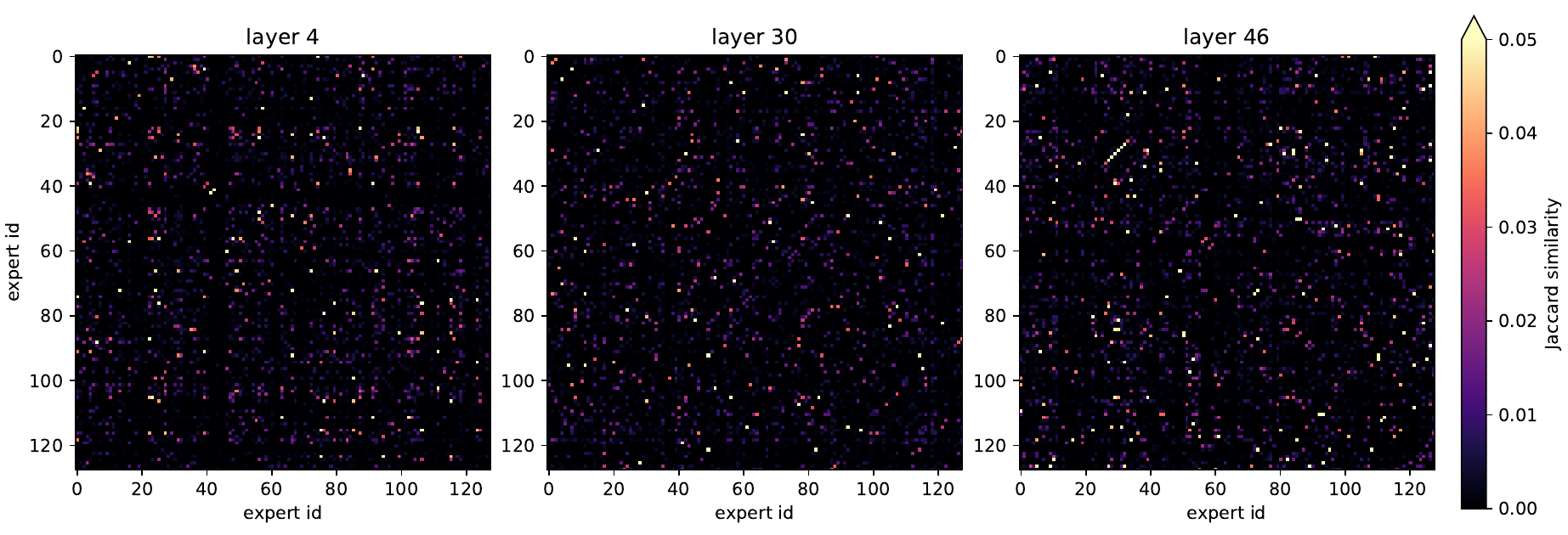}
  \caption{Pairwise Jaccard similarity of the expert dominant sets at $\tau = 0.2$, with the
  experts in their original index order on both axes. The three layers share one color scale,
  which saturates at $0.05$, and the diagonal is masked.}
  \label{fig:jaccard-heatmap}
\end{figure}

\modify{\Cref{fig:jaccard-heatmap} shows the full similarity matrices across all three layers,
with summary statistics reported in \Cref{app:jaccard-heatmap} (\Cref{tab:jaccard}). Each
matrix is a dark field with only a few isolated bright cells, showing neither blocks of
mutually overlapping experts nor bright rows from generalist experts.} The median of $J$ stays
near $10^{-4}$ in all three layers, and between $37.5\%$ and $40.5\%$ of the pairs share no
latent at all. \modify{Moreover, $46\%$ to $59\%$ of the latents in an expert's dominant set
are unique to that expert on average.}

\modify{The remaining overlap is strongly skewed across pairs. The mean of $J$ is an order of
magnitude above the median, and the top $10\%$ of pairs by $J$ account for $83.8\%$, $78.5\%$
and $81.0\%$ of all shared latents at layers 4, 30 and 46. At most $0.57\%$ of pairs exceed
$J = 0.05$, with maximum values of $0.138$, $0.098$ and $0.317$.}

The tail is organized into pairs rather than groups. \modify{If we link expert pairs with
$J > 0.05$, most involved experts connect to only one other expert, and no connected component
exceeds $10$ experts. Across all $128$ experts in layers 4, 30 and 46, moreover, $26$, $33$
and $30$ pairs are mutual nearest neighbors under $J$. Nor is one dominant set a subset of the
other. Even among pairs with $J > 0.05$, the shared latents cover a median of only $16\%$ to
$18\%$ of the smaller set. \Cref{subsec:within-domain-overlap} returns to these high-overlap
pairs once the experts are labeled.}

\section{\modify{How experts divide labor across domains and subfields}}
\label{sec:label}

\Cref{sec:qwen3specialized} establishes that every expert holds a dominant set of latents,
that about half of this set is dominated by that expert alone, and that substantial overlap
occurs only between a few pairs of experts. The set attached to an expert is therefore a
description of that expert, so interpreting the latents it contains tells us which features the
expert is chiefly responsible for. \Cref{sec:dominance_graph meta} and
\Cref{subsec:dominance-overlap} further show that the structure of dominance and its trends are
the same at layers 4, 30 and 46, and we therefore report only layer 30 in the rest of this
section.

\modify{Interpreting these dominant sets reveals a clear hierarchical organization. Across the
layer, experts cluster into coarse domains of strongly unequal sizes
(\Cref{subsec:domain-imbalance}). Within each coarse domain, individual experts specialize in
distinct fine-grained subfields (\Cref{subsec:fine-grained}). Meanwhile, substantial feature
overlap occurs almost exclusively between neighboring subfields inside the same domain
(\Cref{subsec:within-domain-overlap}).}

\subsection{\modify{Interpreting experts from their dominant sets}}
\label{subsec:wordcloud}

\begin{figure}[t]
  \centering
  \begin{minipage}[t]{0.3\textwidth}
    \centering
    \includegraphics[width = \textwidth]{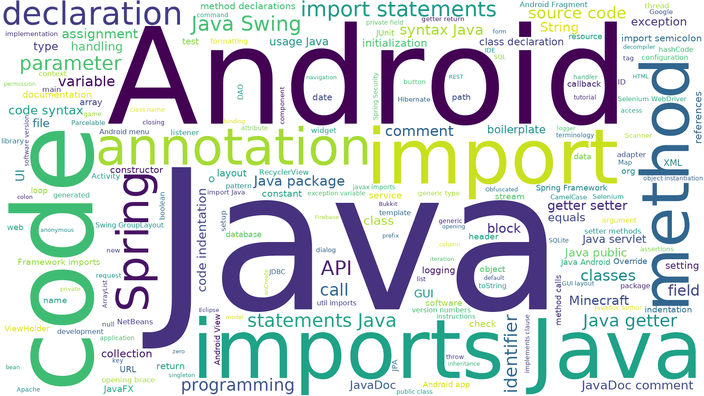}\\[0.3em]
    {\small (a) E50: Java and Android}
  \end{minipage}\hfill
  \begin{minipage}[t]{0.3\textwidth}
    \centering
    \includegraphics[width = \textwidth]{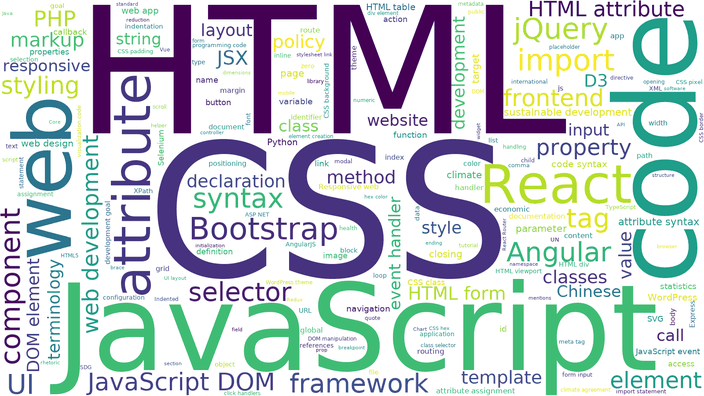}\\[0.3em]
    {\small (b) E57: frontend web}
  \end{minipage}\hfill
  \begin{minipage}[t]{0.3\textwidth}
    \centering
    \includegraphics[width = \textwidth]{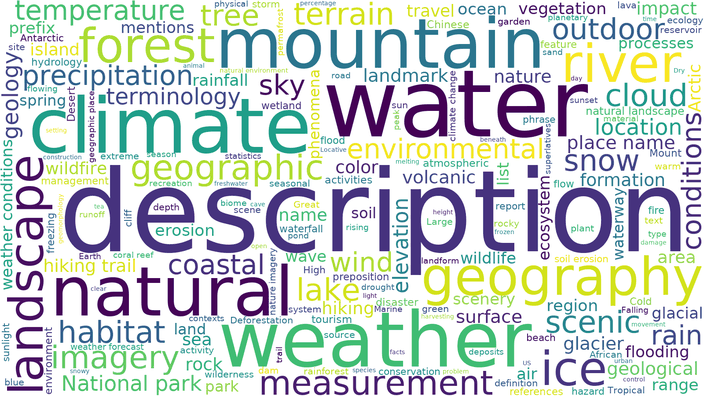}\\[0.3em]
    {\small (c) E93: geography and weather}
  \end{minipage}

  \caption{Explanation word clouds of three experts of layer 30. Word size is the frequency
  among the explanations of the latents sampled from the expert's dominant set.}
  \label{fig:wordcloud_sample}
\end{figure}

For every expert we sample 2000 latents uniformly from its dominant set, taking the whole set
when it holds fewer than 2000 latents. Each sampled latent is interpreted by GPT-5.5 from
activating contexts and logit effects, following the pipeline described in \Cref{app:autointerp}.
Reading the resulting explanations expert by expert, we find that for most experts they
concentrate on a few themes. To make this concentration visible at a glance, we render
the explanations of each expert as a word cloud, in which the size of a word is its frequency
among the explanations of that expert's dominant set. \Cref{fig:wordcloud_sample} shows the
explanation word clouds of experts 50, 57 and 93 of layer 30 (a Java expert, a front-end web
expert and an expert on natural geography and weather), and the word clouds of the remaining
experts are collected in \Cref{app:wordclouds}.

\modify{In \Cref{app:token-vs-feature}, we compare these explanation word clouds with word
clouds built directly from the tokens routed to an expert. Token clouds are dominated by
function words and miss themes whose features fire on context-dependent tokens such as
operators. Explanation word clouds avoid both problems because they retain only the features an
expert drives and summarize the surrounding context.}

\subsection{\modify{Coarse domains receive unequal numbers of experts}}
\label{subsec:domain-imbalance}

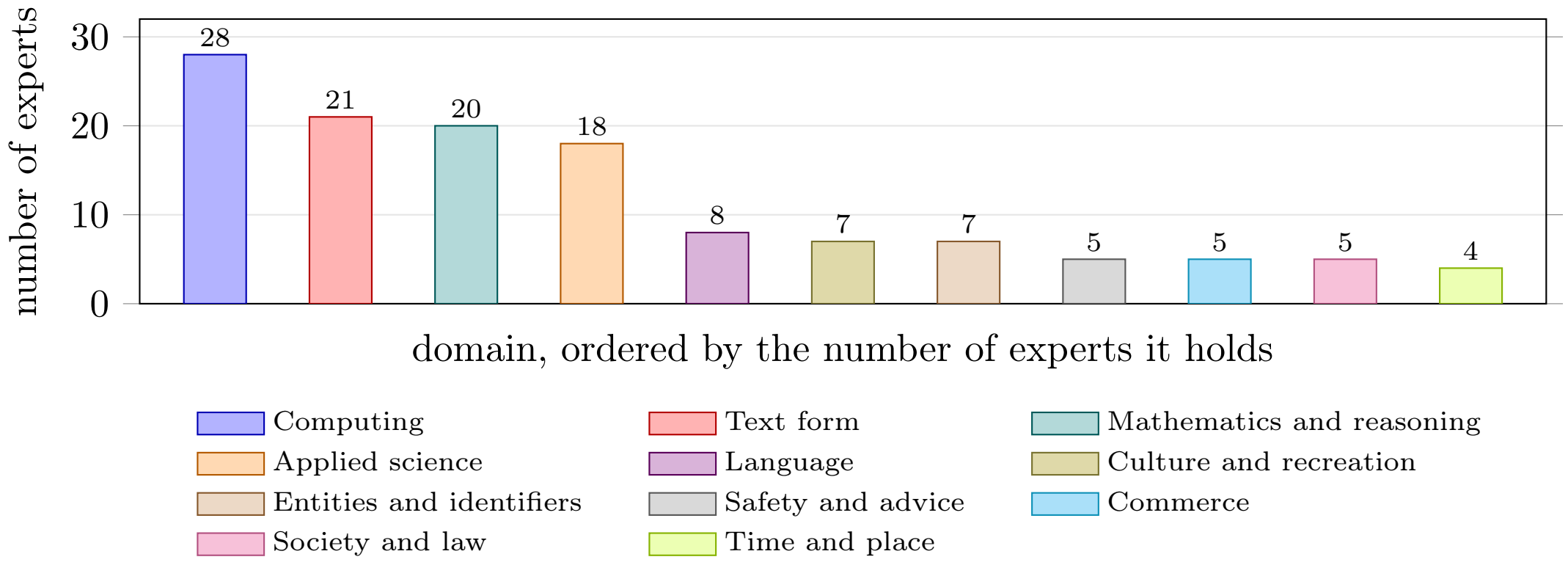
\begin{figure}[t]
  \centering
\begin{tikzpicture}
  \begin{axis}[
    width = 0.86\textwidth, height = 0.25\textwidth,
    ybar,
    area legend,
    bar width = 16pt,
    xlabel = {domain, ordered by the number of experts it holds},
    ylabel = {number of experts},
    xmin = 0.4, xmax = 11.6,
    ymin = 0, ymax = 32,
    xtick = \empty,
    grid = major, grid style = {gray!20},
    tick align = outside,
    nodes near coords,
    every node near coord/.append style = {font = \scriptsize, inner sep = 2pt},
    legend columns = 3,
    legend cell align = left,
    legend style = {
      at = {(0.5,-0.32)}, anchor = north, draw = none, font = \scriptsize,
      /tikz/every even column/.append style = {column sep = 1.5em},
    },
  ]
    \addplot[fill = blue!30, draw = blue!70!black, bar shift = 0pt] coordinates {(1,28)};
    \addlegendentry{Computing}
    \addplot[fill = red!30, draw = red!70!black, bar shift = 0pt] coordinates {(2,21)};
    \addlegendentry{Text form}
    \addplot[fill = teal!30, draw = teal!70!black, bar shift = 0pt] coordinates {(3,20)};
    \addlegendentry{Mathematics and reasoning}
    \addplot[fill = orange!30, draw = orange!70!black, bar shift = 0pt] coordinates {(4,18)};
    \addlegendentry{Applied science}
    \addplot[fill = violet!30, draw = violet!70!black, bar shift = 0pt] coordinates {(5,8)};
    \addlegendentry{Language}
    \addplot[fill = olive!30, draw = olive!70!black, bar shift = 0pt] coordinates {(6,7)};
    \addlegendentry{Culture and recreation}
    \addplot[fill = brown!30, draw = brown!70!black, bar shift = 0pt] coordinates {(7,7)};
    \addlegendentry{Entities and identifiers}
    \addplot[fill = gray!30, draw = gray!70!black, bar shift = 0pt] coordinates {(8,5)};
    \addlegendentry{Safety and advice}
    \addplot[fill = cyan!30, draw = cyan!70!black, bar shift = 0pt] coordinates {(9,5)};
    \addlegendentry{Commerce}
    \addplot[fill = magenta!30, draw = magenta!70!black, bar shift = 0pt] coordinates {(10,5)};
    \addlegendentry{Society and law}
    \addplot[fill = lime!30, draw = lime!70!black, bar shift = 0pt] coordinates {(11,4)};
    \addlegendentry{Time and place}
  \end{axis}
\end{tikzpicture}
  \caption{Number of experts in each of the eleven domains of layer 30. The domains and their
  names are produced by GPT-5.5 from the expert labels alone. \Cref{app:expert-labels}
  lists the experts of every domain.}
  \label{fig:expert-group-share}
\end{figure}

\modify{Inspecting word clouds one at a time reveals individual roles, but does not show how
all $128$ experts cover the feature space as a whole. To obtain a layer-wide view without
fixing categories in advance, we summarize each expert's word cloud into a short label and
group those labels into coarse domains.}

The procedure has two steps, carried out by GPT-5.5. We first name each expert. We hand the
model the words of one expert in the word cloud together with their weights and ask for a short
label naming what that expert is responsible for, with the prompt in
\Cref{app:expert-label-prompt}. \modify{For instance, expert 50 receives ``Java and Android
programming'', expert 93 receives ``physical geography and weather'', and expert 72 receives
``coding algorithms and musculoskeletal fitness'', recovering both themes shown in
\Cref{app:token-vs-feature}.} We then group the experts. A single call receives all 128 labels
at once and returns a partition of experts together with a name for each part, with the prompt
in \Cref{app:expert-group-prompt}. We call these parts domains.

Eleven domains come out of the 128 experts of layer 30, and \Cref{fig:expert-group-share} gives
the number of experts each of them holds. \Cref{app:expert-labels} lists every expert with its
label and its domain.

The allocation across domains is strongly unequal. Computing takes $28$ of the 128 experts and
mathematics and reasoning takes $20$, so these two domains alone hold $37.5\%$ of the layer.
\modify{Most labels in Computing concern code, while the rest cover areas such as cybersecurity
and software platforms. Mathematics and reasoning likewise extends to statistics, parts of the
physical sciences and problem-solving reasoning. Conversely, several categories common in human
taxonomies receive almost no dedicated capacity. Art appears in only one of the 128 labels
(expert 54, paired with animals under Culture and recreation).}

Three of eleven domains (text form, language, and entities and identifiers) are not subject
areas, and together they take $36$ experts, or $28.1\%$ of the layer. The largest of them, text
form with $21$ experts, is the second largest domain of the layer and covers document structure,
punctuation, delimiters and encoding artifacts. \modify{All three emerge directly from the
expert labels without being specified in the prompt. A predefined topical taxonomy would
scatter these $36$ experts across unrelated subject categories and miss all three form-based
groups.}

\modify{This imbalance among topical domains likely reflects the pretraining data mixture,
which typically overweights code and mathematics relative to general text.} Llama 3, for
instance, reports a final data mix of roughly $25\%$ mathematical and reasoning tokens and
$17\%$ code tokens against $50\%$ of general
knowledge~\citep{grattafiori2024llama}.\footnote{The Qwen3 technical report does not disclose
its data mixture, so we cannot check this on the model we study. Raising the proportion of
code, mathematics and reasoning data is nonetheless common practice in current pretraining
pipelines.} On this reading the allocation of experts across domains reflects the proportions
of the corresponding domains in the training data. \modify{Domains with a larger share of the
data receive more experts, allowing the model to draw finer distinctions within them.}

\subsection{\modify{Experts specialize in subfields within coarse domains}}
\label{subsec:fine-grained}

\modify{Why does a single coarse domain such as Computing hold $28$ experts, or Mathematics and
reasoning hold $20$? Looking inside each domain shows that its experts do not duplicate the
same broad role. Instead, individual experts divide a coarse domain into distinct, recognizable
subfields.}

Consider Computing, for instance. \Cref{fig:wordcloud_sample}(a) and (b) show the explanation
word clouds of experts 50 and 57, which both work on code. \modify{Expert 50 specializes in
Java. Its largest words are \emph{Java}, \emph{Android} and \emph{imports}, backed by latents
such as ``Java import statements'', ``JDBC generated keys'' and ``ButterKnife Android
imports''. Expert 57 specializes in front-end web development, with prominent words
\emph{HTML}, \emph{CSS} and \emph{JavaScript} alongside \emph{React}, \emph{Bootstrap} and
\emph{jQuery}, driven by latents on DOM events, CSS selectors and responsive layout. Beyond
these two, Computing also contains dedicated experts for competitive programming (experts 9, 40
and 118), system administration (expert 51) and Python API documentation (expert 113), each
covering a different aspect of working with code.}

The same holds inside Mathematics and reasoning. \Cref{app:wordclouds} gives experts 1 and 76.
Expert 1 is a geometry expert, dominated by \emph{geometry}, \emph{coordinate}, \emph{triangle}
and \emph{angle}. Expert 76 instead concentrates on arithmetic and polynomial computation, with
\emph{polynomial}, \emph{matrix}, \emph{division} and \emph{decimal} in the largest font, over
latents such as ``Polynomial long division'' and ``arbitrary-precision integers''.

\modify{To quantify this separation,} we use GPT-5.5 to judge whether each latent explanation
of an expert belongs to the corresponding theme, with the prompt reproduced in
\Cref{app:theme-check-prompt}. \Cref{tab:fine-grained} reports the share of each expert's
explanations falling in the corresponding theme among the latents sampled from its dominant
set. For example, $86.7\%$ of the latent explanations in expert 50's dominant set relate to
Java development. Against a random sample of $2000$ latents drawn from the dominant sets of the
other $127$ experts, the share of the same theme falls to between $1.9\%$ and $6.7\%$.

\begin{table}[t]
  \centering
  \small
  \caption{Share of latent explanations judged by GPT-5.5 to belong to each theme at layer 30,
  among the latents sampled from the expert's dominant set (Share) and among $2000$ latents
  drawn uniformly from the pooled samples of the other $127$ experts (Random).}
  \begin{tabular}{lcccc}
    \hline
    Expert & 50 & 57 & 1 & 76 \\
    \hline
    Theme & Java development & Front-end development & Geometry & Arithmetic and polynomial \\
    Share  & $86.7\%$ & $78.6\%$ & $78.1\%$ & $59.1\%$ \\
    Random & $1.9\%$ & $3.5\%$ & $2.0\%$ & $6.7\%$ \\
    \hline
  \end{tabular}
  \label{tab:fine-grained}
\end{table}

\subsection{\modify{Dominant-set overlap concentrates within coarse domains}}
\label{subsec:within-domain-overlap}

\begin{figure}[t]
  \centering
  \begin{minipage}[c]{0.52\textwidth}
    \centering
    \includegraphics[width = \textwidth]{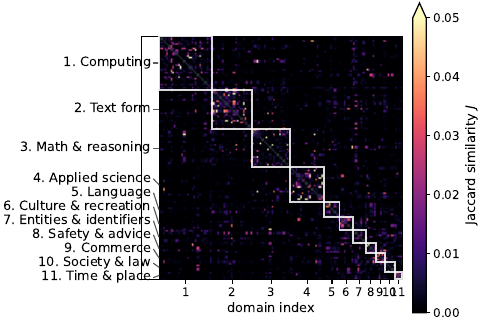}
    \caption{\modify{Pairwise Jaccard similarity $J$ of dominant sets at layer 30
    ($\tau = 0.2$), ordered by coarse domain on both axes. White boxes mark the eleven domains,
    the diagonal is masked, and the color scale saturates at $J = 0.05$.}}
    \label{fig:jaccard-domain-grouped}
  \end{minipage}\hfill
  \begin{minipage}[c]{0.45\textwidth}
    \centering
    \small
    \makeatletter\def\@captype{table}\makeatother
    \caption{\modify{Share of expert pairs at layer 30 belonging to the same coarse domain
    across bins of Jaccard similarity $J$. Across all $8{,}128$ pairs, $12.8\%$ share a
    domain.}}
    \label{tab:overlap-by-domain}
    \vspace{0.15cm}
    \begin{tabular}{lrr}
      \hline
      $J$ & Pairs & Same domain \\
      \hline
      $0$                   & $3{,}290$ & $5.5\%$  \\
      $(0, 10^{-4}]$        & $769$     & $7.8\%$  \\
      $(10^{-4}, 10^{-3}]$  & $2{,}228$ & $13.1\%$ \\
      $(10^{-3}, 10^{-2}]$  & $1{,}544$ & $23.4\%$ \\
      $(10^{-2}, 0.05]$     & $277$     & $44.4\%$ \\
      $> 0.05$              & $20$      & $95.0\%$ \\
      \hline
    \end{tabular}
  \end{minipage}
  \vspace{-0.5cm}
\end{figure}

\modify{These subfields do not form a strictly disjoint partition. Although dominant sets share
almost nothing on average, \Cref{subsec:dominance-overlap} showed that a small tail of expert
pairs shares a substantial number of latents. Using the coarse domains from
\Cref{subsec:domain-imbalance}, we examine how this overlap relates to domain boundaries.}

\modify{We reorder the layer 30 Jaccard matrix of \Cref{fig:jaccard-heatmap} so that experts in
the same coarse domain sit adjacent along both axes. As \Cref{fig:jaccard-domain-grouped}
shows, bright high-similarity entries lie predominantly inside the diagonal domain blocks,
whereas off-block entries are sparse and weak. Across all $8{,}128$ pairs, the mean Jaccard
similarity is $5.2 \times 10^{-3}$ within the same domain, roughly five times the cross-domain
mean of $1.0 \times 10^{-3}$.}

\modify{\Cref{tab:overlap-by-domain} quantifies this pattern across six bins of $J$. As $J$
increases, the share of pairs belonging to the same coarse domain rises monotonically from
$5.5\%$ at $J = 0$ (below the layer-wide baseline of $12.8\%$) to $44.4\%$ for
$J \in (10^{-2}, 0.05]$ and $95.0\%$ ($19$ of $20$ pairs) for $J > 0.05$. Without binning, a
randomly drawn same-domain pair has a larger $J$ than a randomly drawn cross-domain pair with
probability $0.712$.}

\modify{Inspecting the $20$ pairs with $J > 0.05$ confirms that strong overlap connects
neighboring subfields inside the same domain. Within Text form, experts 83 and 105 both handle
punctuation ($J = 0.098$), and experts 77 and 107 both handle Unicode mojibake ($J = 0.092$).
Within Computing, experts 57 and 69 both cover JavaScript web programming ($J = 0.051$). The
single cross-domain pair above $J = 0.05$ links expert 11 (LaTeX algebraic expressions in
Mathematics and reasoning) with expert 95 (chemical and algebraic notation in Applied science),
which likewise share symbolic notation. Even among these $20$ pairs, shared latents account for
a median of only $12.3\%$ of either dominant set, so each expert's label still reflects mainly
its unshared features.}

\section{Conclusion}
\label{sec: conclusion}

\modify{We have introduced a general framework for analyzing expert specialization in the
feature space of an MoE layer rather than in a preset domain taxonomy.} A layer output is a
router-weighted sum of the outputs of the activated experts and the SAE encoder is affine in
that output, so the pre-activation of every latent decomposes exactly over the experts that
produced it. \modify{Tracing latents through this decomposition yields each expert's dominant
set, the latents it primarily drives. Interpreting this set describes the expert in the model's
own features rather than in predefined categories.}

\modify{Applying this framework to layers 4, 30 and 46 of Qwen3-30B-A3B yields a unified
picture of expert specialization. First, at the feature level, expert contributions are far
steeper than router gating weights, and most latents are dominated by only one or two experts.
Dominant sets are largely disjoint across experts. About half of the latents in an expert's
dominant set are unique to that expert, and most expert pairs share almost no latents. Second,
auto-interpreting layer 30 reveals a hierarchical division of labor. (a) Grouping the
interpreted roles recovers eleven coarse domains of strongly unequal sizes, led by computing,
text form, and mathematics and reasoning. (b) Within each coarse domain, individual experts
specialize in distinct subfields, well below conventional domain labels. (c) These subfields
are not a hard partition, as meaningful feature overlap concentrates within coarse domains and
connects neighboring subfields.}

\modify{Because our method requires only the expert outputs and an SAE on the layer output, it
applies to any MoE layer that linearly combines expert outputs. It can therefore also compare
how different routing objectives and load-balancing losses shape expert specialization. Finally,
dominant sets suggest a testable hypothesis for expert pruning, merging and feature-level
steering. Removing or editing an expert should primarily affect the latents it dominates.} We
discuss our limitations in \Cref{sec: limitations}.


\clearpage

\bibliography{iclr2027_conference}
\bibliographystyle{iclr2027_conference}

\clearpage

\beginappendix






\section{Limitations and future work}
\label{sec: limitations}

\begin{itemize}
  \item Our analysis inherits the imperfections of the SAE. The reconstruction of the layer
  output carries an error, and the latents are not a clean basis, since one feature can be
  split across several latents and can be absorbed into another~\citep{chanin2026absorption}. A better decomposition of an
  MLP into features would sharpen the picture of an MoE layer that we draw here.

  \item The pretraining mixture of Qwen3 is not disclosed, and our statistics are computed on a
  corpus of our own choosing. What we measure is therefore routing on our data rather than on
  the data the model was trained on, and the reading of \Cref{subsec:domain-imbalance} that
  relates the allocation of experts to the proportions of domains in the training data remains
  a conjecture.

  \item We report a single model. Extending the analysis to more and to larger MoE models would
  show which of these observations are general, and the same instrument can be used to compare
  how different load balancing losses shape the division of labor among experts.
\end{itemize}

\section{AI use statement}
\label{app:llm-use}

In this work, we used generative AI tools for implementing code, analyzing data, interpreting
experimental results, and polishing the writing. Separately, GPT-5.5 is a component of our method, used to explain latents, judge theme membership, label experts and group them into domains, with every prompt
reproduced in \Cref{app:autointerp,app:theme-check-prompt,app:expert-label-prompt,app:expert-group-prompt}. We have reviewed all AI-assisted work. We take responsibility for the final content
of this work, including text, claims or artifacts produced with the aid of generative AI.

\section{\modify{Summary statistics of dominant-set overlap}}
\label{app:jaccard-heatmap}

\modify{\Cref{tab:jaccard} reports summary statistics of the pairwise Jaccard similarity $J$
across all $8{,}128$ expert pairs in layers 4, 30 and 46, complementing the heatmaps in
\Cref{fig:jaccard-heatmap}.}

\begin{table}[t]
  \centering
  \small
  \caption{Pairwise Jaccard similarity over the $8{,}128$ expert pairs of each layer. The
  intersection in the top $10\%$ of pairs is the share of the total intersection carried by the
  $813$ pairs with the largest $J$. The lower block describes the pairs with
  $J > 0.05$\modify{. Here, a} single partner means exactly one other expert above this
  threshold; a connected group is a connected component of the graph formed by these pairs.}
  \begin{tabular}{lccc}
    \hline
                                       & Layer 4               & Layer 30              & Layer 46              \\
    \hline
    Mean $J$                           & $1.73 \times 10^{-3}$ & $1.57 \times 10^{-3}$ & $2.12 \times 10^{-3}$ \\
    Median $J$                         & $1.06 \times 10^{-4}$ & $1.01 \times 10^{-4}$ & $1.37 \times 10^{-4}$ \\
    Pairs sharing no latent            & $37.8\%$              & $40.5\%$              & $37.5\%$              \\
    Intersection in top $10\%$ of pairs & $83.8\%$             & $78.5\%$              & $81.0\%$              \\
    \hline
    Pairs with $J > 0.05$              & $25$ ($0.31\%$)       & $20$ ($0.25\%$)       & $46$ ($0.57\%$)       \\
    Experts involved (out of $128$)    & $35$                  & $34$                  & $60$                  \\
    \quad with a single partner        & $22$                  & $28$                  & $36$                  \\
    Largest connected group of experts & $10$                  & $4$                   & $8$                   \\
    Largest $J$                        & $0.138$               & $0.098$               & $0.317$               \\
    \hline
  \end{tabular}
  \label{tab:jaccard}
\end{table}

\section{\modify{Comparison with routed-token word clouds}}
\label{app:token-vs-feature}

\begin{figure}[t]
  \centering
  \begin{minipage}[t]{0.3\textwidth}
    \centering
    \includegraphics[width = \textwidth]{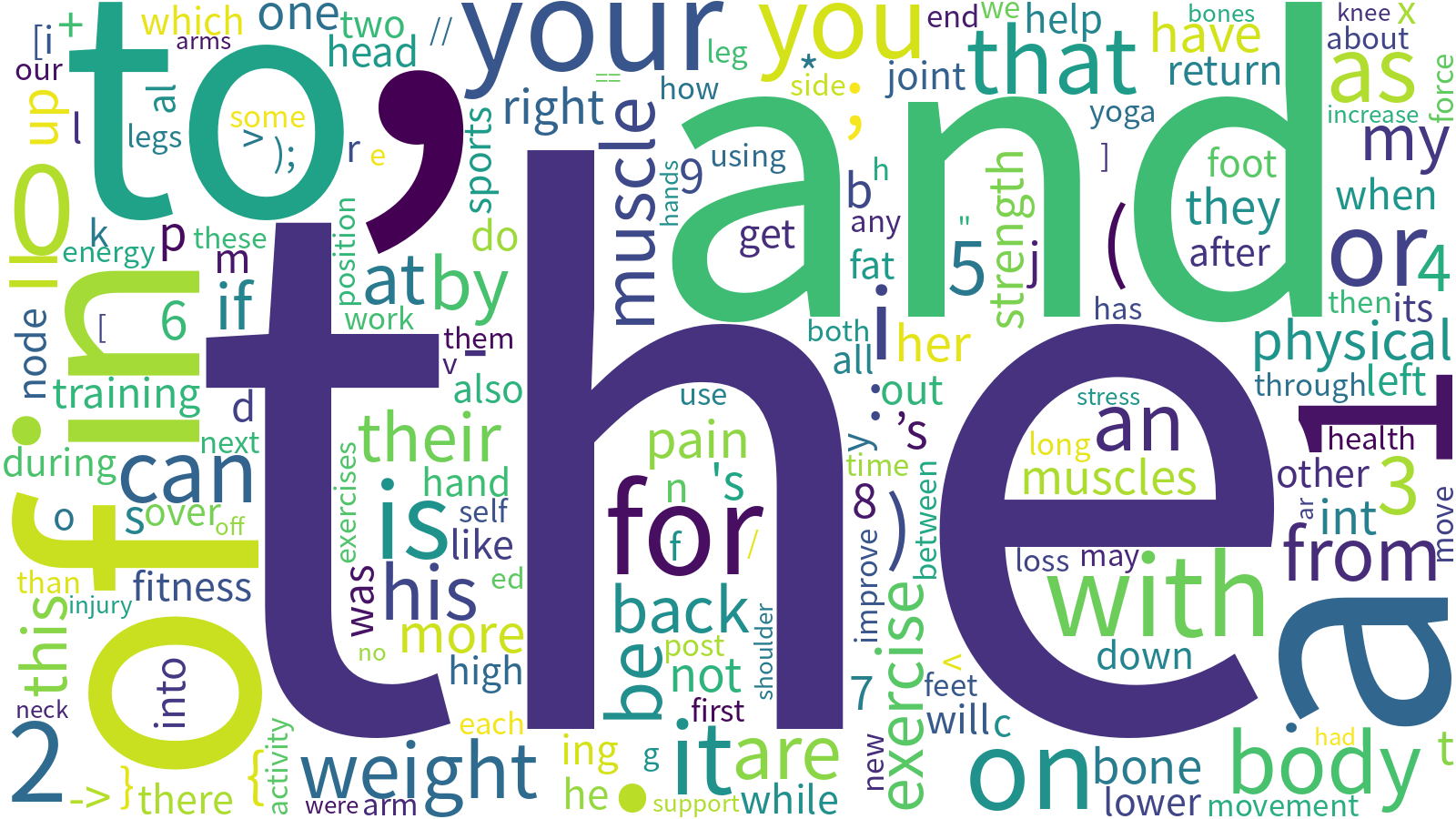}\\[0.3em]
    {\small (a) routed tokens}
  \end{minipage}\hfill
  \begin{minipage}[t]{0.3\textwidth}
    \centering
    \includegraphics[width = \textwidth]{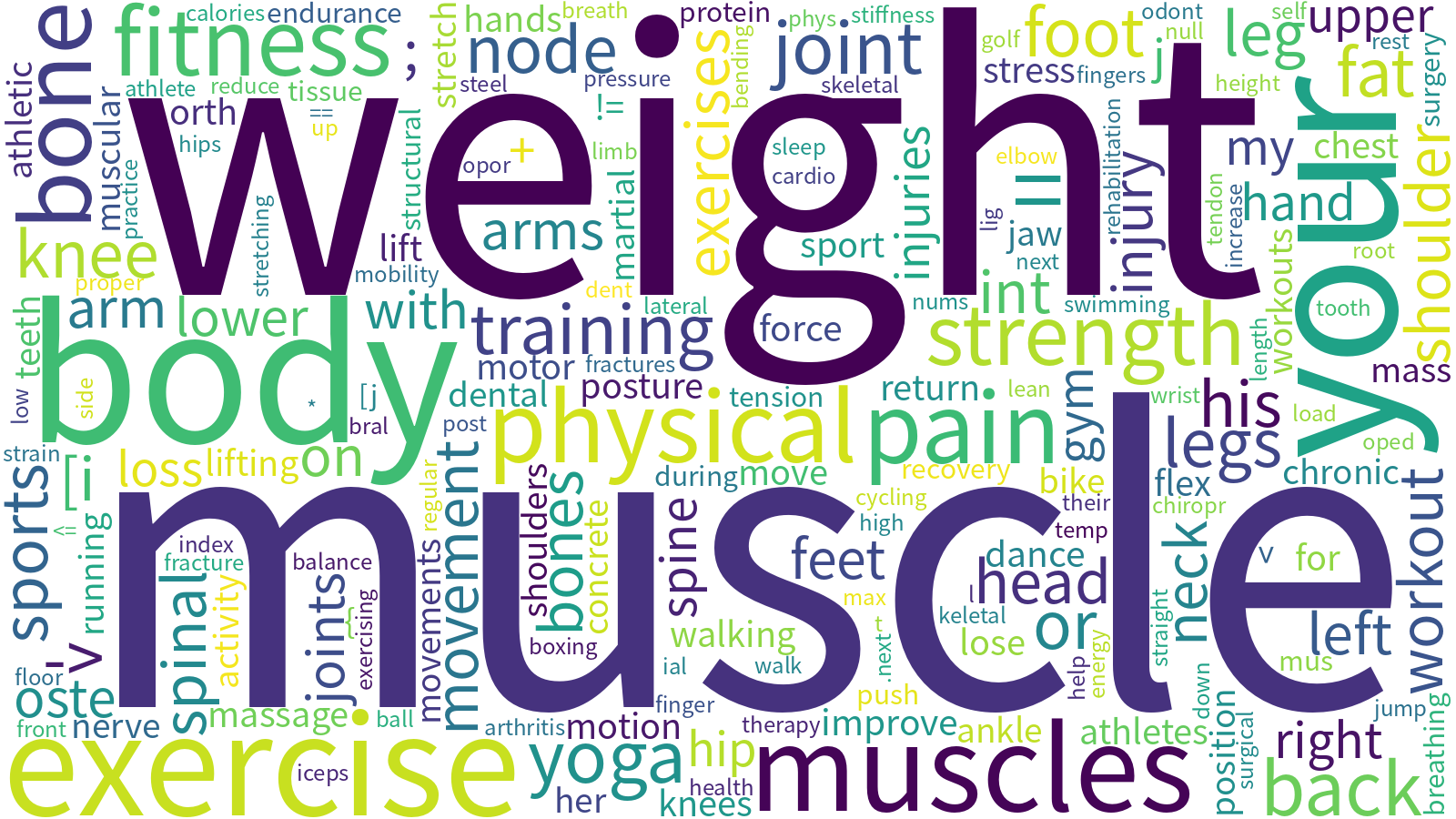}\\[0.3em]
    {\small (b) routed tokens, MI reweighted}
  \end{minipage}\hfill
  \begin{minipage}[t]{0.3\textwidth}
    \centering
    \includegraphics[width = \textwidth]{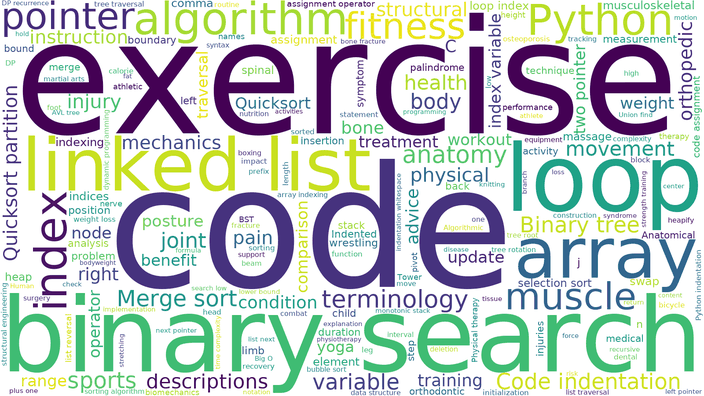}\\[0.3em]
    {\small (c) latent explanations}
  \end{minipage}

  \caption{Word clouds of expert 72 of layer 30, built from (a) routed tokens weighted by
  gating mass, (b) the same tokens reweighted by their contribution to mutual information
  without the top five function words, and (c) latent explanations. The algorithm theme of (c)
  is barely readable in (a) and (b).}
  \label{fig:token-vs-feature}
\end{figure}

\modify{In \Cref{subsec:wordcloud}, we summarize each expert using a word cloud built from the
explanations of latents in its dominant set.} We compare this construction with a word cloud
drawn directly from the tokens routed to an expert. Ours differs in two respects.
\modify{First, the dominant set retains only the latents an expert primarily drives, filtering
out unrelated material. Second, each explanation summarizes the surrounding context rather than
the activating token alone.} \Cref{fig:token-vs-feature}(a) gives the token word cloud of
expert 72 of layer 30, in which a word is weighted by the gating mass the expert accumulates on
it over the corpus. The words it puts forward are \emph{the}, \emph{and}, \emph{of} and other
function words. Such function tokens do carry varied and rich
information~\citep{zhang2025memory}, but that information resides in the features activated at
their positions and cannot be read off the tokens themselves, so at the token level they
contribute nothing but noise.

We remove the five most frequent function
words\footnote{\emph{the}, \emph{of}, \emph{to}, \emph{and}, \emph{a}} and reweight each
remaining token by its contribution to the mutual information between the expert and the token.
\Cref{fig:token-vs-feature}(b) is the result. It is more informative than
\Cref{fig:token-vs-feature}(a), yet part of what the expert does is still lost.
\modify{By contrast, the explanation word cloud in \Cref{fig:token-vs-feature}(c) shows that
expert 72 covers two distinct themes, namely exercise and health alongside algorithms and data
structures.} The second theme is barely readable in \Cref{fig:token-vs-feature}(b), because the
latents that carry it activate only at particular positions within an algorithm. \modify{Latent
F97783, for instance, fires only at the \texttt{<=} operator of a binary search, and only its
explanation (``Binary search loop condition'') connects this position to binary search.} On a
token word cloud this latent can contribute no more than the symbol \texttt{<=}, whose origin
and meaning are left open.

\section{Word clouds}
\label{app:wordclouds}

\Cref{fig:wordcloud-all-1} to \Cref{fig:wordcloud-all-8} give the explanation word clouds of all
128 experts of layer 30, each built as in \Cref{fig:wordcloud_sample}.

\newcommand{\wcpanel}[1]{%
  \begin{minipage}[t]{0.3\textwidth}
    \centering
    \includegraphics[width = \textwidth]{figures/wordcloud/#1f.png}\\[0.3em]
    {\small Expert #1}
  \end{minipage}%
}
\newcommand{\wcrow}[3]{%
  \wcpanel{#1}\hfill\wcpanel{#2}\hfill\wcpanel{#3}\par\vspace{0.8em}%
}

\begin{figure}[!h]
  \centering
  \wcpanel{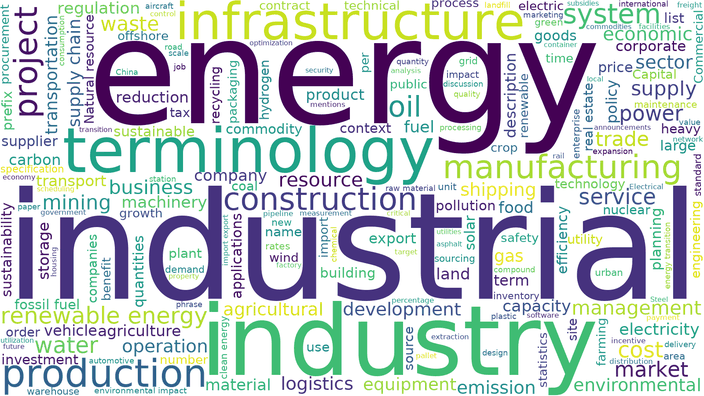}\hspace{0.02\textwidth}\wcpanel{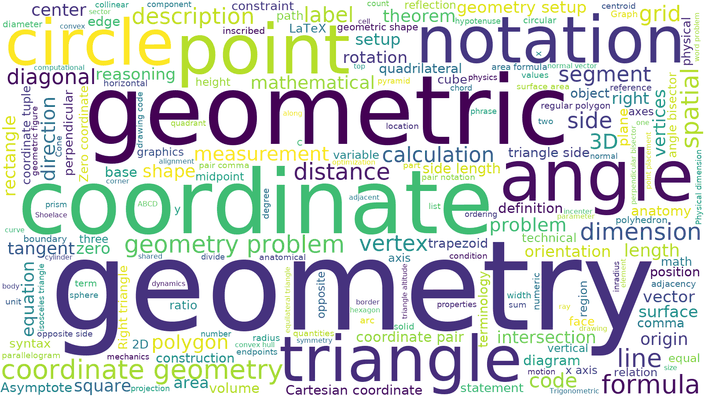}
  \small
  \caption{Explanation word clouds of experts 0 and 1 of layer 30. Each cloud is built from the
  explanations of the latents sampled from that expert's dominant set, and the size of a word
  is its frequency among those explanations.}
  \label{fig:wordcloud-all-1}
\end{figure}

\begin{figure}[p]
  \centering
  \wcrow{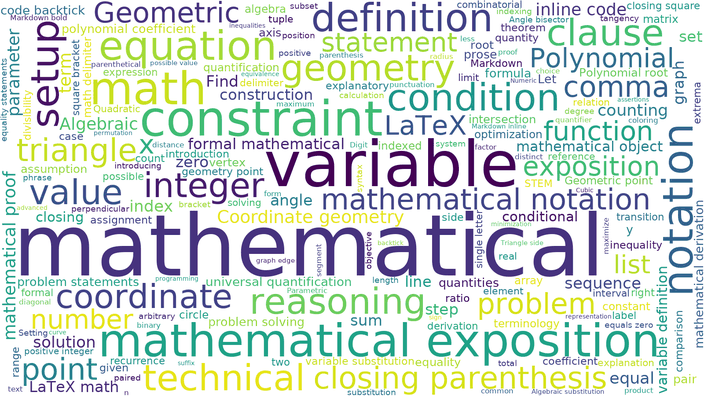}{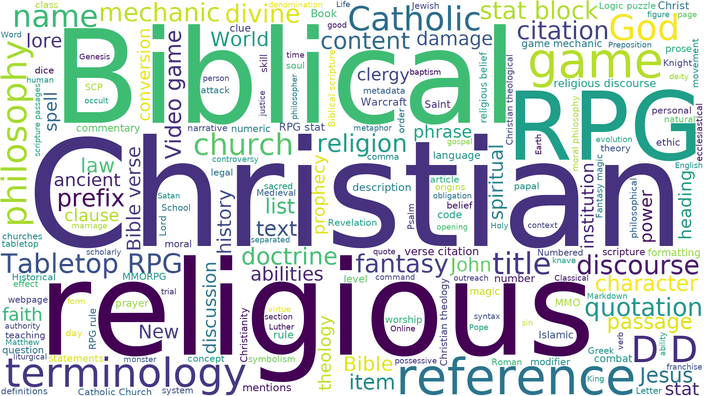}{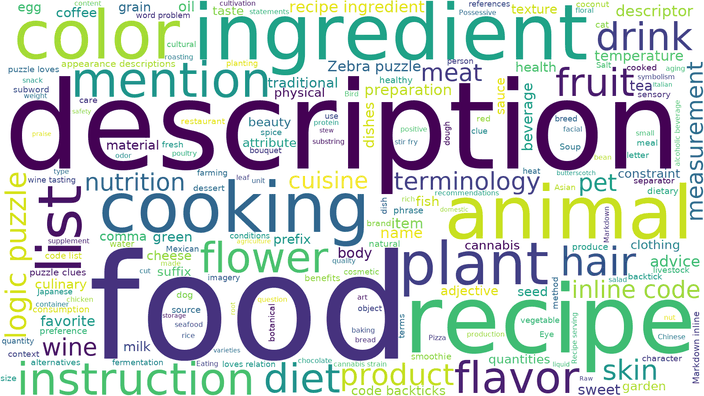}
  \wcrow{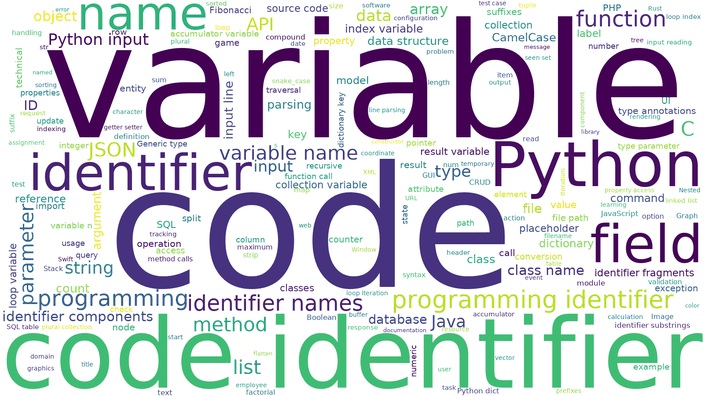}{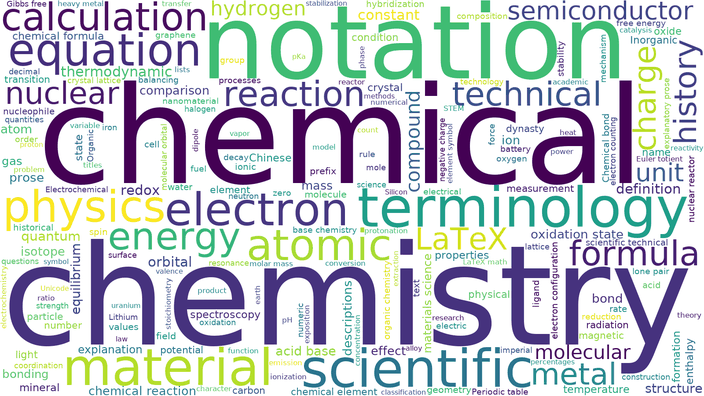}{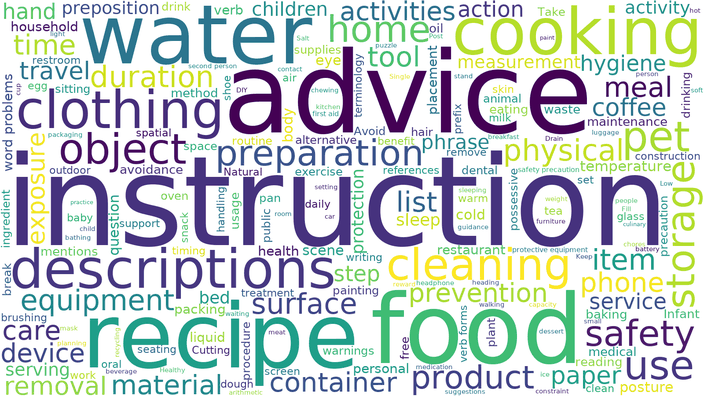}
  \wcrow{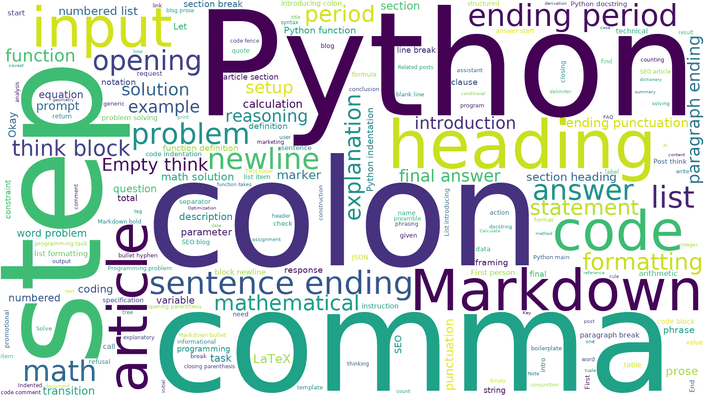}{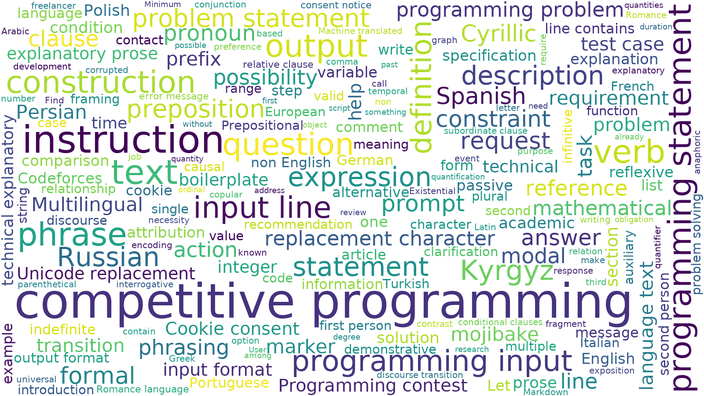}{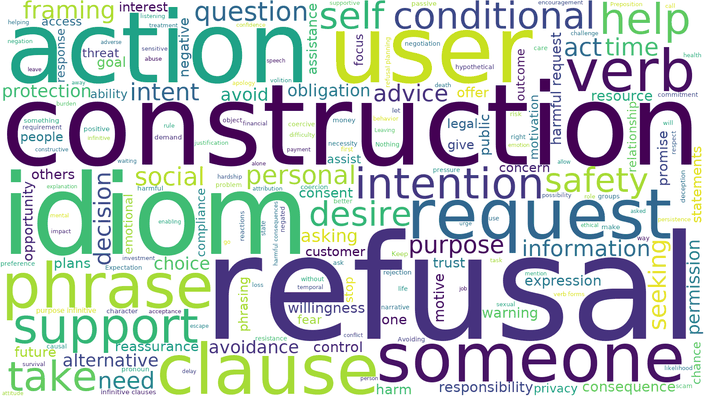}
  \wcrow{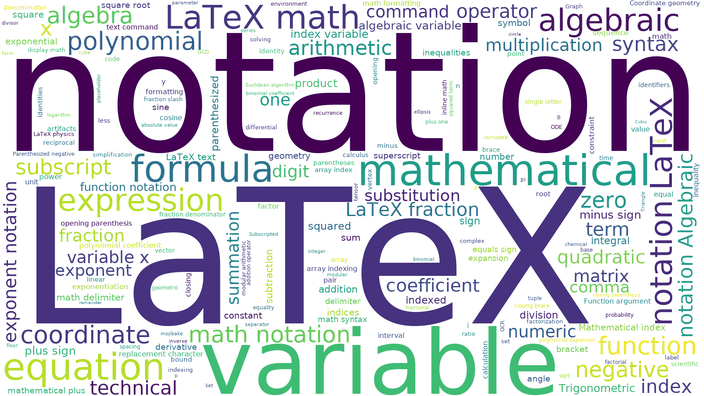}{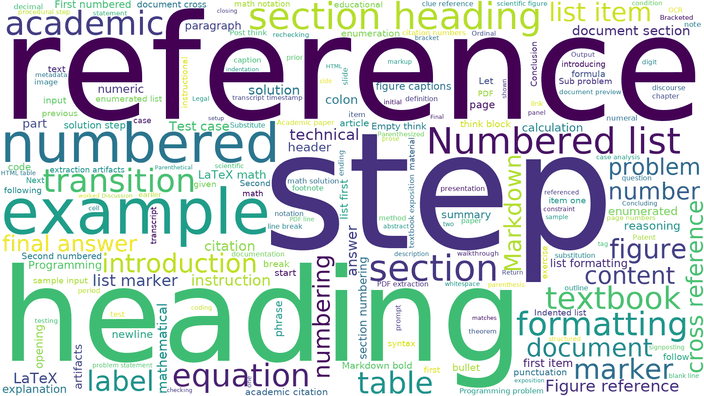}{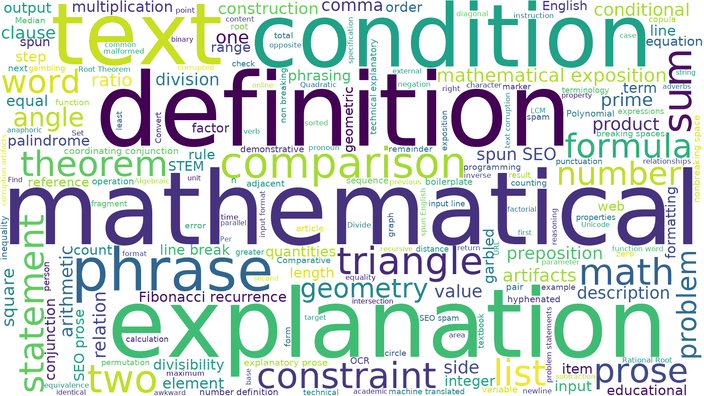}
  \wcrow{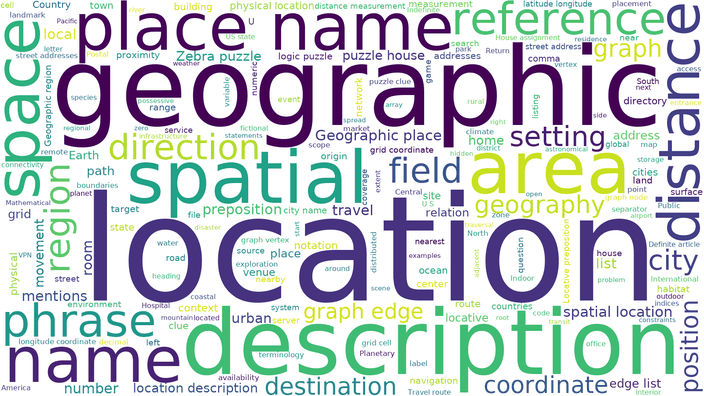}{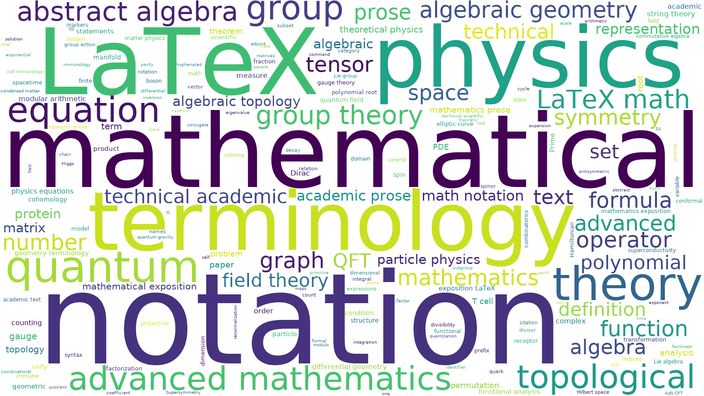}{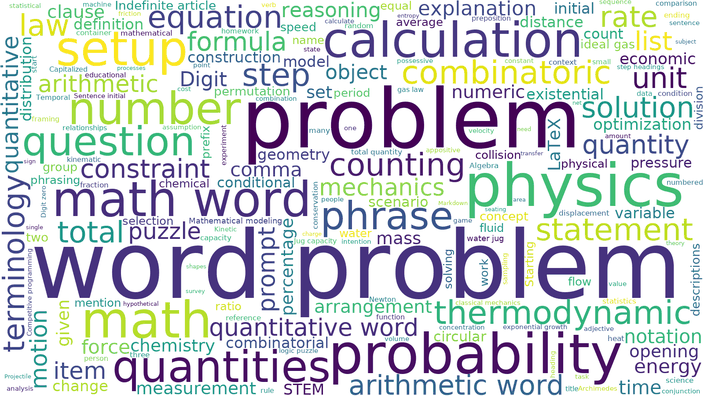}
  \wcrow{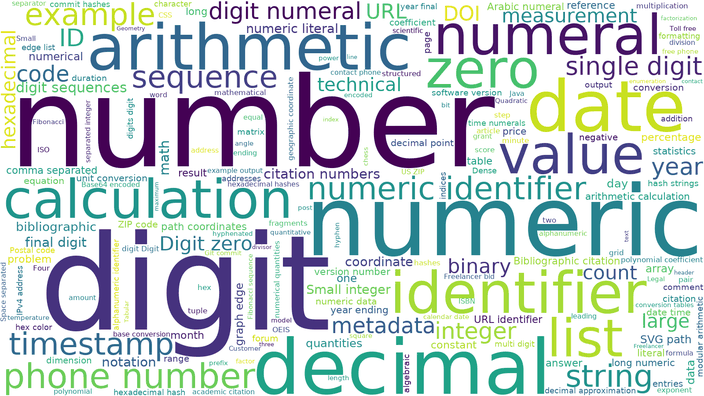}{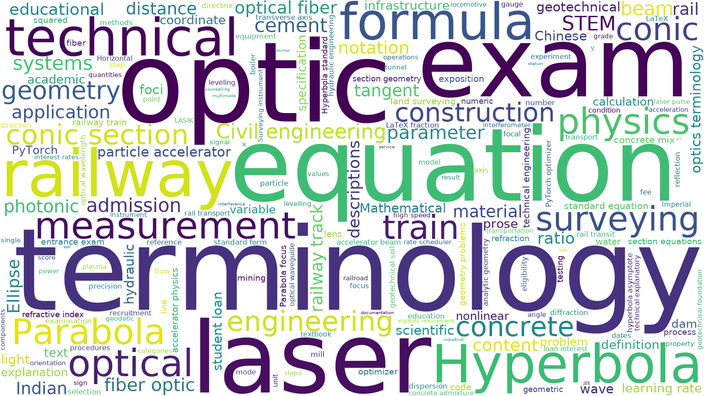}{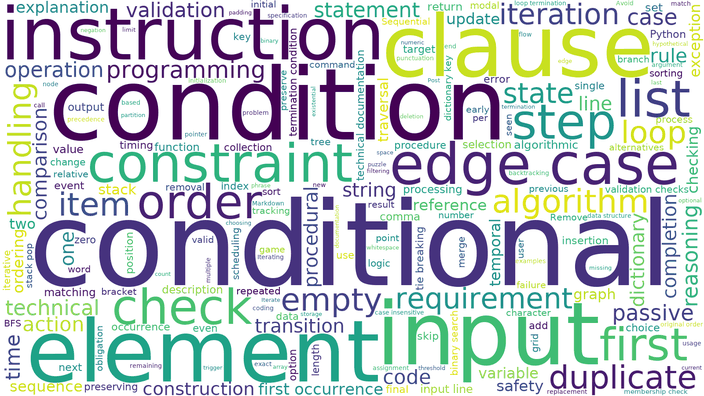}

  \caption{Explanation word clouds of experts 2 to 19 of layer 30, continued from
  \Cref{fig:wordcloud-all-1}.}
  \label{fig:wordcloud-all-2}
\end{figure}

\begin{figure}[p]
  \centering
  \wcrow{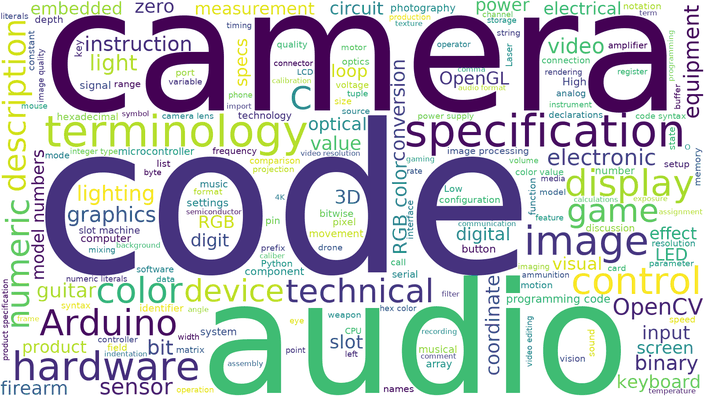}{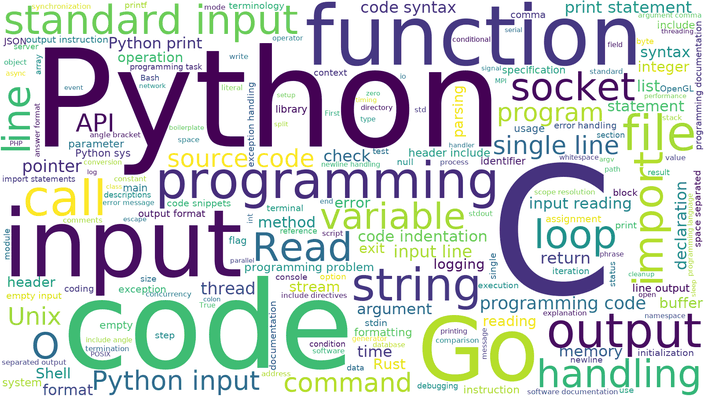}{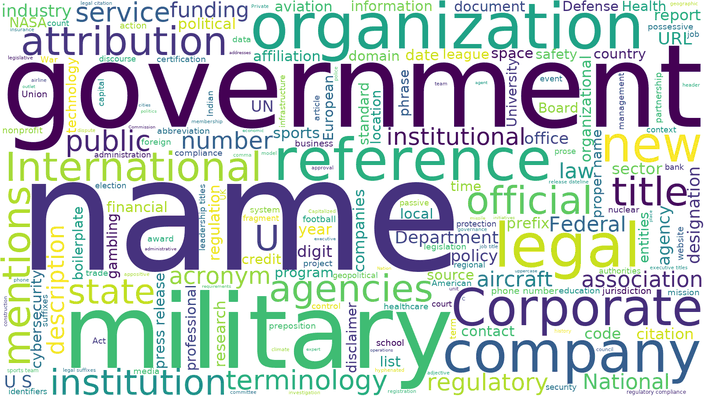}
  \wcrow{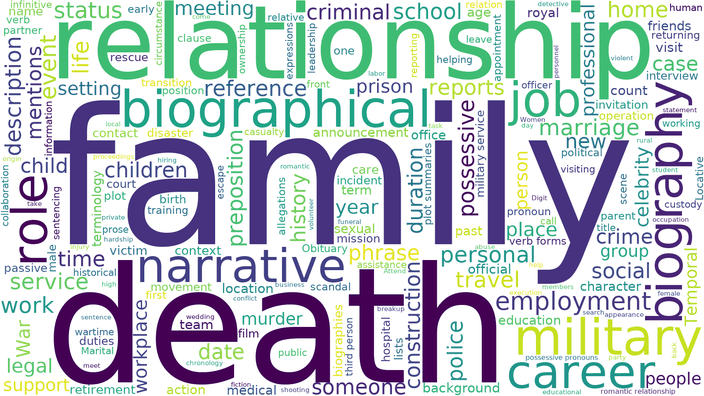}{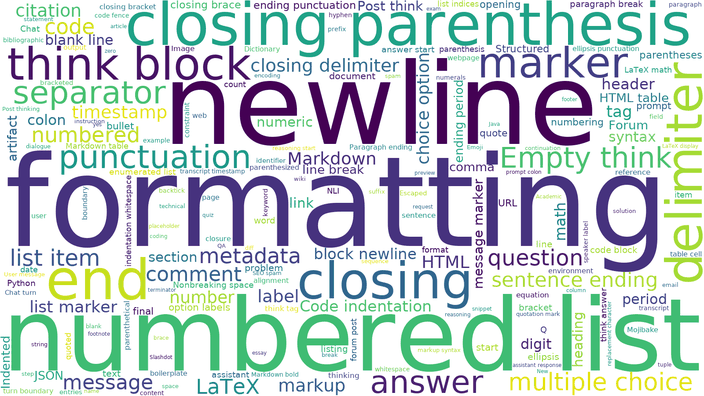}{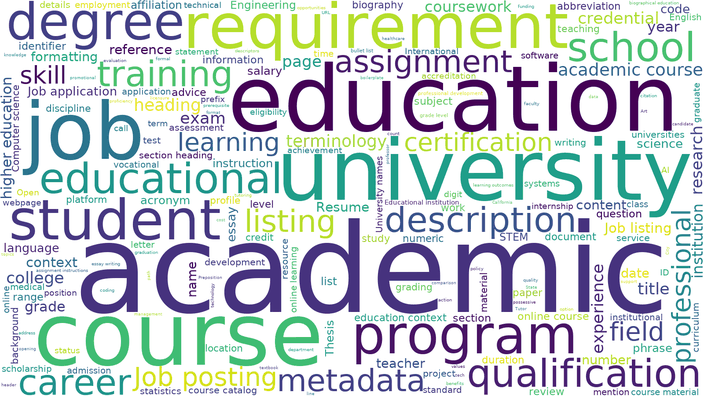}
  \wcrow{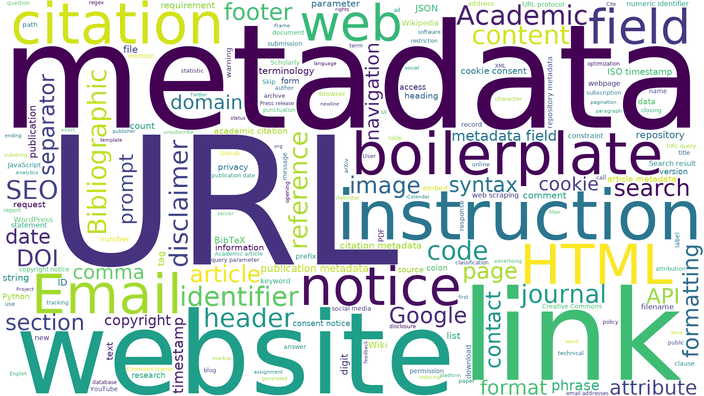}{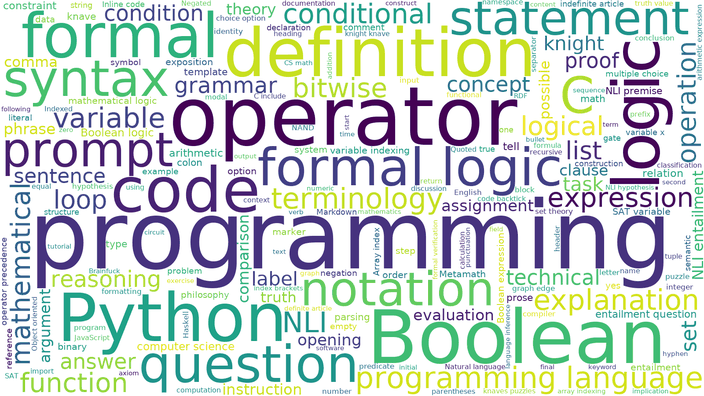}{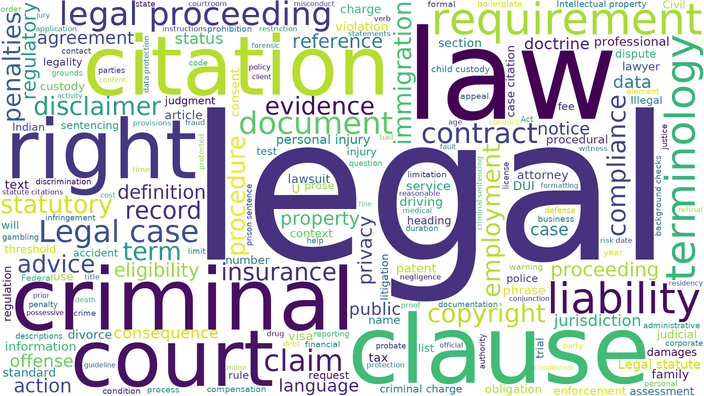}
  \wcrow{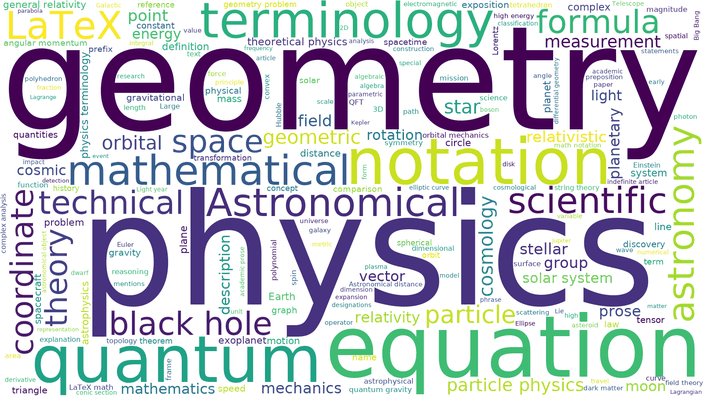}{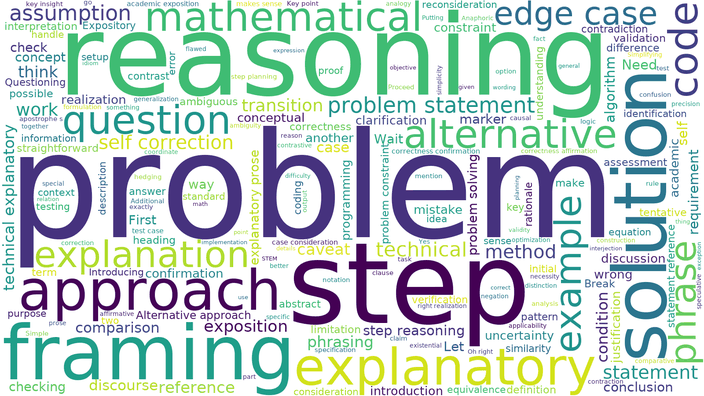}{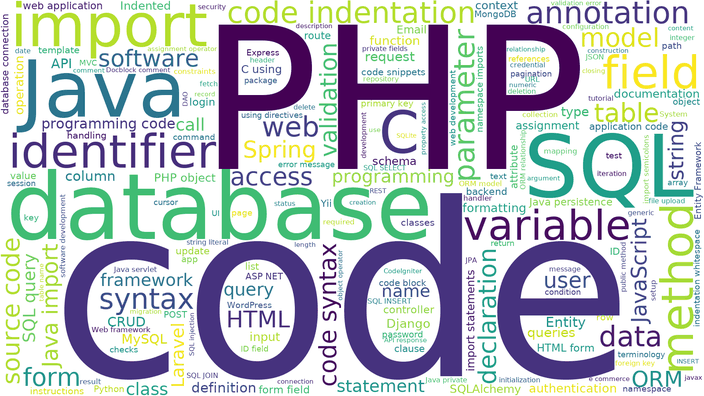}
  \wcrow{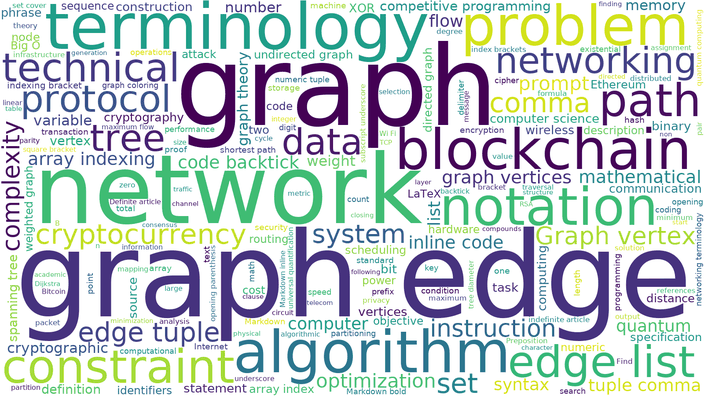}{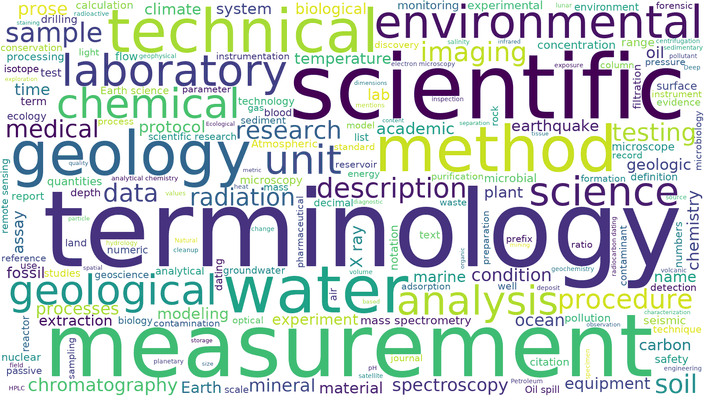}{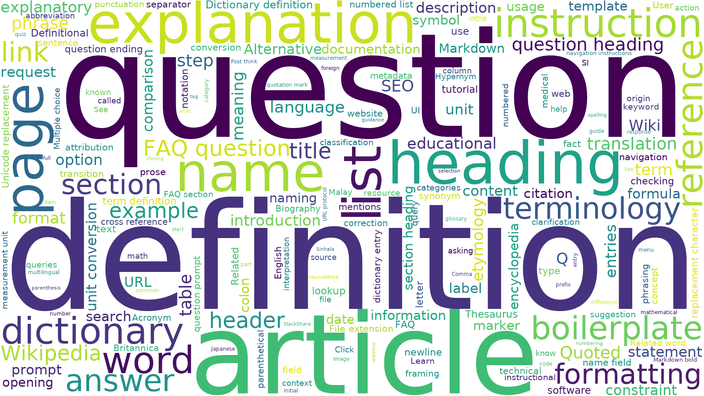}
  \wcrow{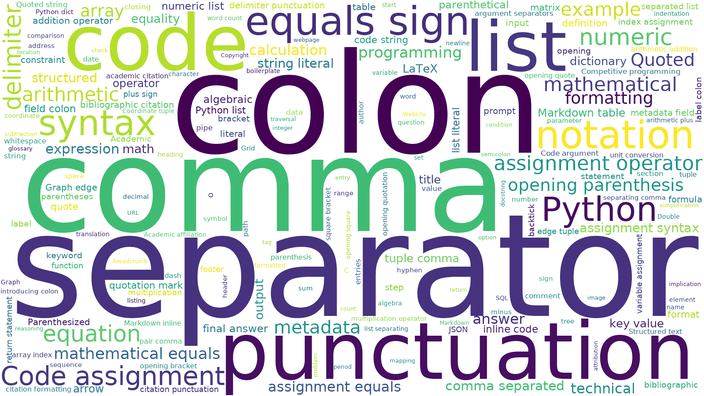}{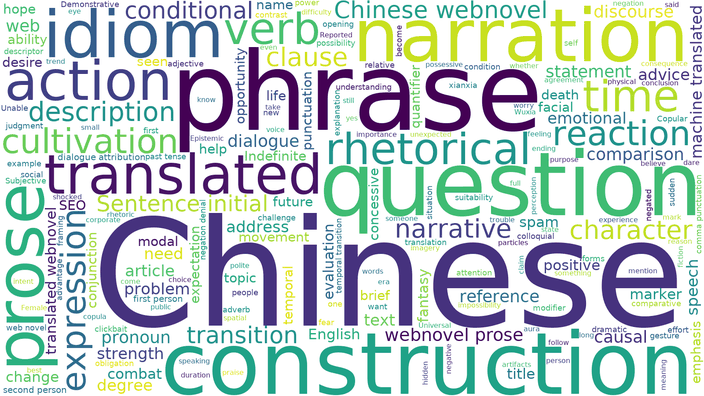}{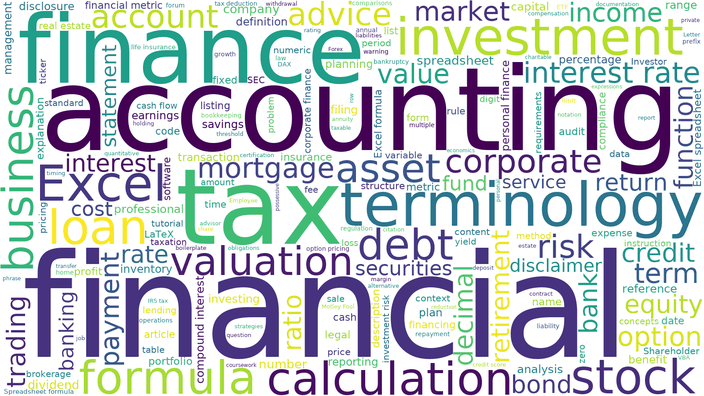}

  \caption{Explanation word clouds of experts 20 to 37 of layer 30, continued from
  \Cref{fig:wordcloud-all-1}.}
  \label{fig:wordcloud-all-3}
\end{figure}

\begin{figure}[p]
  \centering
  \wcrow{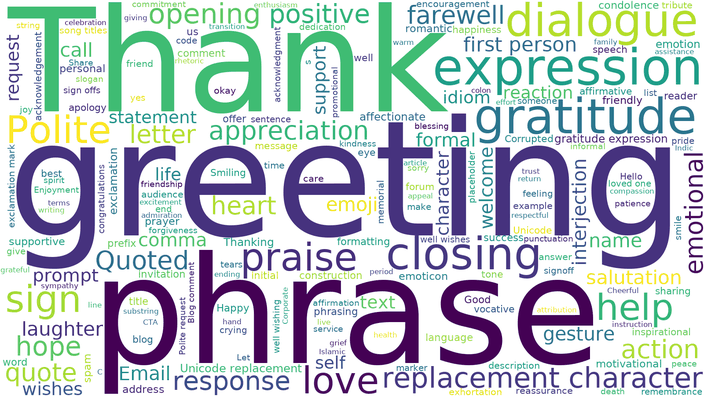}{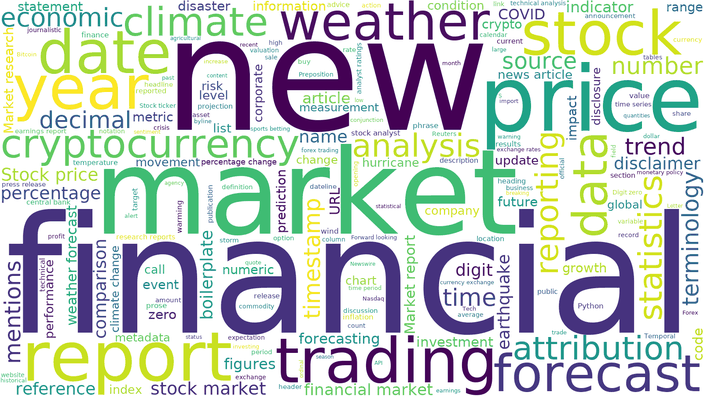}{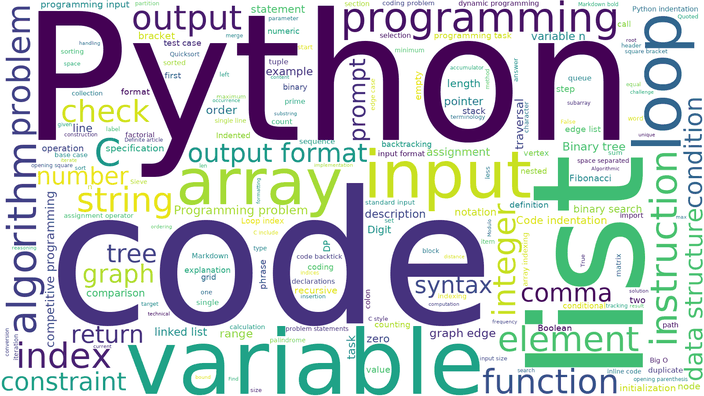}
  \wcrow{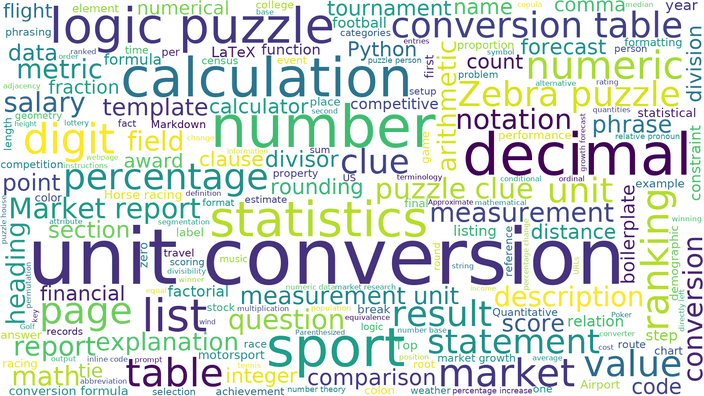}{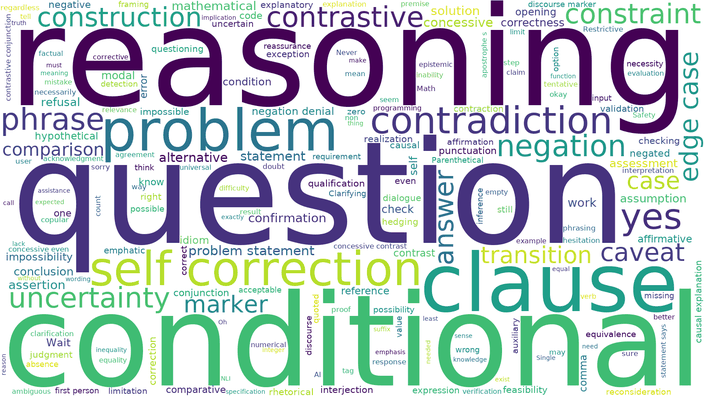}{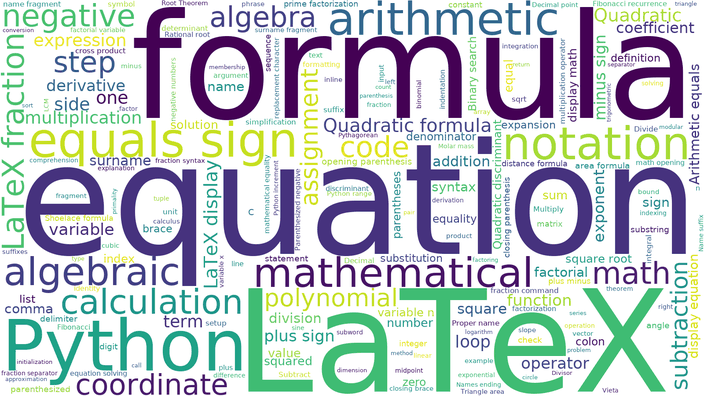}
  \wcrow{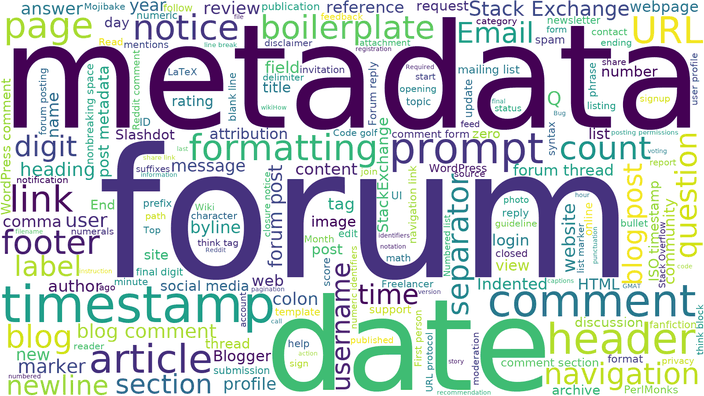}{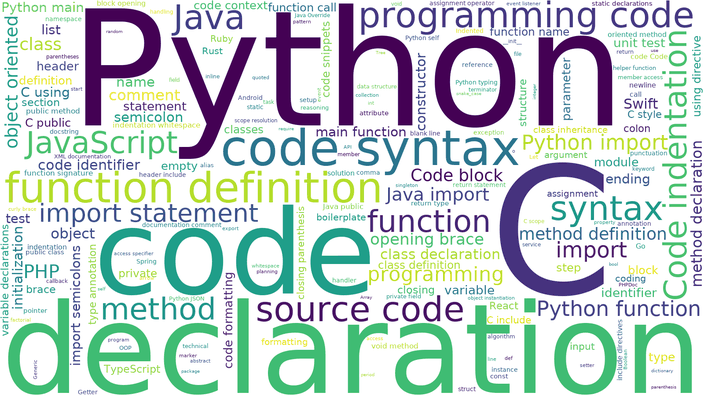}{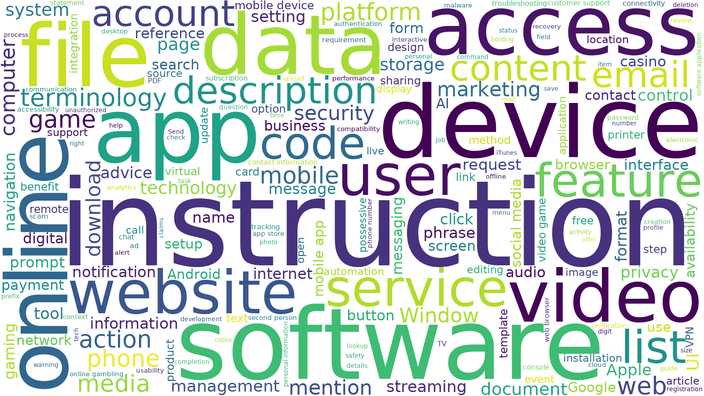}
  \wcrow{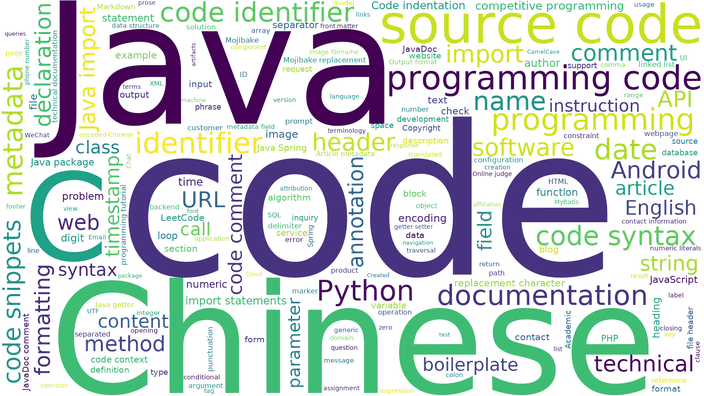}{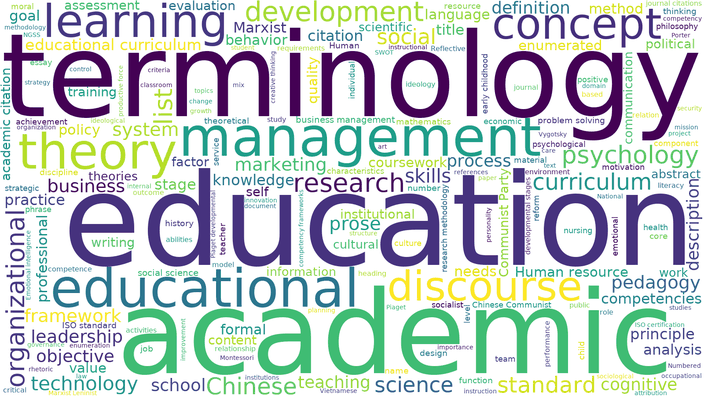}{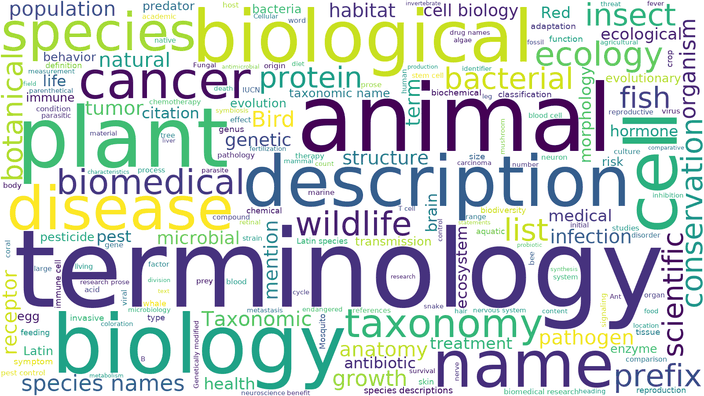}
  \wcrow{50}{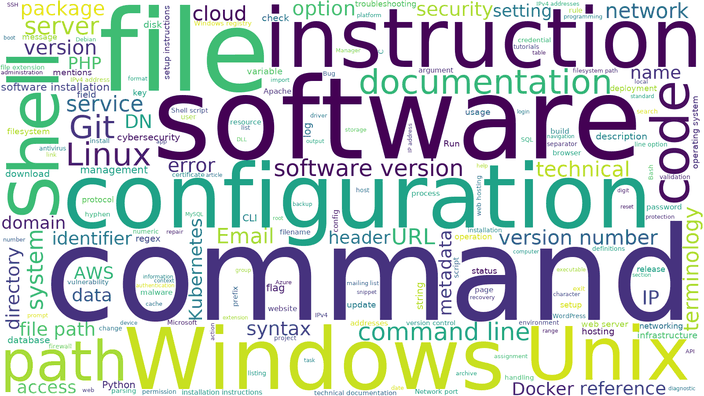}{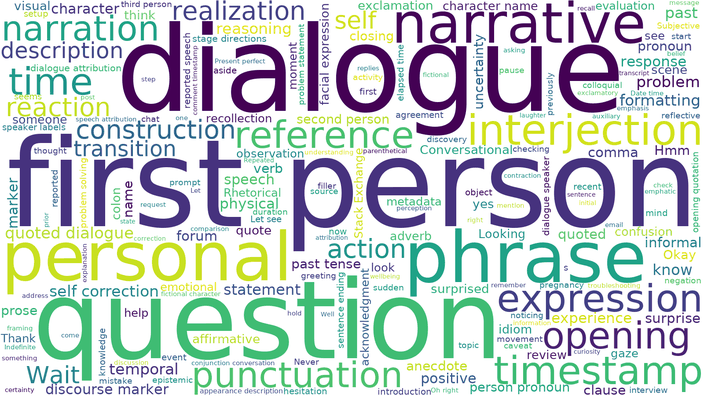}
  \wcrow{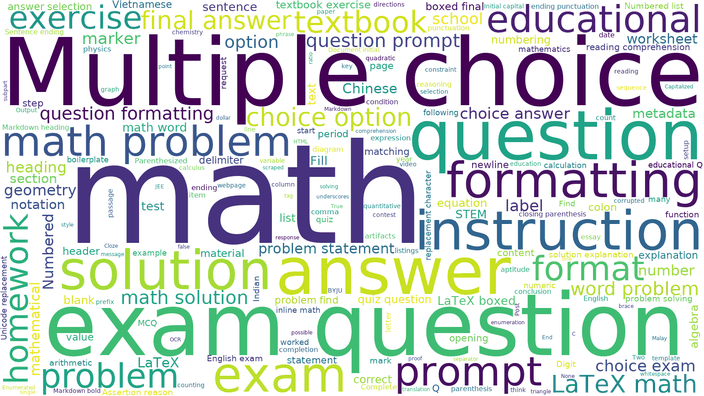}{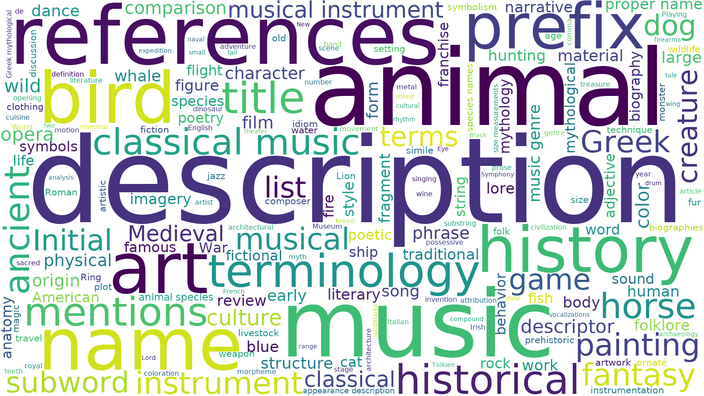}{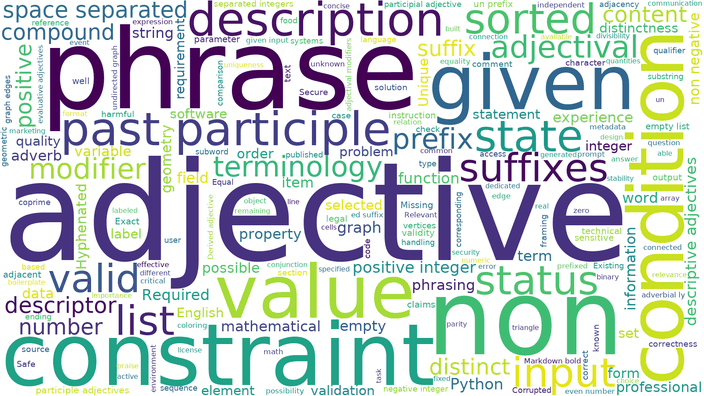}

  \caption{Explanation word clouds of experts 38 to 55 of layer 30, continued from
  \Cref{fig:wordcloud-all-1}.}
  \label{fig:wordcloud-all-4}
\end{figure}

\begin{figure}[p]
  \centering
  \wcrow{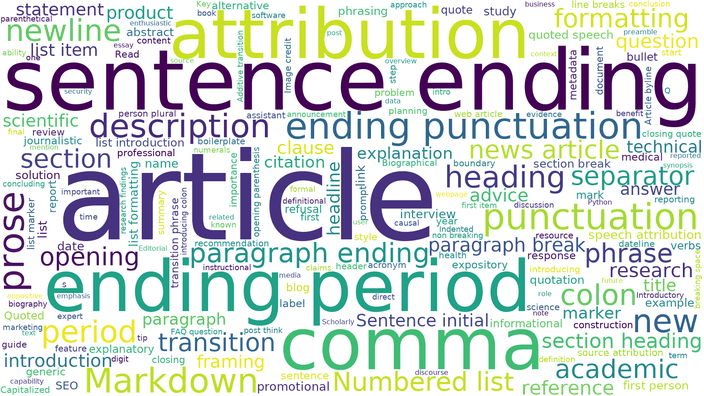}{57}{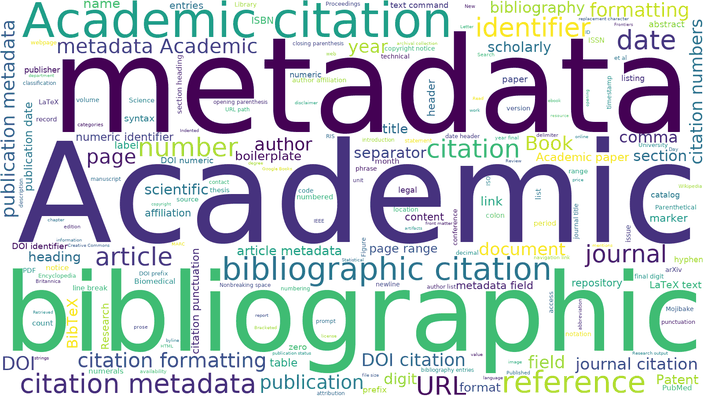}
  \wcrow{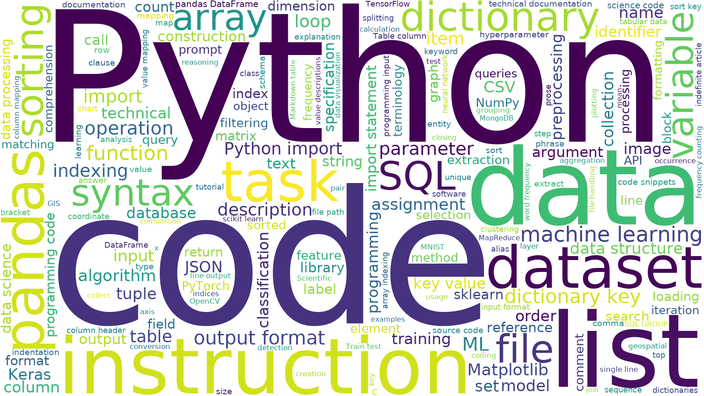}{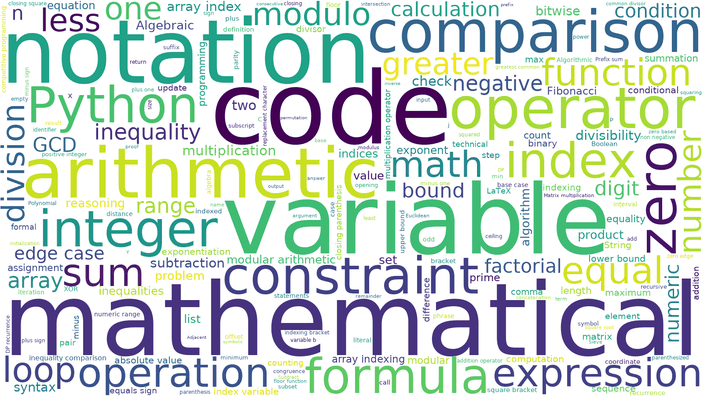}{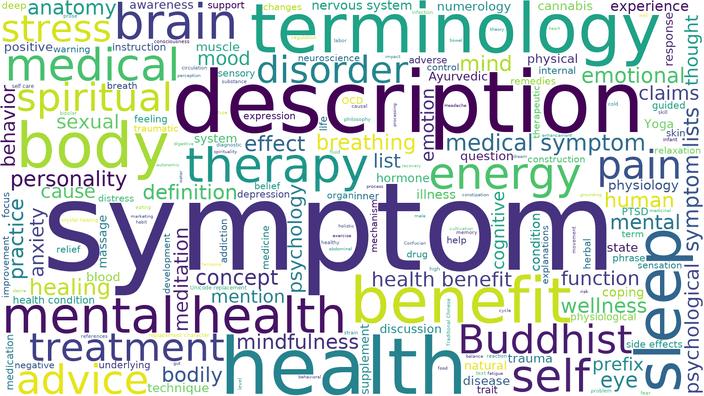}
  \wcrow{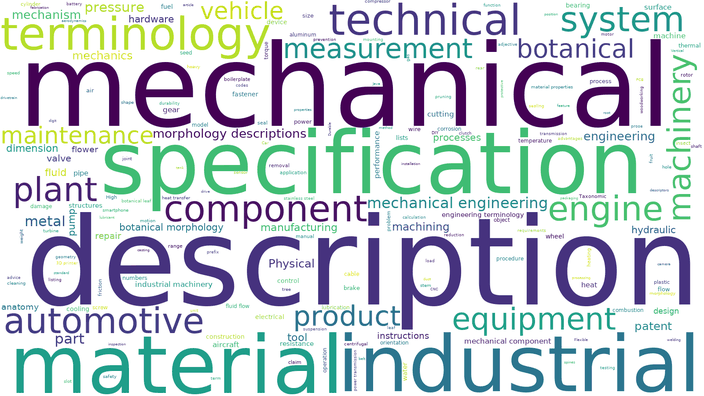}{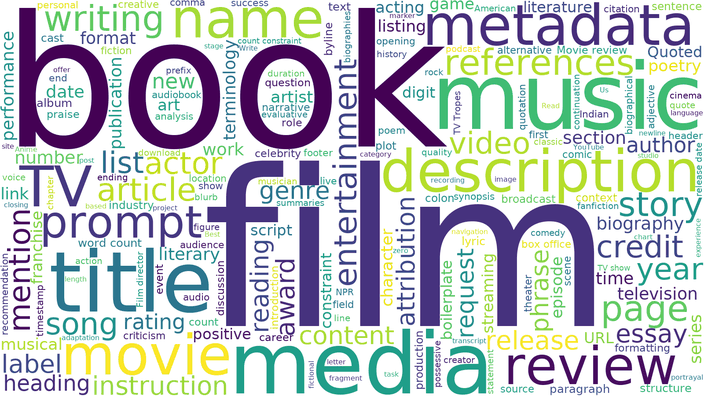}{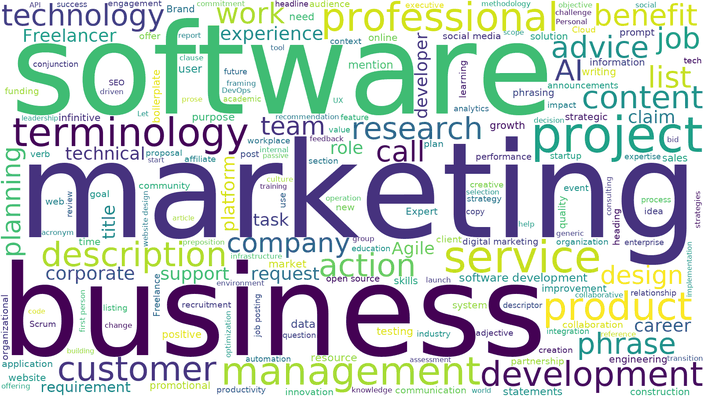}
  \wcrow{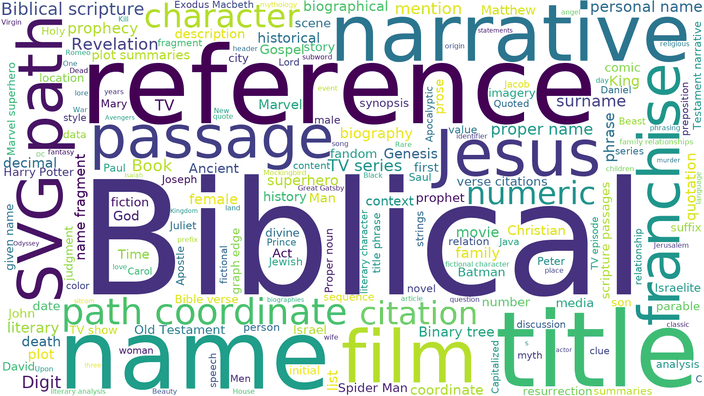}{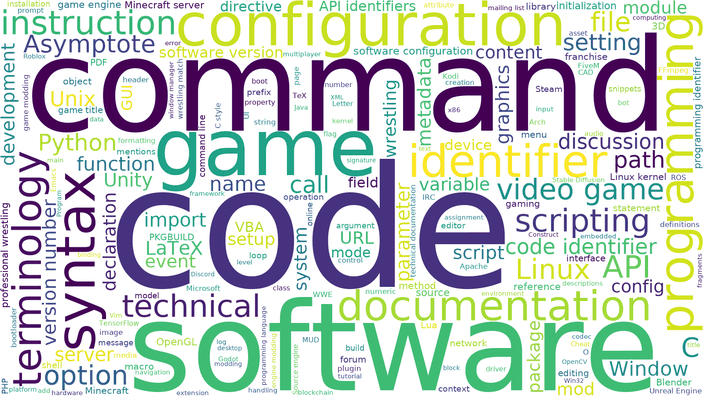}{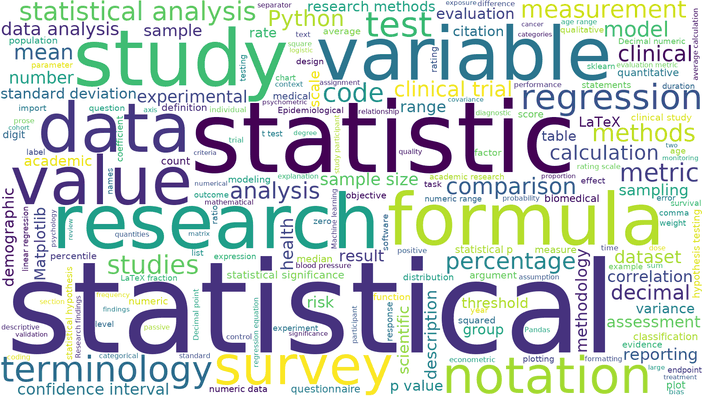}
  \wcrow{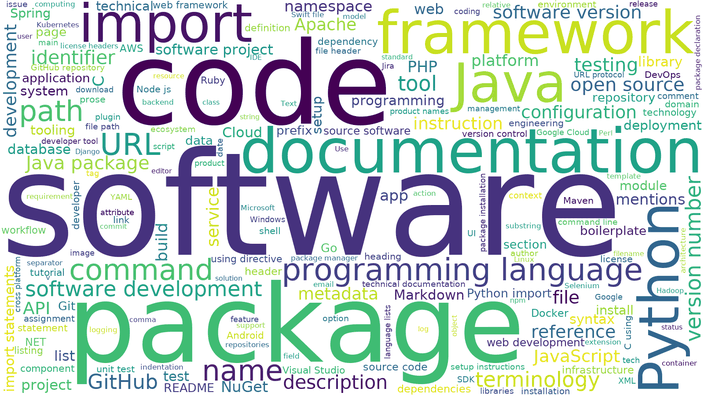}{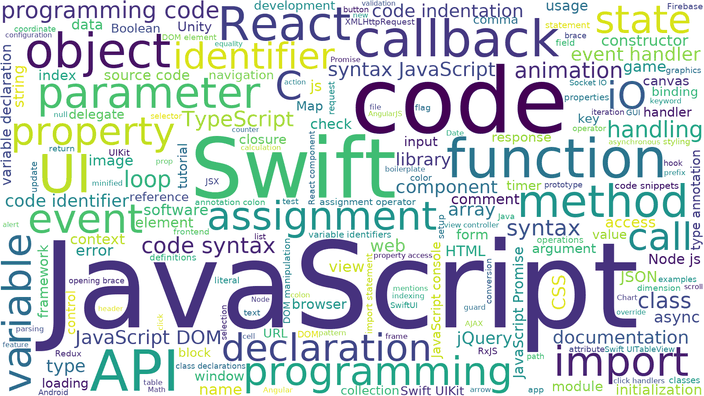}{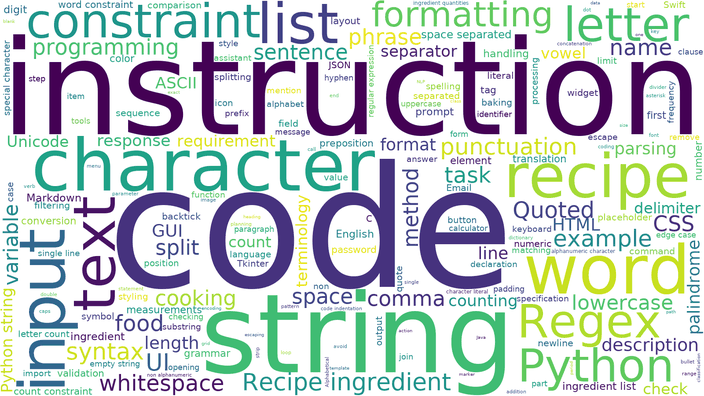}
  \wcrow{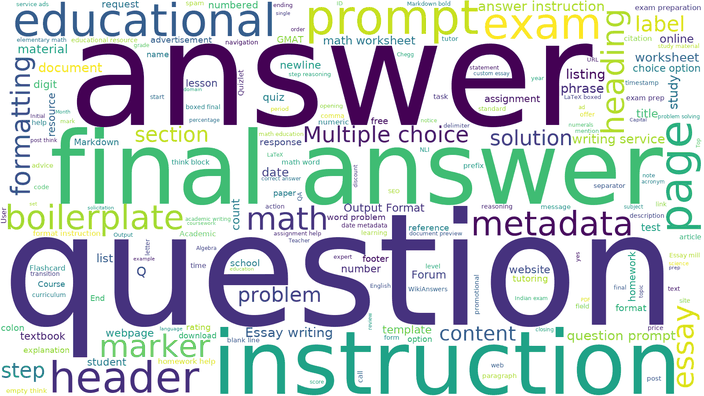}{72}{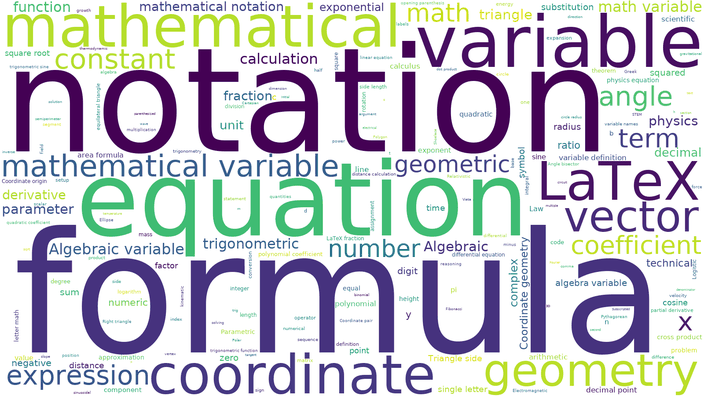}

  \caption{Explanation word clouds of experts 56 to 73 of layer 30, continued from
  \Cref{fig:wordcloud-all-1}.}
  \label{fig:wordcloud-all-5}
\end{figure}

\begin{figure}[p]
  \centering
  \wcrow{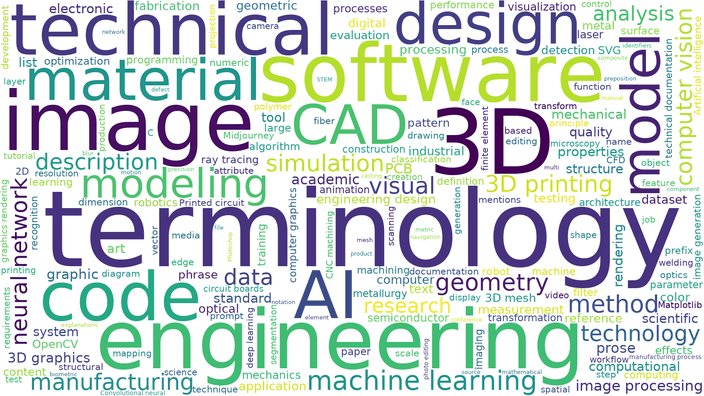}{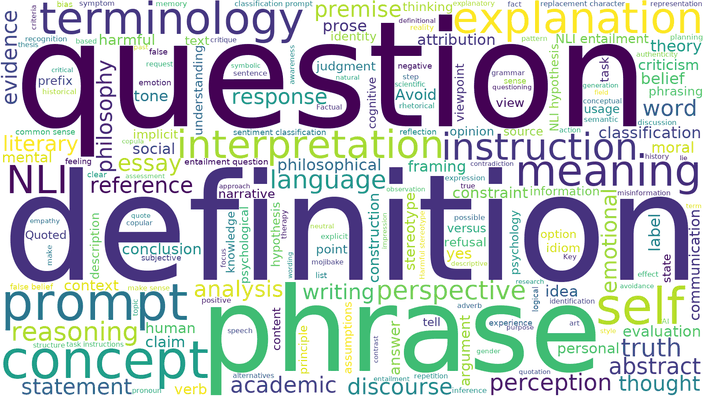}{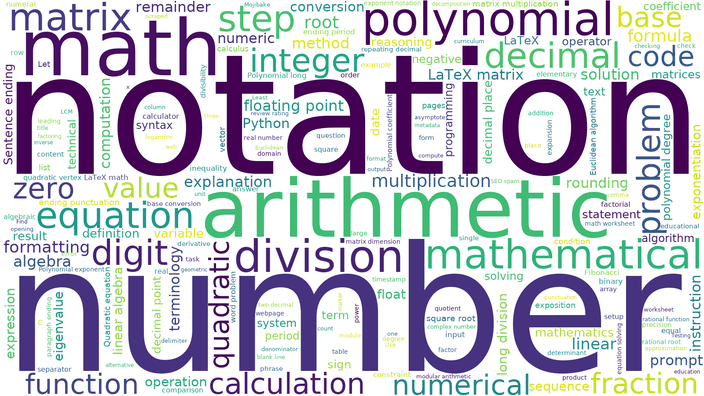}
  \wcrow{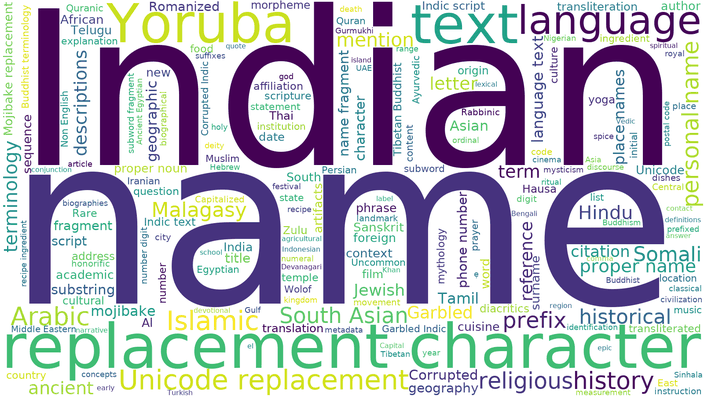}{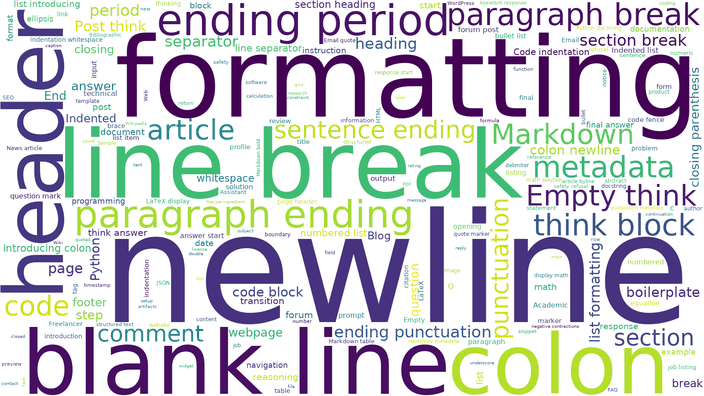}{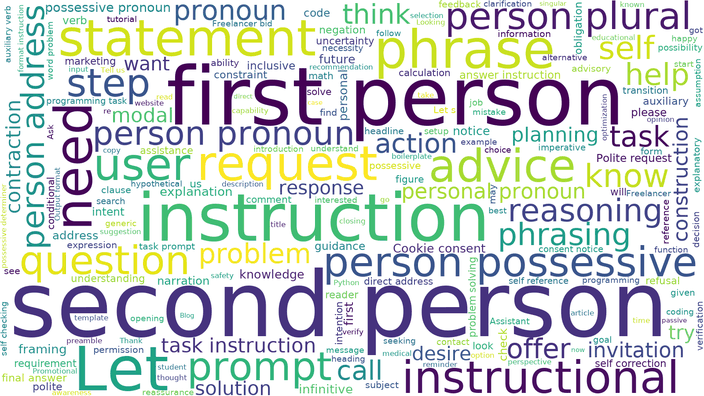}
  \wcrow{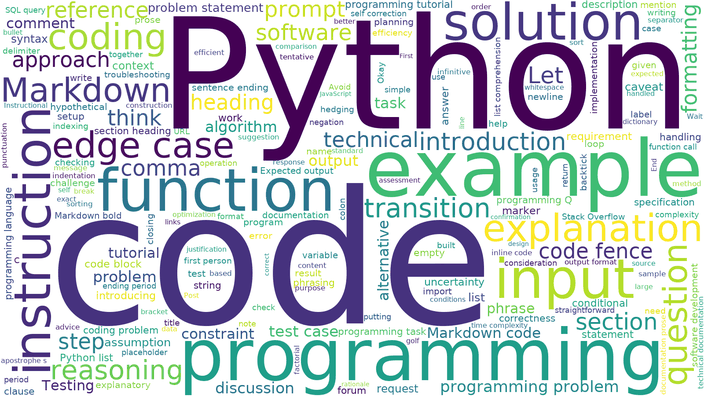}{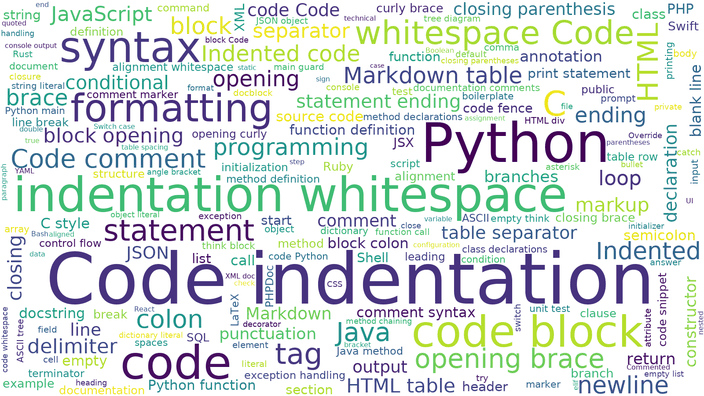}{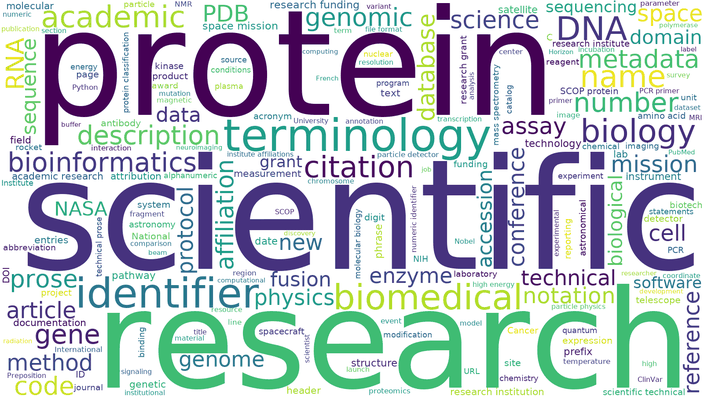}
  \wcrow{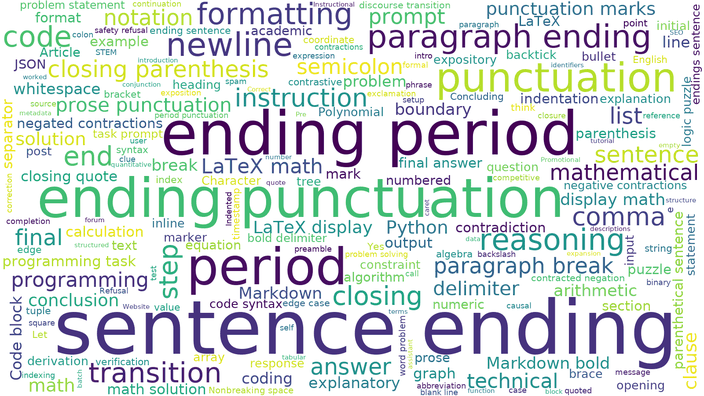}{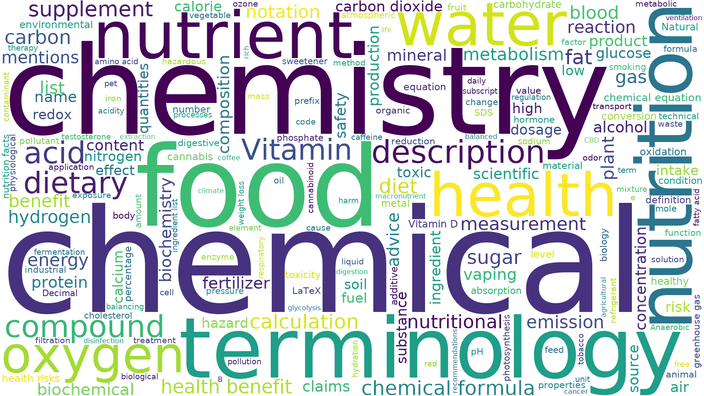}{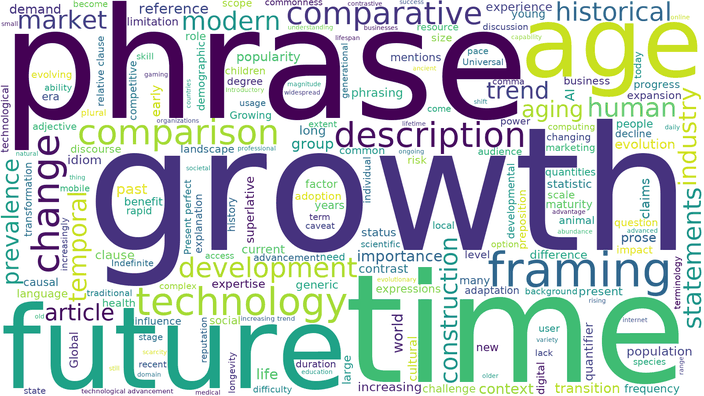}
  \wcrow{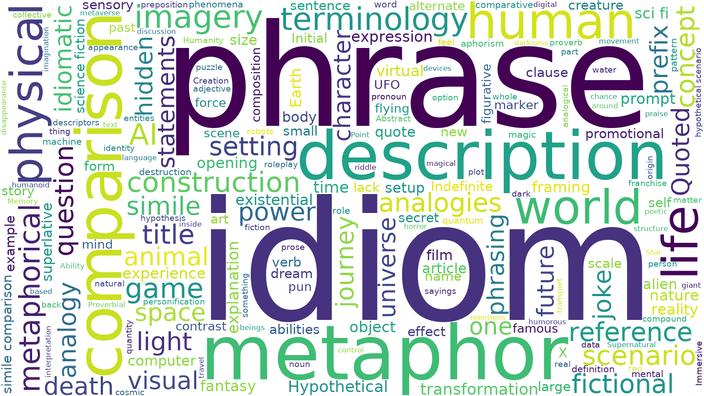}{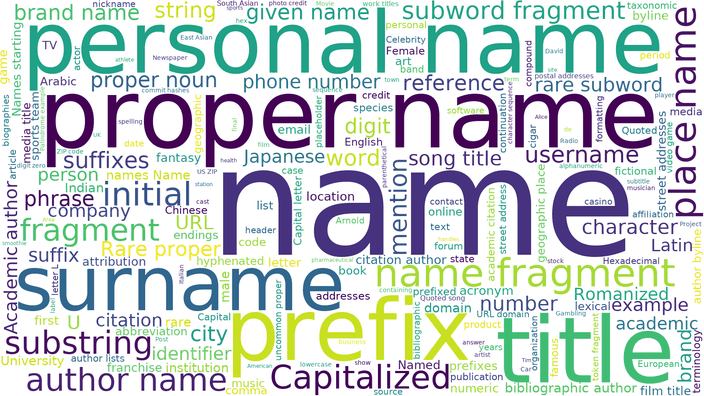}{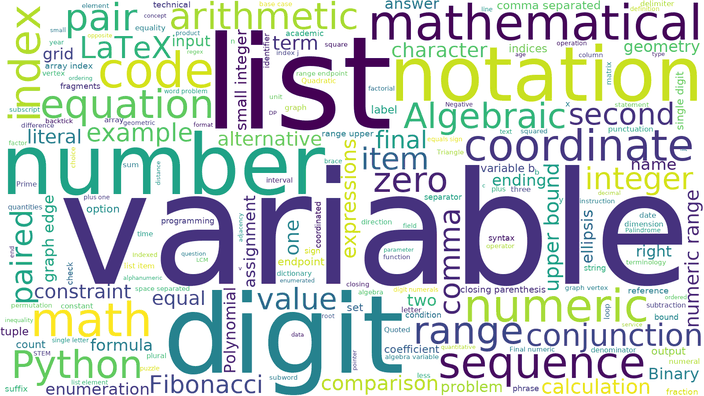}
  \wcrow{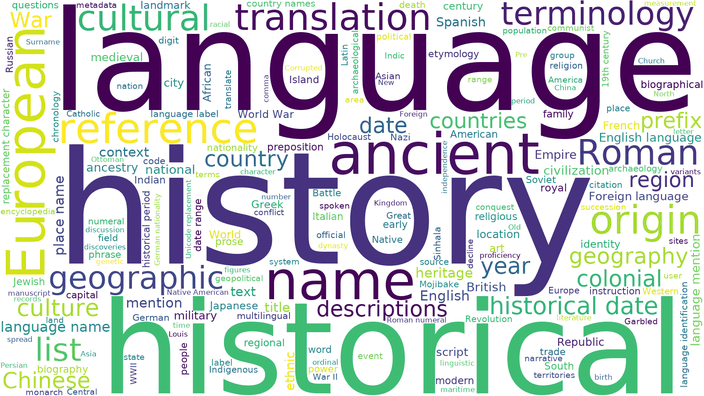}{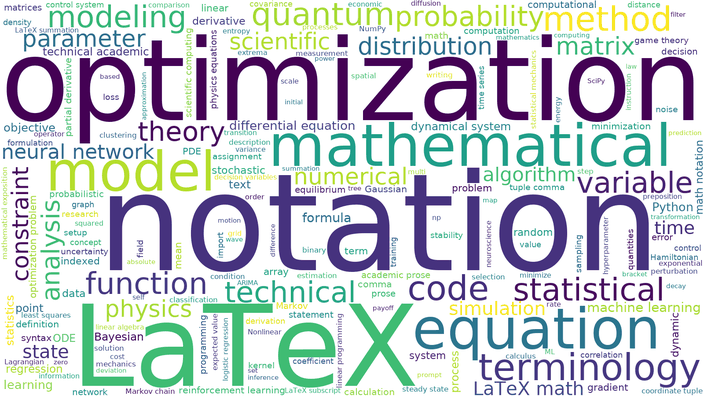}{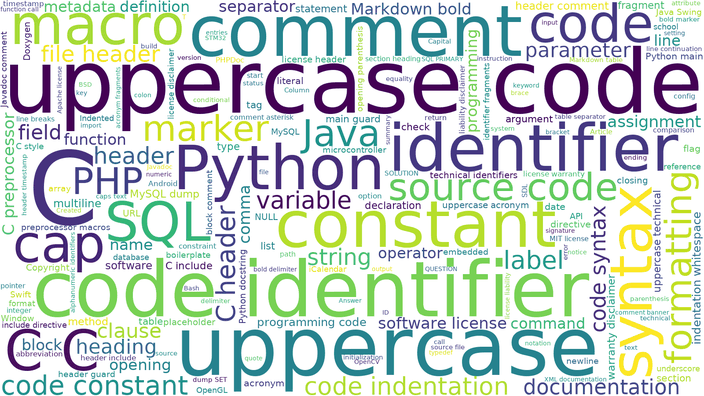}

  \caption{Explanation word clouds of experts 74 to 91 of layer 30, continued from
  \Cref{fig:wordcloud-all-1}.}
  \label{fig:wordcloud-all-6}
\end{figure}

\begin{figure}[p]
  \centering
  \wcrow{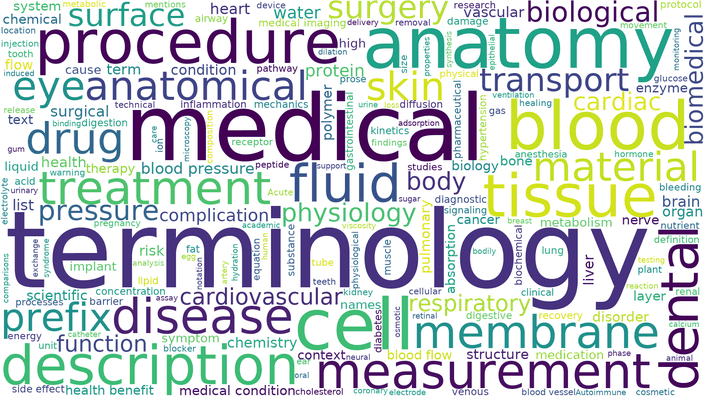}{93}{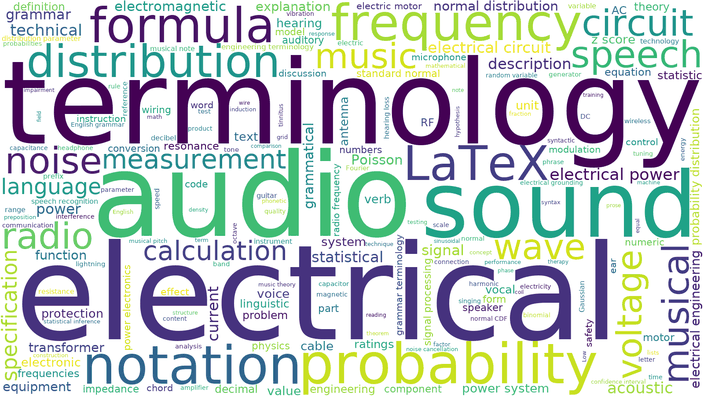}
  \wcrow{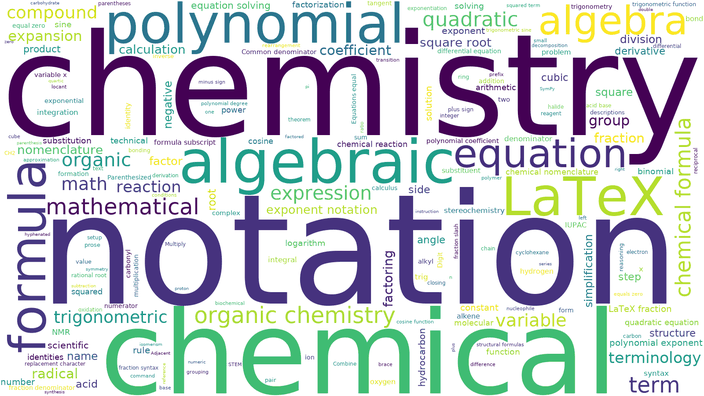}{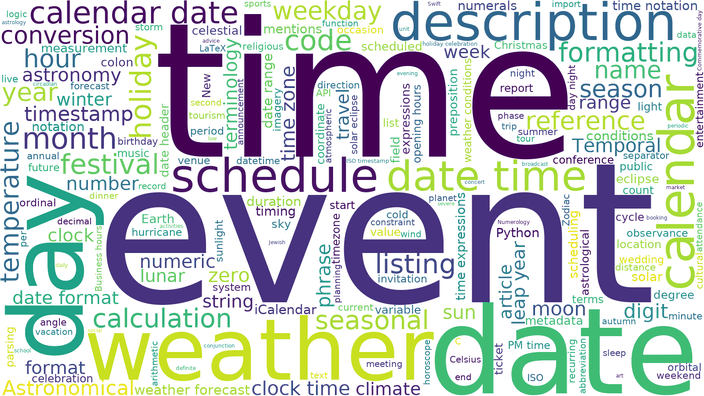}{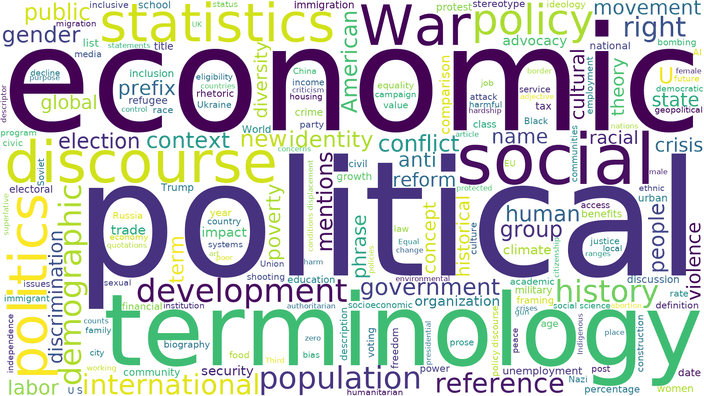}
  \wcrow{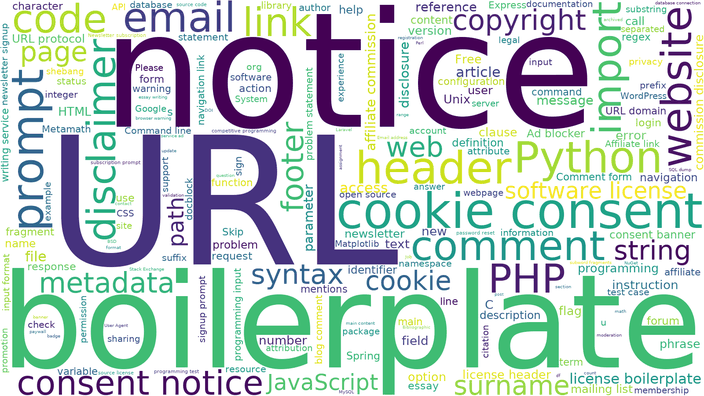}{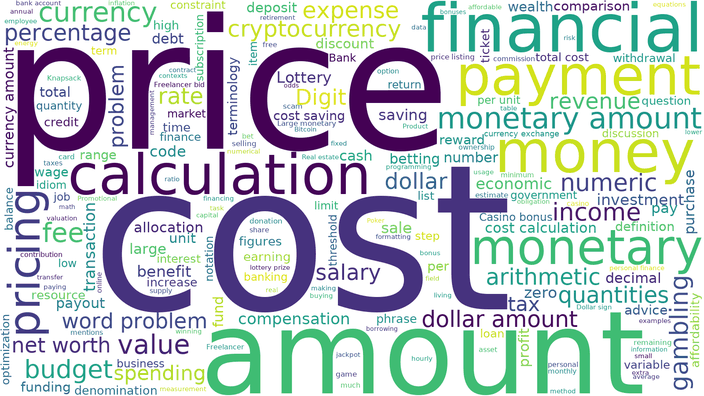}{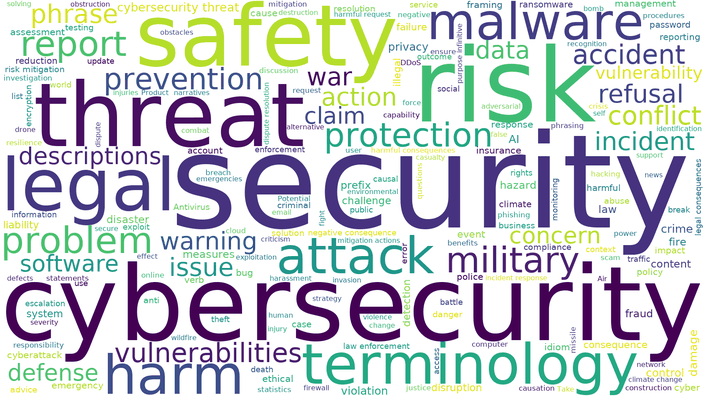}
  \wcrow{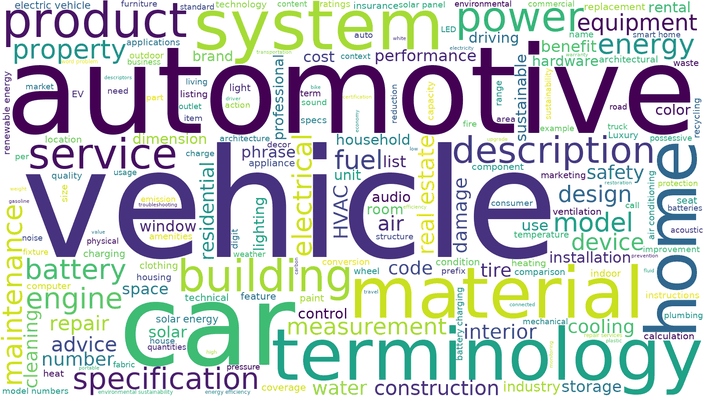}{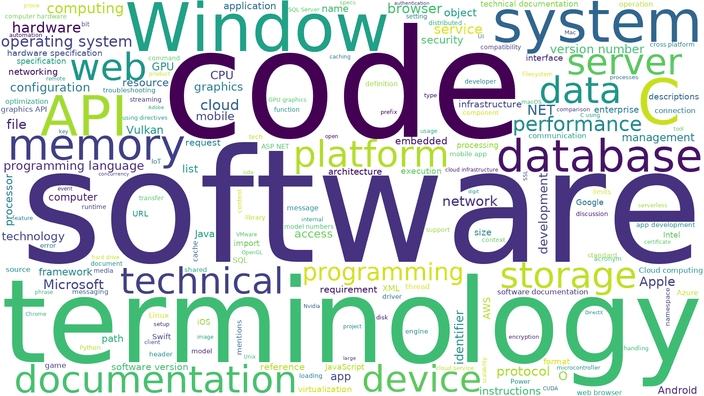}{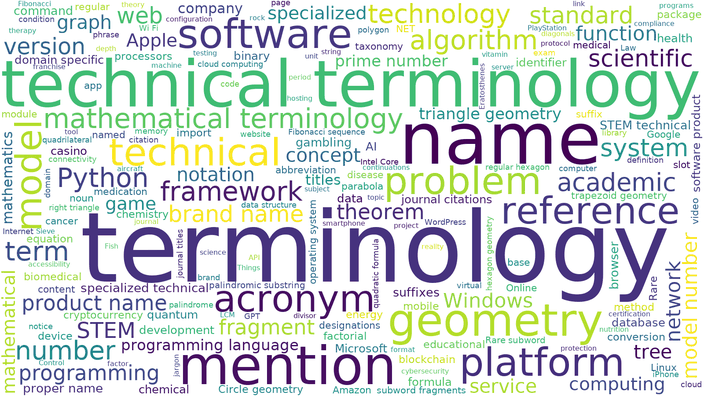}
  \wcrow{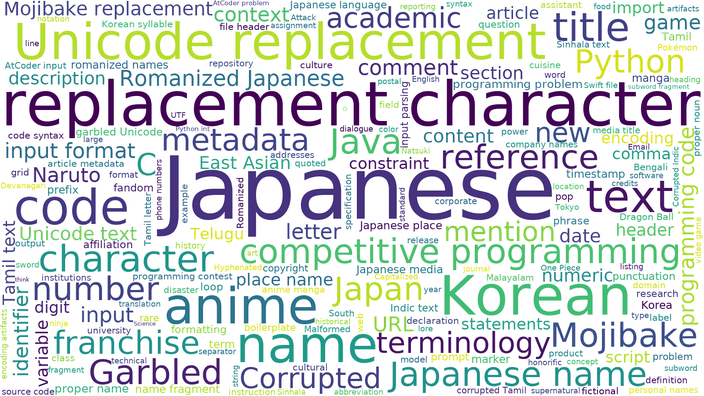}{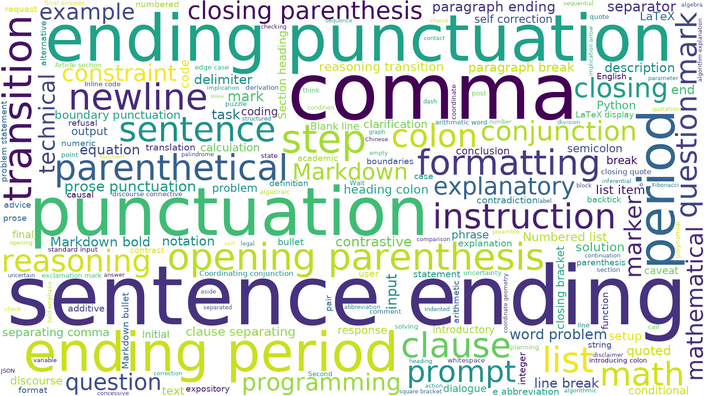}{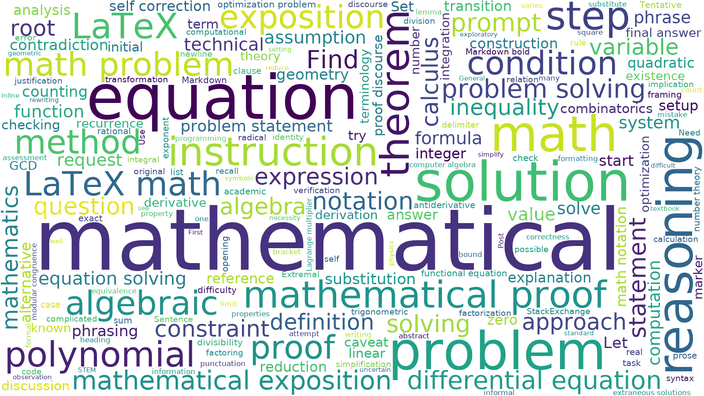}
  \wcrow{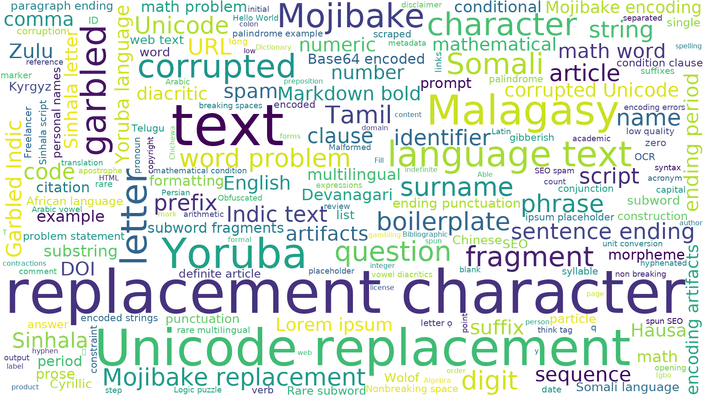}{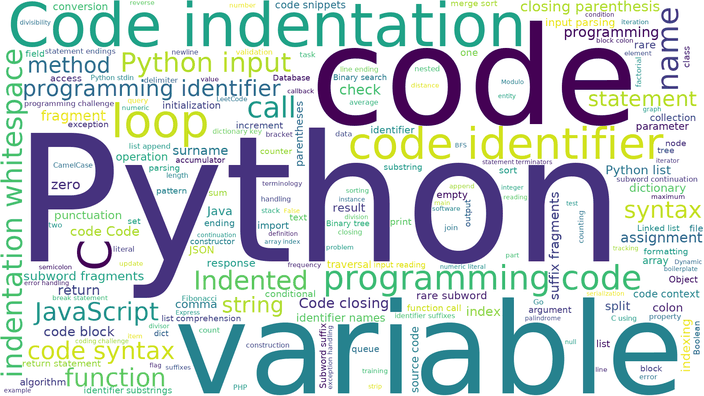}{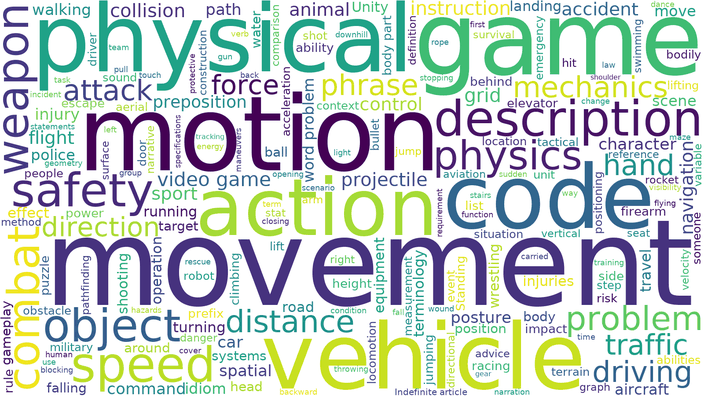}

  \caption{Explanation word clouds of experts 92 to 109 of layer 30, continued from
  \Cref{fig:wordcloud-all-1}.}
  \label{fig:wordcloud-all-7}
\end{figure}

\begin{figure}[p]
  \centering
  \wcrow{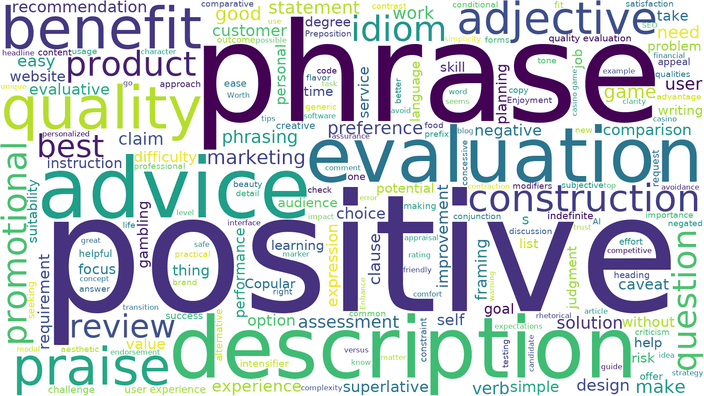}{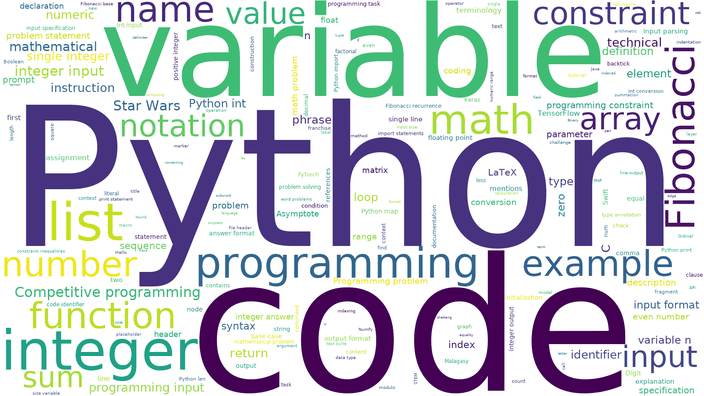}{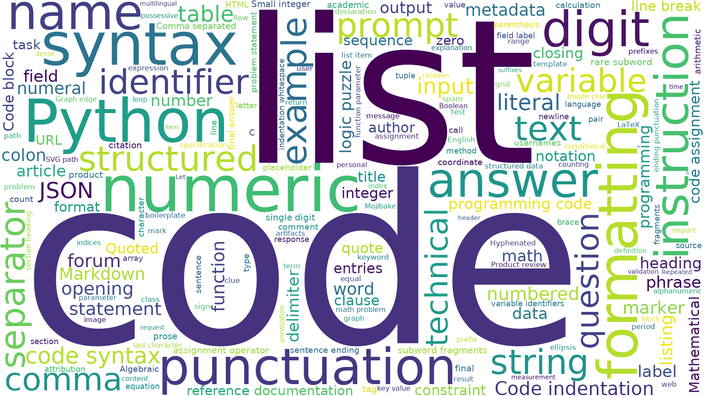}
  \wcrow{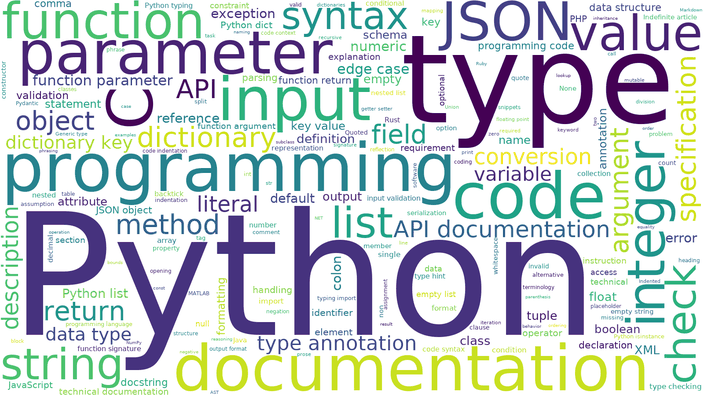}{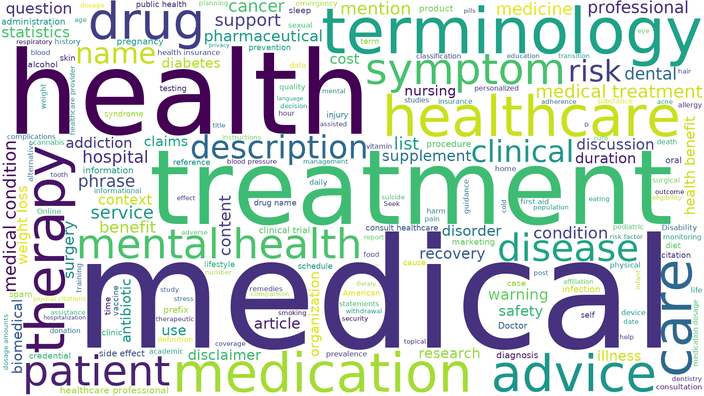}{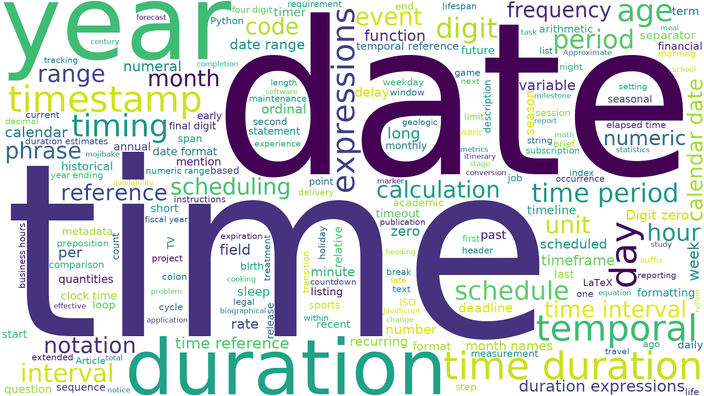}
  \wcrow{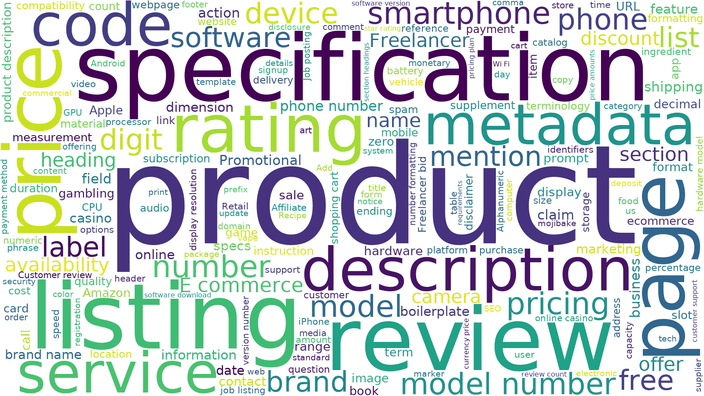}{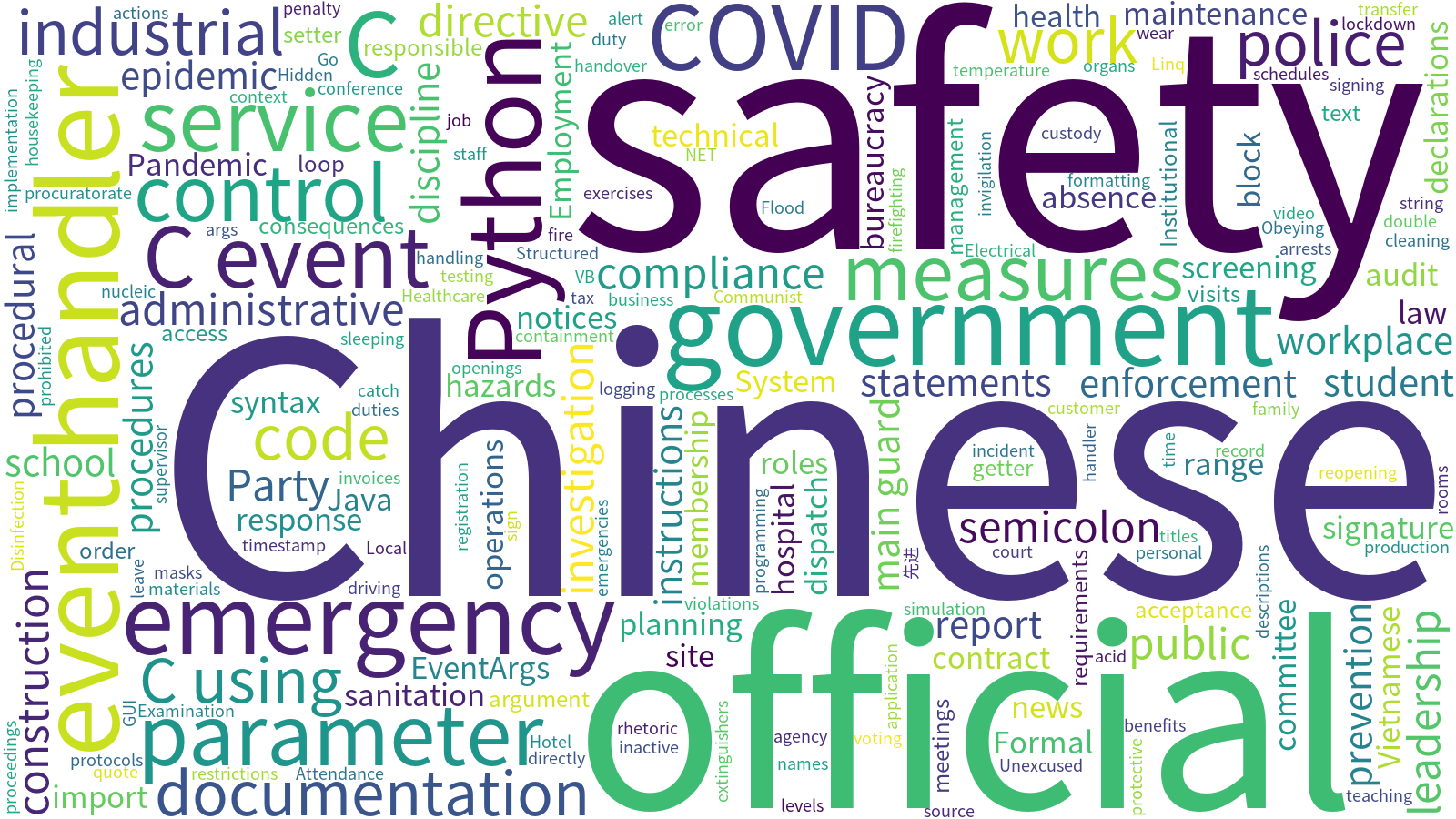}{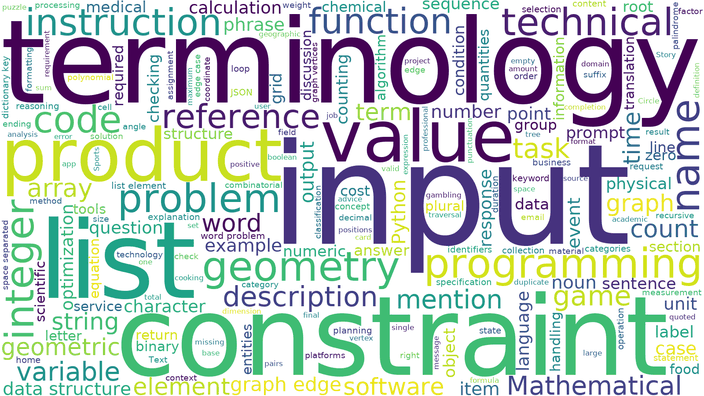}
  \wcrow{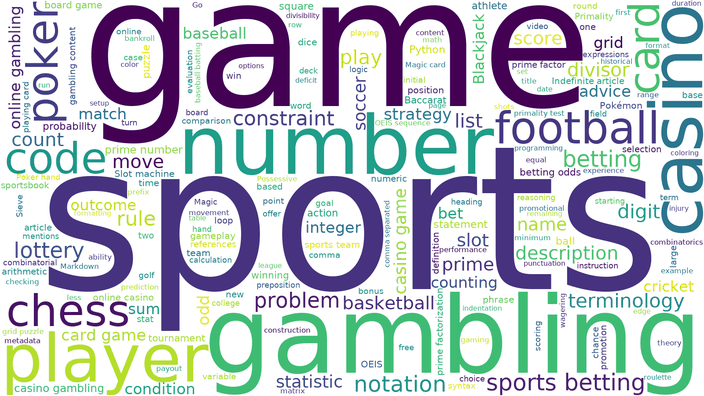}{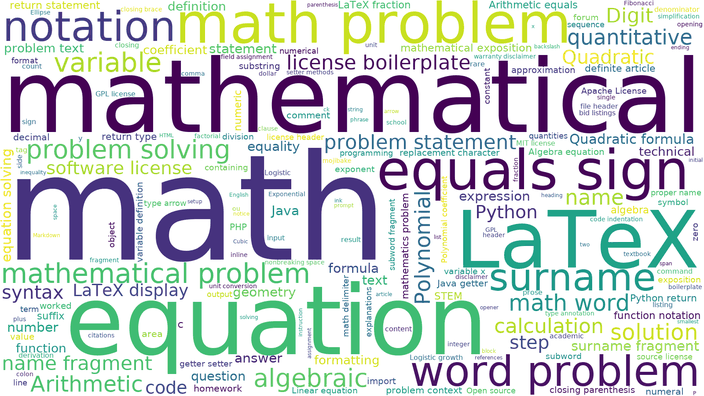}{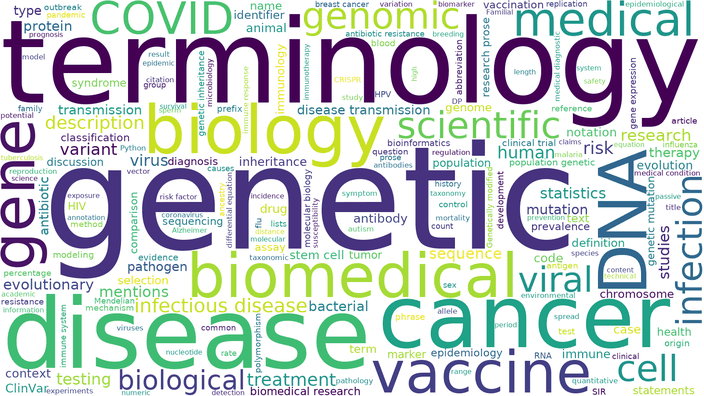}
  \wcrow{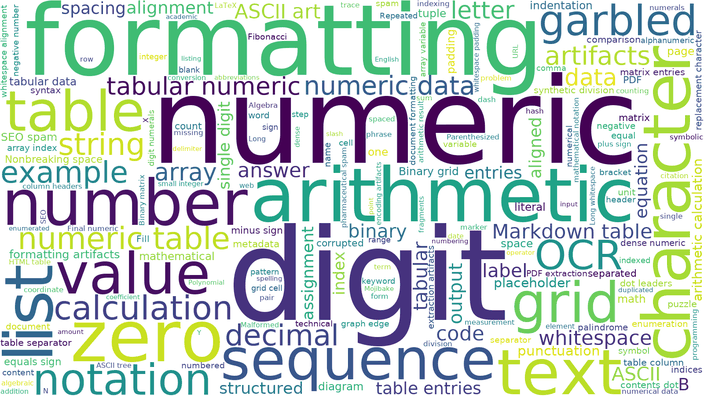}{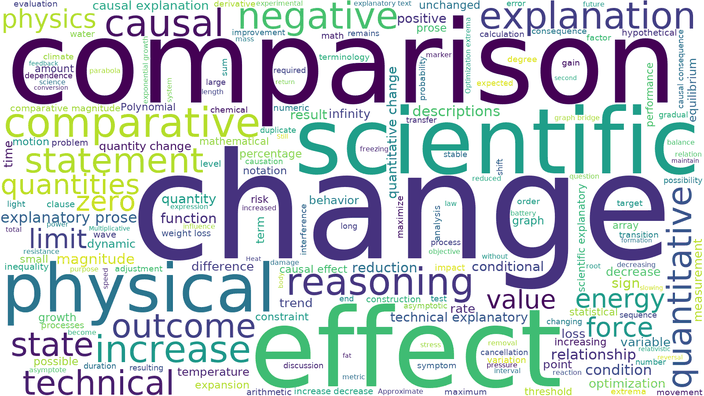}{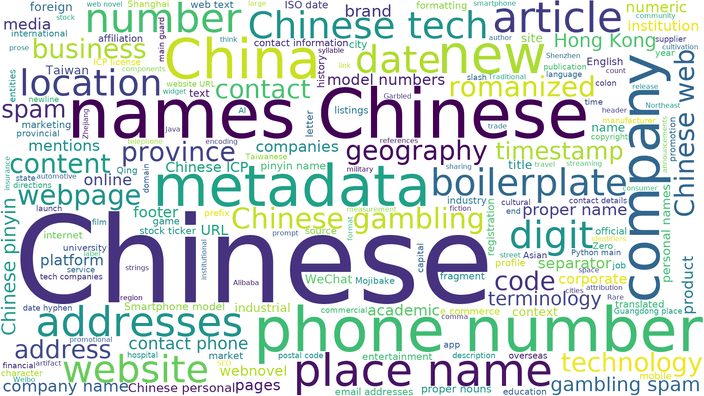}
  \wcrow{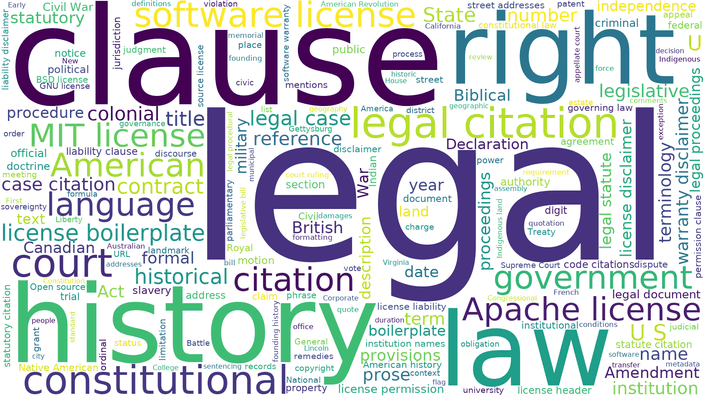}{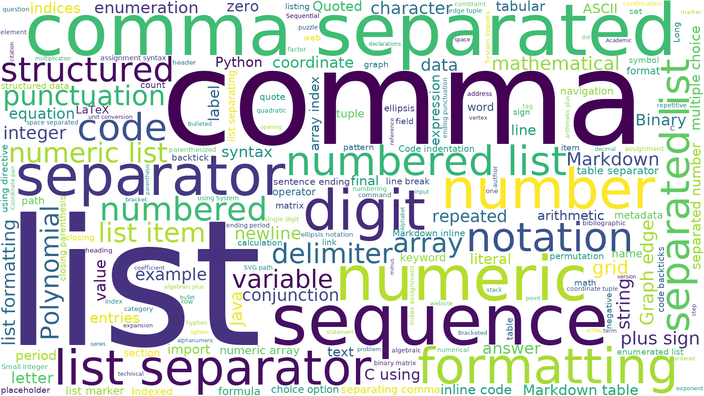}{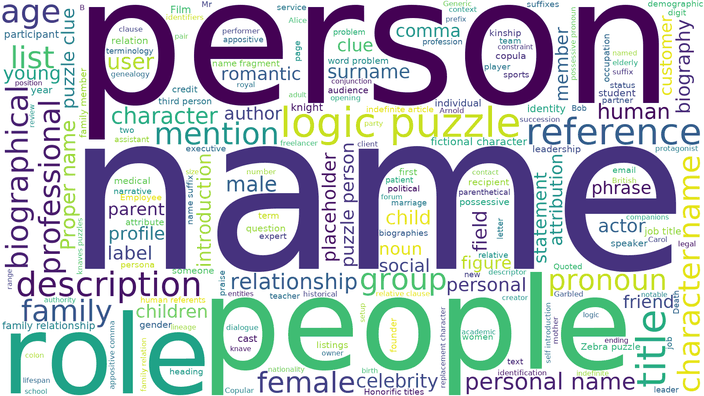}

  \caption{Explanation word clouds of experts 110 to 127 of layer 30, continued from
  \Cref{fig:wordcloud-all-1}.}
  \label{fig:wordcloud-all-8}
\end{figure}

\clearpage

\section{Prompt to judge whether a latent belongs to a theme}
\label{app:theme-check-prompt}

The theme name is substituted for \texttt{\{theme\}} in the prompt below, and the latent
explanation is appended after the final line. The model is asked to answer only
\texttt{yes} or \texttt{no}. The same prompt is used for the share row of
\Cref{tab:fine-grained} and for the random baseline, which draws $2000$ latents from the
dominant sets of every expert other than the one under study.

\begin{promptbox}
You are an expert in mechanistic interpretability, specializing in deciding whether a Sparse Autoencoder (SAE) latent of a large language model belongs to a given theme.

## Background

We study expert specialization in Qwen3-30B-A3B. Every MoE layer holds 128 experts. For each expert we recover the set of SAE latents whose activation that expert principally drives, and we interpret latents sampled from that set into a short natural language explanation, a noun phrase such as "Sailing knot names", "Origami folding steps" or "Falconry equipment terms".

## Task

Decide whether the latent explanation below belongs to the theme named after this paragraph. Belonging means the explanation is about that theme, not merely adjacent to it or sharing a generic word with it. For example, under the theme "Beekeeping", "Hive frame inspection" and "Queen bee marking" belong, while "Insect wing anatomy" and "Pollinator garden plants" do not, even though they are also about insects and flowers.

The theme is: {theme}

## Output format

Your response should contain only yes or no, in lowercase, with no other content.

The feature explanation is:
\end{promptbox}

\section{Prompt to label an expert from its explanation word cloud}
\label{app:expert-label-prompt}

The word cloud of an expert is passed to the model as one line per entry, of the form
\texttt{word: weight}, appended to the prompt below in decreasing order of weight.

\begin{promptbox}
You are an expert in mechanistic interpretability, specializing in describing what an individual expert of a Mixture-of-Experts (MoE) language model is responsible for.

## Background

We study expert specialization in Qwen3-30B-A3B. Every MoE layer holds 128 experts, and a router sends each token to a few of them. To find out what an expert is responsible for we do not look at the tokens routed to it, because a token is selected by the router but the information the expert adds lives in the features of the layer output.

A Sparse Autoencoder (SAE) with 2^20 latents is trained to reconstruct the output of one MoE layer, so each of its latents is one interpretable feature of what that layer computes. The activated part of a latent is an affine function of the layer output, so the activation of a latent decomposes exactly into a sum of per-expert contributions. Averaging the normalized contribution over a large corpus gives, for every expert and latent pair, the fraction of that latent's activation the expert is responsible for. When this fraction is large we say the expert dominates the latent, and the set of all latents an expert dominates is its dominant set. Dominant sets are nearly disjoint across experts, so the dominant set of an expert is a description of that expert alone.

Latents sampled from each dominant set were separately interpreted by a language model from activating contexts and logit effects. The result is a short natural language explanation, a noun phrase such as "Sailing knot names", "Origami folding steps" or "Falconry equipment terms".

## Input

You are given the word cloud of one expert, built from the explanations of latents sampled from its dominant set. Each entry is a word, or a frequent two word collocation, together with a weight. The weight is the frequency of that word among the explanations divided by the frequency of the most common word, so the heaviest entry always has weight 1.000. Common English stopwords have been removed, plural forms have been merged into their singulars, and at most 200 entries are listed in decreasing weight.

This word cloud is the only evidence you have about the expert. You are seeing the vocabulary of the explanations rather than the explanations themselves, so it tells you which concepts recur and how often, but not how they combined inside any single latent.

## Task

Name what this expert is responsible for, using at most nine words.

- Weigh words by their weights and spend your attention on the heaviest ones. A long tail of light words is normal and does not have to be covered.
- Look for what sets this expert apart from an arbitrary other expert. Words such as code, text, term, terminology, description, name, reference and value are frequent in many experts and are weak evidence on their own. Prefer the words that would be surprising in a random expert.
- Say what the expert is about instead of concatenating its heaviest words. For an expert whose cloud is led by hive, honey, beeswax and queen, "hive, honey, beeswax, queen" is a bad answer and "beekeeping" is a good one.
- You are not choosing from a fixed list of topics. An expert may specialize along an axis that is not a subject area at all. Name that axis when that is what the cloud shows.
- A few experts carry several unrelated themes at once. When the heavy words clearly split into separate groups, name each group and join them with "and", for instance "beekeeping and origami". The nine word limit binds here, so keep every theme to one or two words, and when the groups do not all fit, keep the heaviest ones and drop the rest. Do not force one label onto such an expert, and do not invent a second theme when the cloud only supports one.
- Be as specific as the evidence allows and no more specific than that. If the cloud is about geometry, say geometry rather than mathematics. If it spreads over several unrelated parts of mathematics, say mathematics.

## Output format

Your response should contain only the label with no other content. Write a noun phrase of at most nine words, with no trailing punctuation and no quotation marks. Keep the capitalization of proper nouns such as Kyoto, UNESCO and Darwin, and leave every other word in lowercase.

The word cloud of this expert is:
\end{promptbox}

\section{Prompt to group the experts into domains}
\label{app:expert-group-prompt}

The 128 labels are appended to the prompt below as one line per expert, of the form
\texttt{E\{expert id\}: \{label\}}. The only quantities we set are the bounds on the number of
groups, between 6 and 13, and the returned number of groups, eleven, falls strictly inside them.
The prompt fixes the level of grouping but names no domain that the partition should contain.

\begin{promptbox}
You are an expert in mechanistic interpretability, working on the division of labor among the experts of a Mixture-of-Experts (MoE) language model.

## Background

We study expert specialization in Qwen3-30B-A3B. One MoE layer holds 128 experts, and a router sends each token to a few of them. For every expert we recovered the set of Sparse Autoencoder latents whose activation that expert is chiefly responsible for, we interpreted latents sampled from that set into short natural language explanations, and we summarized those explanations into one short label naming what the expert is responsible for. Those 128 labels are the input below, one per expert.

The labels were written independently of one another, so their wording is not standardized. Two experts can hold closely related responsibilities while sharing no word, and two labels can share a generic word such as terminology, description or reference while describing responsibilities that have nothing to do with each other. Read for what an expert does rather than for surface overlap.

This grouping asks a coarse question. Group at the coarsest level the labels still support, and leave the internal structure of a group alone.

## Task

Partition the 128 experts into between 6 and 13 groups, aiming for about 10.

- Every expert belongs to exactly one group and all 128 experts are placed. Nothing may be dropped and nothing may appear twice.
- A group is good when its name is true of every expert in it. Among the names that stay true, take the most general one rather than the most specific, and prefer one large group under a general name over several small groups under specific ones.
- There is no upper bound on the size of a group, and group sizes do not need to be balanced. Do not split a group into subfields merely to even out the sizes.
- Do not cut a small coherent domain apart either, and do not manufacture groups merely to reach the group count.
- A group may be organized around any responsibility its members share.
- Many labels name two or three themes at once. Place such an expert by the theme that carries most of its label, preferring a theme that already has a home among the other experts. Do not open a group for a single odd pairing.
- Keep one axis in one place. Two experts responsible for the same kind of thing belong in the same group even when their labels attach that thing to different subject matter, so do not leave near duplicates of a single theme sitting in two different groups.
- Avoid a catch-all group. A group is a catch-all whenever some of its members lack the property its name states, and giving that group a specific sounding name does not repair it. Check every group by walking its members one at a time and asking whether the name is true of that member, then move out the ones for which it is not. If a few experts genuinely fit nowhere they may share one residual group, which must stay small and must be named for something its members actually have in common.
- Name each group as generally and as briefly as you can, one or two words wherever the group allows it. Two closely related things may be joined with and. What is not acceptable is a name that enumerates three or more subject areas, of the shape "A, B, and C", because such a name states no organizing principle at all. When the only name you can find for a candidate group is a list of that shape, either find the more general word that covers the list, or split the group along the very boundaries the list names.

## Output format

Respond with JSON and nothing else. No prose before or after it and no markdown code fence. The JSON is a list of objects ordered from the largest group to the smallest, each object having exactly two fields:

- "group", the name of the group
- "experts", the ids of the experts in that group, as integers in ascending order

The shape of the response is

[{"group": "<name of the largest group>", "experts": [0, 7, 19]}, {"group": "<name of the next group>", "experts": [3, 12]}]

The 128 expert labels are:
\end{promptbox}

\section{Expert labels and domains}
\label{app:expert-labels}

\Cref{tab:expert-labels} gives the label of every expert of layer 30 together with the domain it
was grouped into, the experts of one domain being listed together. The domains appear in
decreasing order of size, and the experts within a domain in increasing order of identity.

\small\begin{longtable}{>{\raggedright\arraybackslash}p{2cm}
                  p{1cm}
                  >{\raggedright\arraybackslash}p{\dimexpr\linewidth-3cm-6\tabcolsep\relax}}
  \caption{The 128 experts of layer 30, their labels and the domains they were grouped
  into. A label is written by GPT-5.5 from the explanation word cloud of that expert alone, and
  the domains are recovered from the labels in a single further call. Both prompts are given in
  \Cref{app:expert-label-prompt} and \Cref{app:expert-group-prompt}.}
  
  \label{tab:expert-labels} \\
  \hline
  Domain & Expert & Label \\
  \hline
  \endfirsthead

  \hline
  Domain & Expert & Label \\
  \hline
  \endhead

  \hline
  \endfoot
  \multirow[t]{28}{=}{\raggedright Computing} & 9 & competitive programming statements and formal instruction phrasing \\
   & 19 & conditional programming logic and edge case handling \\
   & 20 & embedded multimedia hardware programming and specifications \\
   & 21 & C and Python programming input/output \\
   & 31 & database-backed web application code \\
   & 32 & graph algorithms and blockchain protocols \\
   & 40 & competitive programming and algorithmic code \\
   & 45 & source code syntax and declarations \\
   & 47 & Chinese Java source code and technical documentation \\
   & 50 & Java and Android programming \\
   & 51 & command-line software configuration and system administration \\
   & 57 & HTML CSS and JavaScript frontend development \\
   & 59 & Python data processing and machine learning code \\
   & 60 & integer arithmetic and algorithmic code notation \\
   & 66 & software configuration command syntax and game modding \\
   & 68 & software development packages and documentation \\
   & 69 & JavaScript web and Swift iOS programming \\
   & 70 & programming string manipulation and recipes \\
   & 72 & coding algorithms and musculoskeletal fitness \\
   & 81 & code indentation and block syntax formatting \\
   & 100 & cybersecurity and safety risk management \\
   & 102 & software engineering and computing platforms \\
   & 108 & Python code syntax and indentation \\
   & 111 & Python competitive programming and Star Wars \\
   & 112 & code syntax and structured list formatting \\
   & 113 & Python programming types and API documentation \\
   & 117 & government safety and event handler programming \\
   & 118 & competitive programming problem specifications \\
  \hline
  \multirow[t]{21}{=}{\raggedright Text form} & 8 & Markdown answer formatting and Python/math problem solving \\
   & 12 & academic document structure and formatting \\
   & 24 & structured text formatting and delimiters \\
   & 26 & webpage metadata and academic citation boilerplate \\
   & 34 & definitions and reference page formatting \\
   & 35 & punctuation and separators in code and notation \\
   & 44 & web forum metadata and boilerplate \\
   & 56 & article prose punctuation and formatting \\
   & 58 & academic citation metadata and bibliography formatting \\
   & 71 & educational Q\&A and exam page formatting \\
   & 77 & non-English proper names and Unicode mojibake \\
   & 78 & newlines and structured text formatting \\
   & 80 & Python programming solutions and Markdown formatting \\
   & 83 & sentence endings and punctuation formatting \\
   & 98 & web and source code boilerplate notices \\
   & 104 & Japanese text, Unicode mojibake, and competitive programming \\
   & 105 & punctuation and text formatting \\
   & 107 & Unicode replacement characters and mojibake \\
   & 122 & numeric tables and OCR formatting artifacts \\
   & 124 & Chinese web boilerplate and contact metadata \\
   & 126 & comma separated lists and numeric sequences \\
  \hline
  \multirow[t]{20}{=}{\raggedright Mathematics and reasoning} & 1 & coordinate and Euclidean geometry \\
   & 2 & formal mathematical notation and problem setup \\
   & 11 & LaTeX math notation and algebraic expressions \\
   & 13 & mathematical definitions and corrupted SEO prose \\
   & 15 & LaTeX notation for advanced mathematics and physics \\
   & 16 & quantitative STEM word problems \\
   & 27 & Boolean operators in programming and formal logic \\
   & 29 & mathematical physics, geometry, and astronomy \\
   & 30 & problem-solving reasoning and solution exposition \\
   & 41 & unit conversions, sports statistics, and logic puzzles \\
   & 42 & conditional reasoning and self-correction \\
   & 43 & mathematical equations, LaTeX, and Python syntax \\
   & 53 & multiple-choice math exam questions and solutions \\
   & 67 & statistical analysis and research methods \\
   & 73 & mathematical formulas and notation \\
   & 76 & mathematical notation, arithmetic, and algebra \\
   & 88 & mathematical and programming variables, indices, and lists \\
   & 90 & mathematical optimization and modeling notation \\
   & 106 & mathematical problem solving and proofs \\
   & 120 & mathematical equations and problem solving \\
  \hline
  \multirow[t]{18}{=}{\raggedright Applied science} & 0 & industrial energy and infrastructure sectors \\
   & 6 & chemistry terminology and notation \\
   & 18 & optics, conic sections, and civil engineering \\
   & 33 & scientific measurement and geoscience laboratory methods \\
   & 49 & biological and biomedical terminology \\
   & 61 & medical symptoms and mind-body wellness \\
   & 62 & mechanical engineering specifications and botanical morphology \\
   & 74 & 3D CAD engineering and computer vision \\
   & 82 & biomedical research terminology and identifiers \\
   & 84 & food chemistry and nutrition \\
   & 92 & medical anatomy and physiology terminology \\
   & 94 & electrical/audio engineering, probability notation, and grammar terminology \\
   & 95 & chemical and algebraic notation \\
   & 101 & automotive and residential building systems \\
   & 103 & STEM and software technical nomenclature \\
   & 114 & medical treatment and healthcare advice \\
   & 121 & biomedical genetics and disease terminology \\
   & 123 & scientific causal reasoning about quantitative change \\
  \hline
  \multirow[t]{8}{=}{\raggedright Language} & 36 & Chinese webnovel translationese and discourse constructions \\
   & 38 & polite greetings, thanks, and sign-offs \\
   & 52 & first-person dialogue and narrative discourse markers \\
   & 55 & adjectival morphology and technical constraints \\
   & 75 & definitions and NLI tasks \\
   & 79 & first and second person instructional dialogue \\
   & 85 & temporal and comparative framing of growth and change \\
   & 86 & idioms and figurative comparisons \\
  \hline
  \multirow[t]{7}{=}{\raggedright Culture and recreation} & 3 & Christian theology and role-playing game mechanics \\
   & 54 & animals and classical arts \\
   & 63 & entertainment media metadata and reviews \\
   & 65 & biblical references, fiction franchises and SVG path coordinates \\
   & 89 & history, languages and cultural geography \\
   & 109 & physical movement in games, vehicles, and combat \\
   & 119 & sports, games, gambling, and number theory \\
  \hline
  \multirow[t]{7}{=}{\raggedright Entities and identifiers} & 5 & programming identifier and variable names \\
   & 17 & numeric strings and identifiers \\
   & 22 & official government, military, and corporate entity names \\
   & 23 & biographical life events and relationships \\
   & 87 & proper names titles identifiers and their fragments \\
   & 91 & uppercase code identifiers constants and comments \\
   & 127 & personal names, roles, and logic puzzle characters \\
  \hline
  \multirow[t]{5}{=}{\raggedright Safety and advice} & 4 & food recipes and ingredient descriptions \\
   & 7 & everyday household, cooking, and safety advice \\
   & 10 & refusal constructions and safety request framing \\
   & 46 & digital device and software usage instructions \\
   & 110 & positive evaluative phrasing in advice and marketing \\
  \hline
  \multirow[t]{5}{=}{\raggedright Commerce} & 37 & financial accounting, tax, and investment calculations \\
   & 39 & financial market news and weather forecasts \\
   & 64 & business software marketing and project management \\
   & 99 & monetary amounts and pricing calculations \\
   & 116 & e-commerce product listings and specifications \\
  \hline
  \multirow[t]{5}{=}{\raggedright Society and law} & 25 & academic education and job requirements \\
   & 28 & legal documents and proceedings \\
   & 48 & education and management theory terminology \\
   & 97 & political economy and social policy discourse \\
   & 125 & legal documents, citations, and American history \\
  \hline
  \multirow[t]{4}{=}{\raggedright Time and place} & 14 & spatial and geographic locations \\
   & 93 & physical geography and weather \\
   & 96 & calendar events, date-time expressions, and weather \\
   & 115 & date, time, and duration expressions \\
  \hline
\end{longtable}

\section{SAE training details}
\label{app:sae_training}

We trained our SAE with the latent size of $2^{20} = 1,048,576$ on the layer 4,30,46 MoE layer output stream of Qwen3-30B-A3B~\citep{yang2025qwen3} independently. Activations were collected from a mixture of pre-training and post-training data taken from the Olmo~3~\citep{olmo2025olmo3} training corpus: $5/8$ from pre-training data, $1/4$ from a reasoning chain-of-thought dataset, $1/8$ from an instruction-following supervised fine-tuning dataset.

In order to increase the diversity of the activation sources within a single batch, we dumped all the activations before training. For each example in the dataset, we ran a forward pass of the LLM to obtain the MoE activations and then sampled only $512$ of them from that example into the pool, discarding the remaining activations. For each training-batch, we drew a batch from activations we dumped uniformly at random without replacement.

Our hyperparameters are set following the common practice. We used a square-root warmup
followed by a linear decay. During the first $100$ steps, the learning rate is warmed up from
$0.1\eta_{\max}$ to the peak value $\eta_{\max}$ following a square-root schedule, i.e.,
$\eta(t) = \eta_0 + (\eta_{\max} - \eta_0)\sqrt{t / 100}$ with $\eta_0 = 0.1\eta_{\max}$.
The learning rate then linearly decays from $\eta_{\max}$ to $\eta_0$ during the remainder of
training. We set $\eta_{\max} = 8 \times 10^{-5}$.

We trained the SAE with the AdamW optimizer~\citep{loshchilov2017decoupled} with $(\beta_1,\beta_2) = (0.9, 0.999)$ and a weight decay of 0.01. We used a batch size of $8192$ and trained each SAE on about $3.2$B tokens. 

Our initialization setup generally follows~\citet{conerly2025dictionarylearning} and~\citet{ameisen2025circuit}. Let $n = 2048$ be the hidden size of LLM, and $m = 1,048,576$ be the latent size of SAE. We normalize the activations to $\sqrt{n}$, i.e., $\mathbb{E}\left[\|X\|\right] = \sqrt{n}$, where $X$ denotes the activations of the given layer.

After normalization, we filter out activations whose norm exceeds 20 times the batch-wise
mean, as such outliers would destabilize training.

We initialized every entry of the encoder matrix from $\mathcal{U}(-\frac{1}{\sqrt m}, \frac{1}{\sqrt m})$
and every entry of $W_{\mathrm{dec}}$ from $\mathcal{U}(-\frac{1}{\sqrt {n}}, \frac{1}{\sqrt {n}})$.

Our SAE is trained with activation function TopK. For the number of activations $\kappa$ retained by the activation function, we set $\kappa = 1000$ initially, and let $\kappa$ linearly decay to $100$ during the first $10{,}000$ steps, after which it is held constant. We used an auxiliary loss following~\citet{gao2025scaling} and~\citet{deng2026qwen} with the coefficient of $1/32$. The auxiliary loss always uses the same value of $\kappa$ as our main activation function. After training we normalize each column of $W_{\mathrm{dec}}$ to unit norm and rescale the corresponding row of $W_{\mathrm{enc}}$ by the reciprocal factor, which yields an equivalent SAE. We likewise fold the activation normalization factor into the weights, so the exported SAE applies directly to the raw layer output as in \Cref{eq:encode,eq:decode}.

We use fraction of variance unexplained (FVU) to evaluate our SAE. The FVU values of each layer are shown in \Cref{tab:FVU of sae}. Consistent with \citet{gao2025scaling}, we also observe that the intermediate layers of LLMs are more difficult to reconstruct using SAE, compared with both shallow and deep layers.

\begin{table}[t]
  \centering
  \caption{The FVU of our SAE on each layer. }
  \begin{tabular}{ll}
    \hline
    Layer & FVU\\
    \hline
    4 & $12.6\%$\\
    30 & $25.1\%$ \\
    46 & $7.6\%$\\
    \hline
  \end{tabular}

  \label{tab:FVU of sae}
\end{table}

\section{Pipeline of auto-interpretability}
\label{app:autointerp}
 
Following common practice, we use an LLM to produce a natural-language explanation for each latent. Unlike prior work that relies primarily on activated contexts, we provide the explainer with three sources of evidence: (i) activated contexts, (ii) logit effects, and (iii) layer information.
 
\paragraph{Activated contexts.}
We run inference on the 1.5B-token validation set of \Cref{sec:qwen3specialized}, which does not overlap with the SAE training data. For each latent, we collect all activated contexts within a 64-token window and select the top 5 by activation strength. We additionally sample 5 contexts uniformly at random from the remaining activations. We found that top-activating contexts are often highly similar to one another, which biases the explainer toward the literal surface pattern shared by those examples. Adding randomly sampled contexts increases diversity and encourages the explainer to summarize the latent's more general function rather than its most salient literal trigger. All ten contexts, along with their token-level activation values, are provided to the explainer.
 
\paragraph{Logit effects.}
To quantify a latent's direct effect on the output distribution, we apply the logit lens, projecting the latent's decoder vector through the LM head and extracting the top-10 boosted and top-10 suppressed tokens by logit value. Boosted tokens indicate what the latent promotes, while suppressed tokens indicate what it suppresses.
 
\paragraph{Layer information.}
We also tell the explainer the total number of layers, the layer at which the latent lives, and its relative depth, together with a soft prior on how features are typically distributed across depth: shallow layers tend to contain \emph{input features} that respond to specific tokens, local syntax, or surface patterns; middle layers tend to contain \emph{abstract features} that capture discourse patterns, reasoning states, or semantic categories; and deep layers tend to contain \emph{output features} that directly shape next-token prediction. This prior guides how the explainer weighs the evidence but is explicitly framed as a tiebreaker rather than a constraint. For input features, activated contexts are usually sufficient; for output features, the logit effects are more informative, since their activated contexts are often diffuse; for abstract features in the middle layers, neither source alone is typically conclusive, and the explainer is instructed to combine both. For output features, we adopt a fixed naming convention (\texttt{Say X} for boosted tokens and \texttt{Don't Say X} for suppressed tokens) so that output-oriented explanations are easily distinguishable from concept-level ones.
 
\paragraph{Output format.}
In addition to a concise explanation phrase, we ask the explainer to report a confidence rating on a three-level scale (\texttt{High}, \texttt{Medium}, \texttt{Low}) reflecting how well the provided evidence supports its explanation. The explainer is also allowed to output \texttt{Uninterpretable} when the activated contexts and logit effects do not exhibit a consistent, describable pattern, rather than being forced to produce a speculative explanation.
 
The full system prompt and user prompt are given in \Cref{app:autointerp_prompts}.
 
\subsection{Prompts}
\label{app:autointerp_prompts}
 
We provide the full system prompt and user prompt template below. In the user prompt, fields in double braces are filled in per latent.
 
\begin{promptbox}
You are an expert in mechanistic interpretability, specializing in interpreting features extracted by Sparse Autoencoders (SAEs) from a large language model.
 
## Task
 
Given one SAE feature, infer the concept, behavior, pattern, or output tendency it represents. Produce:
1. A concise interpretation phrase (2-6 words preferred; clarity > brevity).
2. A confidence rating.
 
---
 
## Input
 
You may receive up to three sections. Any may be missing.
 
### 1. Context Activations
Text examples that strongly activate the feature, with token-level activation highlights.
 
[Context 1]
[Text]
"The cat sat on the mat in Paris"
[/Text]
Top Activations: [mat|0.92] [Paris|0.87]
[/Context 1]
 
How to read them:
- Global pass: identify topic, domain, task, language, register, formatting, or discourse situation shared across examples.
- Local pass: examine the highest-activating tokens -- do they share a semantic field, syntactic role, position, or discourse function?
- Cross-example pass: isolate the single common thread that explains every activation. Reject hypotheses contradicted by any example.
 
### 2. Logit Effect
The feature's output influence at its own layer, presented as a direct logit-lens projection of the decoder vector: a list of boosted / suppressed tokens.
 
Use it to infer what the feature pushes the model to say or avoid saying. Logit effects can be noisy (especially in earlier layers), so weight context activations more heavily when they are clearer.
 
### 3. Layer Information
Soft prior based on layer depth:
- Shallow -> surface features: lexical items, local syntax, formatting, named entities, languages, domain terminology.
- Middle -> abstract / behavioral features: discourse patterns, reasoning states, task structure, style, intent, semantic categories.
- Deep -> output-oriented features: answer formats, completion templates, specific phrases, refusal / compliance, "say X" behaviors.
 
The layer prior is a tiebreaker, not a constraint. Override it whenever activations or logit effects give stronger evidence.
 
---
 
## Reasoning Procedure
 
1. Activations -- what topic / pattern, which tokens, what common thread?
2. Logit effect -- what is boosted, what is suppressed, is there a clear output push?
3. Layer prior -- what type of feature does the depth suggest?
4. Synthesis:
   - Activations + logit effect agree -> combine into one label.
   - They diverge -> take the clearer signal; do not force a unified story.
   - Both incoherent -> output `Uninterpretable`.
5. Phrase -- specific, 2-6 words, no filler.
6. Confidence -- see scale below.
 
---
 
## Output Format
 
Output only the two tags below. No reasoning, preamble, or commentary.
 
<explanation>concise interpretation phrase</explanation>
<confidence>High | Medium | Low</confidence>
 
Confidence scale:
- High -- activations and logit effect are coherent, repeated, and specific.
- Medium -- plausible interpretation; some evidence is partial, noisy, or only one channel is clear.
- Low -- weak, sparse, or contradictory evidence; label is tentative.
 
If evidence is too noisy, contradictory, or insufficient, output exactly:
 
<explanation>Uninterpretable</explanation>
<confidence>Low</confidence>
 
---
 
## Label Conventions
 
### Surface features (shallow)
Name the concept or pattern directly.
- `European capitals`
- `medical terminology`
- `Python indentation`
- `Chinese punctuation`
- `HTML attribute syntax`
 
### Behavioral features (middle)
Name the behavior, capability, discourse function, or abstract state.
- `formal register`
- `negation logic`
- `Chinese language response`
- `step-by-step reasoning`
- `instruction-following mode`
- `uncertainty expression`
 
### Output-oriented features (deep)
- Boosted output -> prefix with `Say`: `Say however`, `Say comma`, `Say closing parenthesis`, `Say final answer`.
- Suppressed output -> prefix with `Don't Say`: `Don't Say period`, `Don't Say apology`, `Don't Say the`.
 
Do not substitute `Produce`, `Express`, `Output`, `Generate`, etc. for `Say`.
 
---
 
## Specificity Rules
 
Be precise. Replace vague labels with concrete ones:
 
| Avoid                | Prefer                          |
|----------------------|---------------------------------|
| `locations`          | `European capitals`             |
| `science-related`    | `scientific terminology`        |
| `punctuation tokens` | `sentence-ending punctuation`   |
| `negative words`     | `negation / denial`             |
| `medical stuff`      | `medical symptom description`   |
 
Banned patterns: `xxx-related`, `xxx tokens`, `xxx words`, `xxx stuff`, `xxx things`, and meta-descriptions of the feature itself (e.g. `activates on nouns`, `high activation feature`).
 
---
 
## Honesty Rule
 
Do not fabricate coherence. When evidence does not support a specific interpretation, output the `Uninterpretable` template above. Forced labels are worse than honest abstention.
\end{promptbox}
 
\begin{promptbox}
## Context Activations
Two complementary sets of examples are provided: the strongest (top) activations and a random sample. Read both -- the random set guards against over-fitting to the most extreme cases.
 
### Top Activating Contexts
The 5 examples where this feature fires most strongly.
 
[Top Context 1]
[Text]
{{top_context_1.text}}
[/Text]
Top Activations: {{top_context_1.top_activations}}
[/Top Context 1]
 
... (Top Context 2-5 in the same format)
 
### Random Contexts
5 examples sampled at random from this feature's activating set (a less-biased view of where it fires).
 
[Random Context 1]
[Text]
{{random_context_1.text}}
[/Text]
Top Activations: {{random_context_1.top_activations}}
[/Random Context 1]
 
... (Random Context 2-5 in the same format)
 
## Logit Effect
Source: decoder-vector projection. This feature lives at layer {{layer_idx}} (single decoder; SAE-style).
Decoder L2 norm = {{decoder_l2_norm}}
 
Top 10 boosted logits:
Rank 1: [{{token}}], logit={{logit}}
... (Rank 2-10)
 
Top 10 suppressed logits:
Rank 1: [{{token}}], logit={{logit}}
... (Rank 2-10)
 
## Model Information
Model: {{model_name}} -- {{num_layers}} transformer layers (0-indexed).
Feature extractor: Sparse Autoencoder (SAE), trained at layer {{layer_idx}}. Each feature has a single decoder vector at its own layer.
 
## Layer Information
The model has {{num_layers}} layers in total (0-indexed); the current feature's layer index is {{layer_idx}} (relative depth {{layer_idx / num_layers}}). Use the relative position as a soft prior, but override it with the logit-effect signal whenever the two disagree.
 
Analyze the context activations, logit effect, and model / layer information, then respond in the exact output format specified in the system prompt above.
\end{promptbox}

\end{document}